\documentclass[sigconf]{acmart}
\pdfoutput=1
\usepackage{color}
\usepackage{marvosym}

\usepackage{amsmath,amssymb}
\usepackage[utf8]{inputenc}
\usepackage{graphicx}
\usepackage{subfigure}
\usepackage{booktabs}
\usepackage{witharrows}
\usepackage{multirow}
\usepackage{pifont}

\usepackage{bbding}
\def\ie{\emph{i.e.}}
\def\etal{\emph{et al.}}
\def\eg{\emph{e.g.}}
\AtBeginDocument{%
  \providecommand\BibTeX{{%
    \normalfont B\kern-0.5em{\scshape i\kern-0.25em b}\kern-0.8em\TeX}}}

\copyrightyear{2021}
\acmYear{2021}
\setcopyright{acmlicensed}
\acmConference[MM '21] {Proceedings of the 29th ACM Int'l Conference on Multimedia}{October 20--24, 2021}{Chengdu, Sichuan Province, China.}
\acmBooktitle{Proceedings of the 29th ACM Int'l Conference on Multimedia (MM '21), Oct. 20--24, 2021, Chengdu, Sichuan Province, China}
\acmPrice{15.00}
\acmISBN{978-1-4503-8651-7/21/10}
\acmDOI{10.1145/XXXXXX.XXXXXX}

\begin{document}
\fancyhead{}
\title{
% RELE: Reinforcement Learning for Zero-shot Low-Light Image Enhancement
% ReLLIE: Towards Customized Low-Light Image Enhancement via Deep Reinforcement Learning}
ReLLIE: Deep Reinforcement Learning for Customized Low-Light Image Enhancement}

%
% The "author" command and its associated commands are used to define
% the authors and their affiliations.
% Of note is the shared affiliation of the first two authors, and the
% "authornote" and "authornotemark" commands
% used to denote shared contribution to the research.
% \author{Rongkai Zhang, Lanqing Guo}
% \affiliation{%
% \institution{Institute for Clarity in Documentation}
%  \streetaddress{P.O. Box 1212}
%  \state{Ohio}en
%  \country{USA}
% }, lanqing001@e.ntu.edu.sg, siyu.hang@ntu.edu.sg, bihan.wen@ntu.edu.sg
\author{Rongkai Zhang}
\authornote{Both authors contributed equally to this research.}
\email{rongkai002@e.ntu.edu.sg}
\author{Lanqing Guo}
\authornotemark[1]
\email{lanqing001@e.ntu.edu.sg}
\affiliation{\institution{Nanyang Technological University}}
\author{Siyu Huang}
\affiliation{\institution{Nanyang Technological University}}
\email{siyu.huang@ntu.edu.sg}
\author{Bihan Wen}
\authornote{Bihan Wen is the corresponding author.}
\affiliation{\institution{Nanyang Technological University}}
\email{bihan.wen@ntu.edu.sg}

\begin{abstract}
Low-light image enhancement (LLIE) is a pervasive yet challenging problem, since: 1) low-light measurements may vary due to different imaging conditions in practice;
2) images can be enlightened subjectively according to diverse preference by each individual. To tackle these two challenges, this paper presents a novel deep reinforcement learning based method, dubbed ReLLIE, for customized low-light enhancement. ReLLIE models LLIE as a markov decision process, \textit{i.e.}, estimating the pixel-wise image-specific curves sequentially and recurrently. Given the reward computed from a set of carefully crafted non-reference loss functions, a lightweight 
network is proposed to estimate the curves for enlightening a low-light image input. As ReLLIE learns a policy instead of one-one image translation, it can handle various low-light measurements and provide customized enhanced outputs by flexibly applying the policy different times. Furthermore, ReLLIE can enhance real-world images with hybrid corruptions, \textit{e.g.}, noise, by using a plug-and-play denoiser easily. Extensive experiments on various benchmarks demonstrate the advantages of ReLLIE, comparing to the state-of-the-art methods. (Code is
available:~\href{https://github.com/GuoLanqing/ReLLIE}{https://github.com/GuoLanqing/ReLLIE}.)
%benefiting from the simple network, 
\end{abstract}





\maketitle
\section{Introduction}
% \begin{figure}[!t]
% 	\begin{center}
% 			\includegraphics[width=.325\linewidth]{img/intro/1.png}
% 			\includegraphics[width=.325\linewidth]{img/intro/3.png}
% 			\includegraphics[width=.325\linewidth]{img/intro/2.png} \\
% 			(a) Low-light inputs of different degeneration degrees \\
% 			\includegraphics[width=.325\linewidth]{img/intro/6.png}
% 			\includegraphics[width=.325\linewidth]{img/intro/5.png}
% 			\includegraphics[width=.325\linewidth]{img/intro/4.png} \\
% 			(b) Customized enhancement outputs by ReLLIE
% 	\end{center}
% 	\vspace{-3mm}
% \caption{The proposed ReLLIE can deal with (a) low-light image inputs of different degeneration degrees and output (b) customized enhancement results using different enhancement parameters.}
% % \vspace{-2mm}
% \label{fig:intro} 
% \end{figure}
\begin{figure}[t]
  \centering
  \includegraphics[width=\linewidth]{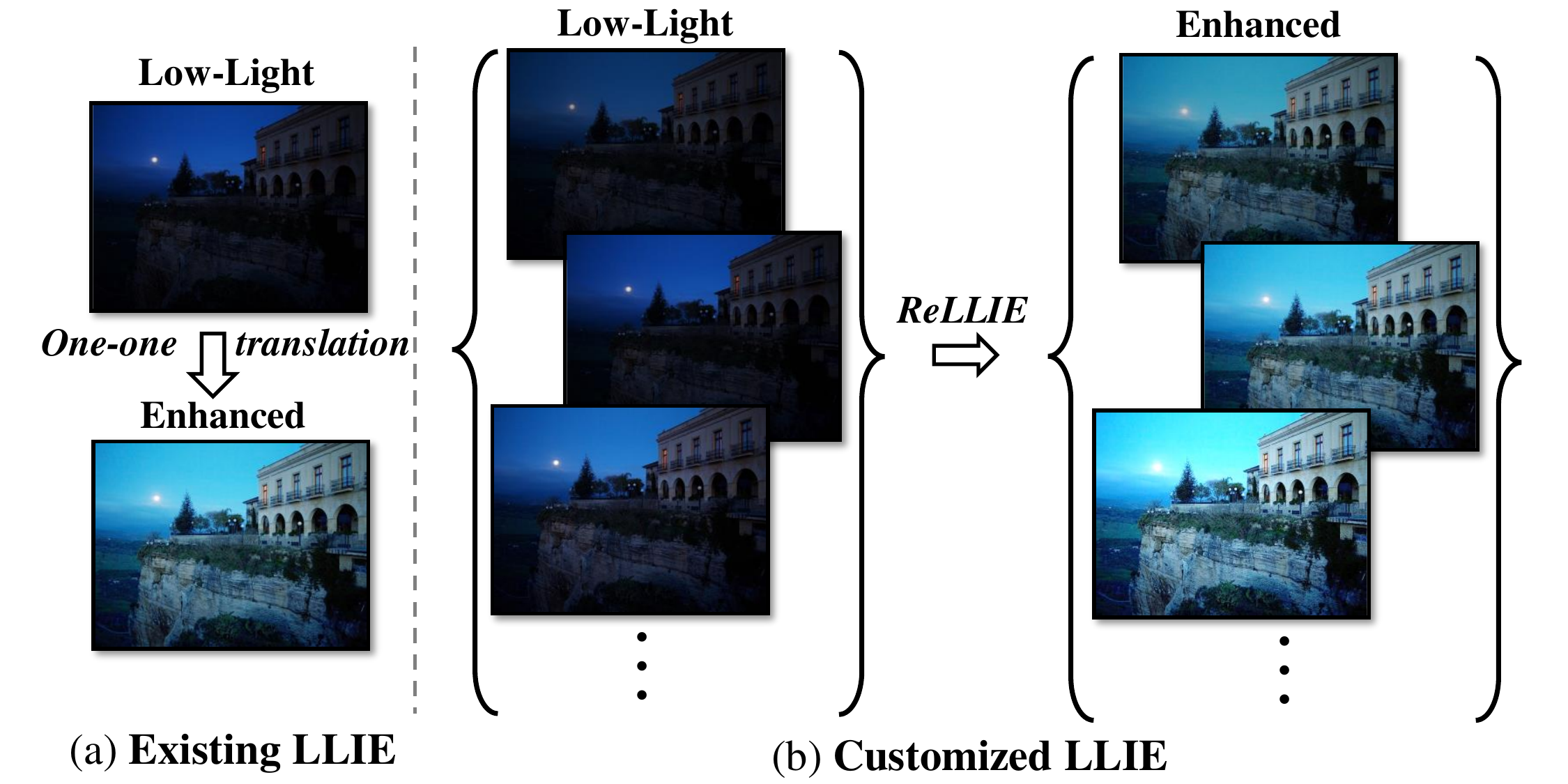}
  \caption{(a) Most of the existing LLIE methods produce one-one image translation; (b) We propose a customized LLIE scheme using the ReLLIE method.}
%   \Description{A woman and a girl in white dresses sit in an open car.}
\label{fig:intro}
\end{figure}

Low-light images captured under insufficient lighting conditions are pervasive in real-life scenarios due to inevitable environmental/technical constraints. Suffering from compromised aesthetic quality and unsatisfactory transmission of information, such low-light images are forbidden from many computer vision applications which therefore motivate plenty of low-light image enhancement (LLIE) methods \cite{guo2016lime,lore2017llnet,zhang2019kindling,gharbi2017deep,jiang2019enlightengan,guo2020zero,zheng_2020_ECCV}. Based on Retinex theory~\cite{land1977retinex, wei2018deep, zhu2020zero}, a low-light image can be modeled by the following degradation process:
\begin{equation}
    S_{low} = R \circ I_{low} + n_{add},
    \label{eq:1}
\end{equation}
where $S_{low}$ is the low-light image, $R$ denotes the underlying reflectance, $I_{low}$ is the insufficient illumination, $n_{add}$ is the additive noise, and $\circ$ denotes the element-wise multiplication. The LLIE task aims to recover the ``optimal'' illumination $I_{opt}$ from the observation $S_{low}$ with the consistent reflectance $R$, meanwhile, suppressing the noise $n_{add}$. Most of the existing methods establish a one-one image translation model under the assumption that there exists only one deterministic output for an input. However, as an intrinsic nature, the LLIE task is complicated in practice, since both $S_{low}$ and $I_{opt}$ may be diverse for different individuals/applications. As shown in Fig. \ref{fig:intro}, an LLIE method should be more customized that it can 1) handle inputs $S_{low}$ with \emph{varied degrees of degeneration} (which are possibly different from the training data) and 2) provide candidate outputs with \emph{different subjective} $I_{opt}$ so as to meet the preference of different users.
% However, such simplification may distort some important intrinsic nature of this task. Firstly, low-light image enhancement is highly subjective that both the low-light observation $S_{low}$ and the "optimal" illumination $I_{opt}$ may vary largely and dependent on different persons and applications, \eg, the example shown in Figure \ref{fig:intro}. 

%captured/assessed 
%, ignoring the customization of LLIE task

%However, such simplification may distort the intrinsic nature of LLIE task due to the following aspects:
%(a) Most of the existing methods build a one-one image translation model under the assumption that there exists only one deterministic output for an input, ignoring the subjective assessment of LLIE task;
%(b) The learning-based LLIE methods require sufficient paired images for training which may be expensive to collect in practice. The zero-shot LLIE \cite{li2020zero} would be a better alternative for real-world scenarios.
%(c) The additional noise $n_{add}$ is usually handled by the supervision of ground truth. To the best of our knowledge, the hybrid degeneration is rarely explored in existing literature of unsupervised/zero-shot LLIE.
%, which largely limits their applications in practice. \par
%except the recent work \cite{zhu2020zero} which considers additional noise during enhancement, 
 %The curve used is carefully formulated so that 1) it is differentiable, 2) it maintains the range of the enhanced image, and 3) preserves the contrast of neighboring pixels. 
 
In this paper, we present a deep reinforcement learning (DRL) based method named ReLLIE, to achieve more customized LLIE results. Instead of simply performing one-one paired image translation, ReLLIE reformulates LLIE as a \emph{sequential image-specific curve estimation problem}. Specifically, ReLLIE takes a low-light or intermediate image as input and produces second-order curves as its output at each step following a learned policy. The policy is parameterized by a lightweight fully convolutional network and trained using a set of non-reference loss functions specially designed for LLIE. In a recurrent manner, ReLLIE employs the image-specific curves to deliver a robust and accurate dynamic range adjustment. 

To the best of our knowledge, ReLLIE is the first non-reference DRL based method for pixel-wise LLIE. Compared to the existing methods, ReLLIE has the following advantages. Firstly, ReLLIE learns a more flexible stochastic policy other than the deterministic one-one image translation. It can deal with inputs of different low-light degrees and provide customized enhancement outputs. The number of enhancement steps can be flexibly determined by the users (\ie~ less or more than which used in training).
Secondly, while existing deep LLIE methods require large-scale paired images or additional high-quality images for training which are expensive to collect. ReLLIE adopts non-reference loss functions as its reward function such that it does not require any paired or even unpaired data in its training process. Therefore, ReLLIE enables non-reference \cite{guo2020zero} and zero-shot \cite{guo2020zero}  image enlightening which are more flexible for real-world scenarios.
Thridly, ReLLIE can be flexibly equipped with additional enhancement modules, \eg,~denoiser, to tackle the hybrid image degeneration according to personalized preference.
%To the best of our knowledge, this is the first work handling the additive noise $n_{add}$ in unsupervised/zero-shot LLIE scnarios.
%Besides the training, with the supervision of ground truth, can be handled,
In extensive experiments, we show that ReLLIE can perform on par with other existing LLIE methods that require paired or unpaired data for training. ReLLIE also achieves the state-of-the-art performance on zero-shot scenarios.\par
%Finally, thanks to the lightweight network, ReLLIE can be effectively applied to zero-shot learning, which makes it more suit the real-world scenarios. 
%It is noteworthy that recent work has adopted similar ideas \cite{guo2020zero}. However, our method is more customized and significantly differs from it well in the following aspects.
%\textcolor{red}{list contributions}
Our contributions are summarized as follows.
\begin{enumerate}
    \item Recognizing the gap between real-world scenarios and the limitations of existing LLIE methods, we present a DRL based lightweight framework namely ReLLIE, towards a more customized LLIE scheme.
    \item Accompanied with ReLLIE, we propose a new non-reference LLIE loss namely channel-ratio constancy loss (CRL) and a new channel dependent momentum update (CDMU) module, for training more robust LLIE models. We also propose enhancement-guided refinement (RF) module to handle the additive noise in LLIE scenarios.
    \item Extensive experiments show that the proposed ReLLIE can be effectively applied to zero-shot and unsupervised LLIE benchmarks. 
\end{enumerate}

\section{Related Work}
\subsection{Deep Reinforcement Learning for Image Restoration}
Recently, DRL has gathered considerable interest in image processing tasks. For instance, Yu \etal~ \cite{yu2018crafting} proposes RL-Restore to learn a policy for selecting appropriate tools from predefined toolbox to progressively restore the quality of a corrupted image. However, it requires sufficient paired training data to train the agent using $\mathcal{L}_2$ loss function. More related to this work, Park \etal~ \cite{park2018distort} proposes a DRL based color enhancement method to tackle the need of paired data via a ``distort-and-recover'' training scheme. Their scheme only requires high-quality reference images for training instead of input and retouched image pairs. In parallel with \cite{park2018distort}, Hu \etal~\cite{hu2018exposure} enables a paired image-free photo retouching method with DRL and generative adversarial networks (GANs).
While these methods focus on global image restoration, Furuta \etal~\cite{furuta2019fully} proposes pixelRL to enable pixel-wise image restoration which is more flexible. More recently, Zhang \etal~\cite{zhang2021r3l} proposes R3L, which applies DRL to pixel-wise image denoising via direct residual recovery. However, the aforementioned methods all require the external set of ``high-quality'' training images, which can be highly limited in practice. Furthermore, no work to date has exploited DRL for LLIE problem.

\subsection{Low-Light Image Enhancement}
The LLIE task aims to increase the image visibility so as to benefit a series of downstream tasks including classification, detection, and recognition. 
Histogram equalization (HE)~\cite{abdullah2007dynamic} and its follow-ups~\cite{lee2013contrast} achieve uniformly contrast improvement by spreading out the most frequent intensity values, providing undesirable amplified noise.
Later on,
Retinex theory~\cite{land1977retinex}, which assumes an image can be decomposed into reflectance and illumination, has been widely used in traditional illumination-based methods~\cite{fu2016fusion,guo2016lime}. 
For instance,
NPE~\cite{wang2013naturalness} jointly enhances contrast and illumination, and LIME~\cite{guo2016lime} proposes a structure-aware smoothing model to estimate the illumination map.
These hand-craft methods impose priors on the decomposed illumination and reflectance, which achieve impressive results in illumination adjustment but presenting intensive noises and artifacts.

Recently, the deep learning based methods commonly apply high-quality normal-light ground truth as guidance to learn how to improve low-light image~\cite{wei2018deep,lore2017llnet,zhang2019kindling}. 
LL-Net~\cite{lore2017llnet} proposes a stacked auto-encoder to simultaneously conduct denoising and enhancement using synthesized low/normal-light image pairs. However, the distribution of synthetic data inevitably deviates from real-world images due to the domain gap, leading to severe performance degradation when transferring to real-world cases.
Later on, Wei \etal~\cite{wei2018deep} collects a real-world dataset with low/normal-light image pairs, based on which the Retinex-Net is proposed to decompose images into illumination and reflectance in a data-driven way.
Following that, various other neural networks~\cite{zhang2019kindling,Yang_2020_CVPR} have been proposed for supervised LLIE.
% These methods rely on a large number of paired dataset, which is hard to collect in practice.
More recent methods \cite{jiang2019enlightengan} focus on unsupervised LLIE which directly enlightens low-light images without any paired training data. The very recent Zero-DCE~\cite{guo2020zero} trains the deep LLIE model using non-reference losses. However, existing deep methods produce one-one image mapping for LLIE, while neglecting different low-light imaging conditions in practice and diverse subjective preference by each individual.
% However, these methods may fail when it comes to some low-quality or extremely dark cases. 

Our proposed ReLLIE is significantly different from other counterparts by achieving a more customized LLIE via learning a stochastic enhancement policy rather than the one-one image translation model. The enhancement operation can be conducted multiple times, which is highly flexible for real-world scenarios. In addition, ReLLIE can be applied to zero-shot and unsupervised LLIE scenarios by employing the non-reference losses as reward function. 

\begin{figure*}[!ht]
\centering 
\includegraphics[width=1\textwidth,height=.4\linewidth]{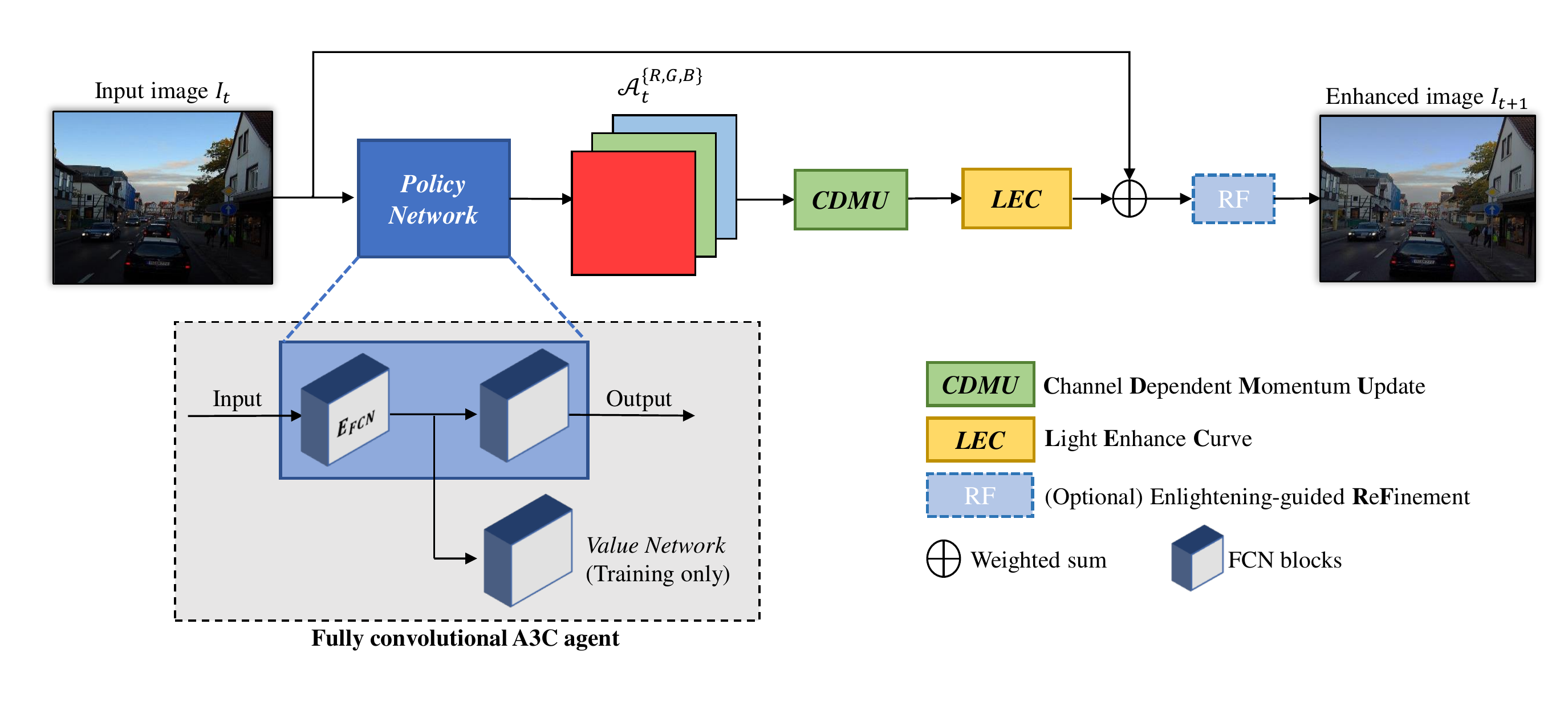}
\vspace{-4em}
\caption{Illustration of the ReLLIE pipeline by taking the $t$-th enhancement step as an example. 
%The agent is the fully convolutional policy network trained with A3C~\cite{mnih2016asynchronous}.
}
\label{fig:self-MM}
\end{figure*}

\section{Problem Definition}
\subsection{LLIE via Curve Adjustment}
LLIE can be achieved by human experts via applying the curve adjustment in photo editing software, where the self-adaptive curve parameters are solely dependent on the input images. The optimal curves for challenging low-light images are often of very high order. Zero-DCE~\cite{guo2020zero} suggests this procedure can be realized equally by recurrently applying the low-order curves. In this work, we apply a second-order light enhancement curve (LEC) at each step, which can be formulated as:
\begin{equation}
    \text{LE}(I(\mathrm{x}); \textbf{A}(\mathrm{x})) = I(\mathrm{x}) + \textbf{A}(\mathrm{x}) I(\mathrm{x})(1-I(\mathrm{x})),
\end{equation}
where $I$ is the input low-light image and $\mathrm{x}$ denotes the pixel coordinates. $LE(I(\mathrm{x}); \textbf{A}(\mathrm{x}))$ outputs the enhanced image at $\mathrm{x}$, using the learned feature parameter $\textbf{A}(\mathrm{x})$, which has the same size as the image. LE can be applied multiple times to approximate higher-order LEC. At the $t$-th step ($t \geq 1$), the enhanced output is:
\begin{equation}
    \text{LE}_t(\mathrm{x}) = \text{LE}_{t-1}(\mathrm{x}) + \textbf{A}_t(\mathrm{x}) \text{LE}_{t-1}(\mathrm{x})(1-\text{LE}_{t-1}(\mathrm{x})),
    \label{eq:output}
\end{equation}
which models the enhancement of a low-light image as a sequential decision making problem by finding the optimal pixel-wise parameter map $\textbf{A}_t(\mathrm{x})$ at each step $t$.

\subsection{LLIE as Markov Decision Process}
\label{sec:MDP}
Based on (3), we show that LLIE can be  formulated as a markov decision process (MDP)~\cite{sutton2018reinforcement} consisting of the task-specific \emph{state}, \emph{action} and \emph{reward}. 
%Commonly, sequential decision making problems, \ie~enhancement via curve adjustment, can be modelled as Markov Decision Process (MDP) and addressed by deep reinforcement learning (DRL). 
%\textcolor{red}{MDP is a ... consisting of state, action, ...}

\noindent\emph{state}: At each step $t$, the low-light image $I_t \in \mathbb{R}$ is the \emph{state} ($s_t \in \mathcal{S}$), where $t=0$ denotes the initial state with raw inputs and $t \geq 1$ denotes the intermediate states with partially enhanced images from the previous step. 
\noindent\emph{action}: The \emph{action} at $s_t$ is to select a parameter $\alpha_t(\mathrm{x})$ for the LEC of each pixel, where $\alpha_t(\mathrm{x})$ is constrained in a predefined range $\mathcal{A}$ and all $\alpha_t$ constitute a parameter map $\textbf{A}_t(\mathrm{x})$. Applying a sequence of parameter maps to the input raw images results in a trajectory $T$ of \emph{states} and \emph{actions}:
\begin{equation*}
    T= (s_0,\textbf{A}_0,s_1,\textbf{A}_1,\cdots,s_{N-1},\textbf{A}_{N-1},s_N,\textbf{A}_N),
\end{equation*}
where $N$ is the number of steps, and $s_N$ is the stopping state. 
\noindent\emph{reward}: The \emph{reward} $r: \mathcal{S} \times \mathcal{A} \to \mathbb{R}$ evaluates the actions given a \emph{state}. Our goal is to obtain a policy $\pi$ that maximizes the accumulated \emph{reward} during the MDP. To this end, we employ a stochastic policy agent parameterized by $\pi_{\theta}(\textbf{A}_t|s_t)$ with trainable parameters $\theta$. The policy $\pi_{\theta}: \mathcal{S} \to \mathbb{P}(\mathcal{A})$ maps the current \emph{state} $s_t \in \mathcal{S}$ to $\mathbb{P}(\mathcal{A})$ the set of probability density functions over the \emph{actions}, as $P(\textbf{A}_t|s_t)$. In summary, when an agent enters a \emph{state}, it samples one \emph{action} according to the probability density functions, receives the \emph{reward}, and transits to the next \emph{state}.

More specifically, given a trajectory $T$, the return $r_k^\gamma$ is the summation of discounted \emph{rewards} after $s_k$:
\begin{equation}
    r_k^\gamma = \sum_{k' = 0}^{N-k} \gamma^{k'}r(s_{k+k'},\textbf{A}_{k+k'}) ,
\end{equation}
where $\gamma \in [0,1]$ is a discount factor, which places greater importance on \emph{rewards} in the nearer future. To evaluate a policy, we have the following objective:
\begin{equation}
    J(\pi_{\theta}) = \mathbb{E}_{s_0 \sim \mathcal{S}_0}[r_0^{\gamma}|\pi_{\theta}],
    \label{eq:objective}
\end{equation}
where $s_0$ is the input image and $\mathcal{S}_0$ is the input distribution, \eg, a dataset. Intuitively, the objective in Eq. \ref{eq:objective} describes the expected return over all possible trajectories induced by the policy $\pi_\theta$. The goal of the agent is to maximize the objective $J(\pi_\theta)$, which is related to the final image quality defined by \emph{reward} $r$, since images (\emph{states}) with a higher quality are more greatly rewarded.

%The non reference losses used are detailed in section 4.2.
%Here, inspired by \cite{guo2020zero},

\section{Proposed ReLLIE}
\subsection{Agent}
%FCNs are widely adopted in pixel-level image processing tasks.
With the MDP formulation of LLIE, we can apply a DRL based agent to conduct such task.  
Inspired by~\cite{furuta2019fully}, we employ fully convolutional networks (FCNs) based asynchronous advantage actor-critic (A3C) \cite{mnih2016asynchronous} framework as our stochastic policy agent. The overall framework of ReLLIE is depicted in Fig. ~\ref{fig:self-MM}. In A3C, we use a policy network $\pi_\theta$ and a value network $V_{\theta_v}$ to make DRL training more stable and efficient \cite{NIPS1999_464d828b}. The FCN-based encoder $\text{E}_\text{FCN}$ extracts the features of the input image $I_t$ then outputs $s_t$, the representation of state $t$. $\text{E}_\text{FCN}$ is shared by both $\pi_\theta$ and $V_{\theta_v}$. Taking $s_t$, the policy network $\pi_\theta$ outputs the probability $P(\textbf{A}_t|s_t,\theta_\pi)$, from which a parameter map $\textbf{A}_t (\mathrm{x})$ is sampled. The value network outputs $V_{\theta_v}(s_{t})$ which is an estimation of the long term discounted rewards:
\begin{equation}
   V_{\theta_v}(s_t) = \mathbb{E}_{s_0=s_t}\left[r_0^\gamma\right].
\end{equation}
We also include a skip link in ReLLIE to make the update of the input image $I_t$ a weighted sum of raw input image $I_0$ and the enhanced one. The update process is
\begin{equation}
    I_{t} = \omega LE_t(\mathrm{x}) + (1-\omega)I_0,
\end{equation}
where $\omega$ is a tunable parameter and empirically set as 0.8. After color enhancement, our framework includes an optional denoising module (which can be arbitrary image enhancing method) for further enhancement. 

Without loss of generality, we consider the one-step learning case ($N=1$) here for convenience. The gradients of the parameters of these two networks $\theta_\pi$,$\theta_v$ are calculated as:
\begin{equation}
\begin{aligned}
     r_t^\gamma &= r_t+\gamma V(s_{t+1}),\\
     d\theta_v &= \nabla_{\theta_v}(r_t^\gamma-V_{\theta_v}(s_t))^2,\\
     d\theta_\pi &= -\nabla_{\theta_\pi}\log P(\textbf{A}_t|s_t,\theta_\pi)(r_t^\gamma-V_{\theta_v}(s^t)).
\end{aligned}
\end{equation}

\begin{figure}[!ht]
\centering 
\includegraphics[width=0.8\linewidth]{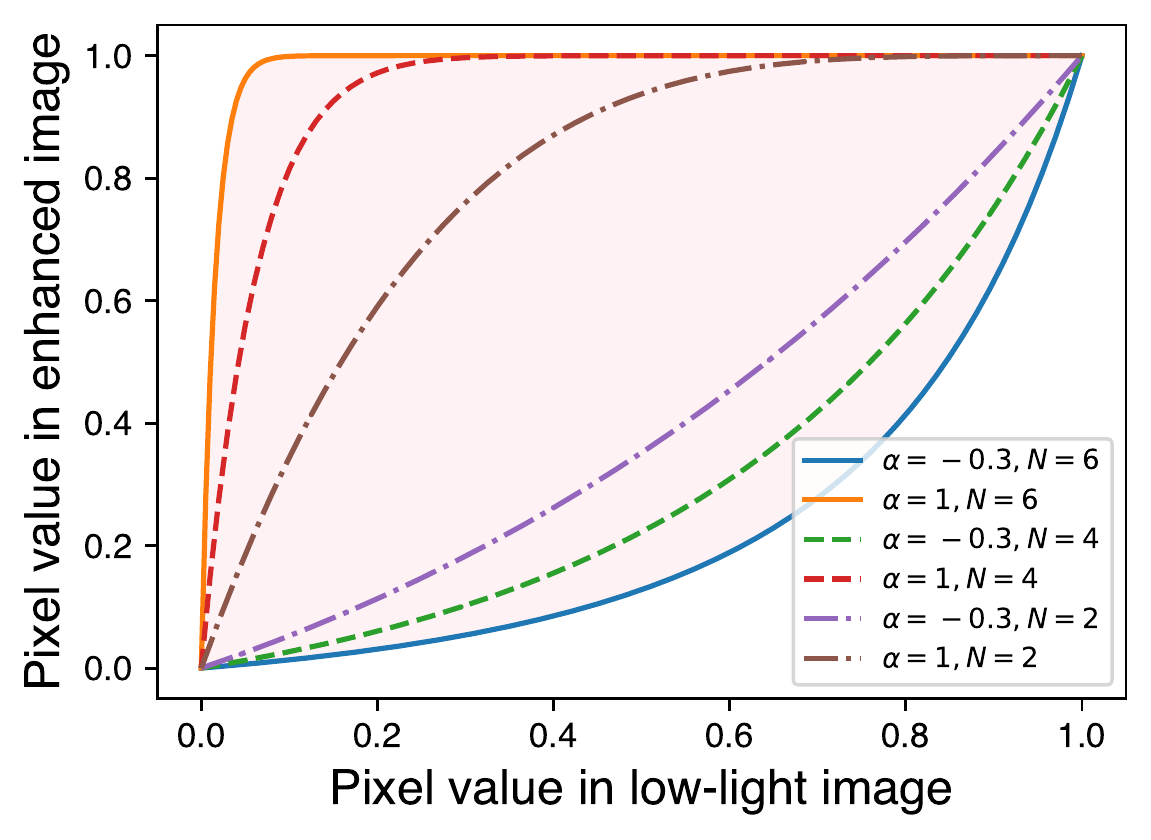}
\vspace{-1em}
\caption{Illustration of how the adjustment ranges with different $N$ and action range $\mathcal{A} \in \left[-0.3, 1\right]$. }
%changed for pixels from low-light images to the enhanced one }
\label{fig:range}
\end{figure}

\noindent
\textbf{Action space.} As mentioned in Section \ref{sec:MDP}, the \emph{action} for \emph{state} $s_t$ selects a parameter $\alpha_t(\mathrm{x})$ for LEC of a pixel, where $\alpha_t(\mathrm{x})$ is constrained in a predefined range $\mathcal{A}$ and all $\alpha_t$ constitute the parameter map $\textbf{A}_t(\mathrm{x})$. The range $\mathcal{A}$ is critical for the performance of our agent, since a too narrow range results in a limited enhancement while a too wide one results in a exhaustively large search space. Here, we empirically set the range $\mathcal{A} \in \left[-0.3, 1\right]$ with graduation as 0.05. This setting ensures that 1) each pixel is in the normalized range of $\left[0,1\right]$ and 2) LEC is monotonous. Meanwhile, it alleviates the cost of searching suitable LEC for low-light image enhancement. Fig.~\ref{fig:range} shows that LEC can effectively cover the pixel value space under the proposed action space setting, with respect to different choices of $N$.\par
%as we focus on low-light image enhancement, using the prior that images are enhanced to be \emph{brighter}, 
\noindent
\textbf{Reward.} Many metrics have been proposed for image quality assessment,~\eg, the $\mathcal{L}_2$ distance between enhanced/groundtruth outputs and the adversarial loss learned from a predefined set of ``high-quality" images. In this work, we adopt four non-reference losses to assess an enhanced image and use the negative weighted sum of them as the \emph{reward} to train our agent. On one hand, the using of non-reference losses gets rid of the need of expensively collected \emph{paired} data and even does not require the so-called ``high-quality'' images. On the other hand, a weighted sum of different non-reference losses 
introduces more flexibility for user preference.

\subsection{Non-Reference Losses}
%aiming at the specific low-light image enhancement task, 
For zero-reference LLIE, spatial consistency loss, exposure control loss, and illumination smoothness loss are exploited in \cite{guo2020zero}. In addition to these losses, in this work we propose a new non-reference loss, namely channel-ratio constancy loss  (CRL), for more robust and effective learning of zero-reference LLIE models. We discuss the details of the four losses in the following.
%The four different losses enable a . 

%Applying these four types of losses, we evaluate the quality of enhanced images and train our agent. The details of these four losses are given below:\par
\noindent
\textbf{Spatial consistency loss.} The spatial consistency loss $L_{spa}$ encourages the preservation of the difference among neighboring regions during the enhancement:
\begin{equation}
    L_{spa} = \frac{1}{K}\sum_{i=1}^K\sum_{j \in \Omega(i)}(|Y_i-Y_j|-|I_i-I_j|)^2,
\end{equation}
where $K$ is the number of local region and $\Omega(i)$ is the four neighboring regions (top, down, left, right) centered at the region $i$. $Y$ and $I$ denote the average intensity value of the local region in the enhanced version and input image, respectively. Here, the local region is set to 4x4 empirically.

\noindent
\textbf{Exposure control loss.} The exposure control loss $L_{exp}$ measures the distance between the average intensity value of a local region to a predefined well-exposedness level $E$, \ie, the gray level in the RGB color space \cite{mertens2009exposure}. It is written as:
\begin{equation}
    L_{exp} = \frac{1}{M}\sum_{k=1}^M|Y_m - E|,
\end{equation}
where $M$ represents the number of non-overlapping local regions of size 16×16, $Y_m$ is the average intensity value of a local region $m$ in the enhanced image. According to \cite{guo2020zero}, $E$ is set to 0.6.

\noindent
\textbf{Illumination smoothness loss.} To avoid aggressive and sharp changes between neighboring pixels, we employ illumination smoothness loss $L_{tvA}$ to control the curve parameter map \textbf{A} at every state, as:
\begin{equation}
    L_{tvA} = \frac{1}{N}\sum_{t=1}^N\sum_{c \in \epsilon}(|\nabla_x \textbf{A}_t^c|+|\nabla_y \textbf{A}_t^c|)^2, \epsilon = {R,G,B},
\end{equation}
where $N$ is the number of iteration and $\nabla_x$ and $\nabla_y$ denote the horizontal and vertical gradient operations, respectively.

\noindent
\textbf{Channel-ratio constancy loss.} In addition to the above three losses, we propose a channel-ratio constancy loss $L_{crl}$ to constrain the ratio among three channels to prevent potential color deviations in the enhanced image. CRL $L_crl$ is formulated as:
\begin{equation}
    L_{crl} = \sum (|\frac{I_R}{I_G}-\frac{Y_R}{Y_G}|+|\frac{I_R}{I_B}-\frac{Y_R}{Y_B}|+|\frac{I_G}{I_B}-\frac{Y_G}{Y_B}|)^2,
\end{equation}
where $\frac{I_R}{I_G}$ denotes the pixel-wise ratio between $R$ channel and $G$ channel of input image $I$, $\frac{Y_R}{Y_G}$ denotes the pixel-wise ratio between $R$ channel and $G$ channel of enhanced one $Y$, and $\sum$ denotes the summation of all the ratios. $L_{crl}$ constrains the intrinsic ratio among channels of the input images and thereby avoiding color casts.
%constrains the potential independent aggressive enhancement among channels and 

\noindent
\textbf{Agent reward.} The total learning objective is
\begin{equation}
    L_{total} = W_{spa}L_{spa} + W_{exp}L_{exp}  + W_{tvA}L_{tvA}+ W_{crl}L_{crl},
\end{equation}
where $W_{spa}$, $W_{exp}$, $W_{tvA}$ and $W_{crl}$ are tunable parameters which can be set according to user preference. Hence, for a given enhanced image, the \emph{reward} $r$ at a certain state $s_t$ is
\begin{equation}
    r(s_t,\textbf{A}_t) = - L_{total}(s_{t+1}).
\end{equation}

\subsection{Channel Dependent Momentum Update}

We further propose a channel dependent momentum update (CDMU) for color images with RGB channels. At each state, the agent outputs $\textbf{A}_R(\mathrm{x})$, $\textbf{A}_G(\mathrm{x})$, $\textbf{A}_B(\mathrm{x})$ for the pixel in different channels respectively. The real parameter maps $\textbf{A}_R^*(\mathrm{x})$, $\textbf{A}_G^*(\mathrm{x})$, and $\textbf{A}_B^*(\mathrm{x})$ applied to each channel is computed as:
\begin{equation}
    \begin{aligned}
         \textbf{A}_R^*(\mathrm{x}) &= \textbf{A}_R(\mathrm{x}),\\
         \textbf{A}_G^*(\mathrm{x}) &= \omega_{CD} \textbf{A}_G(\mathrm{x}) + (1-\omega_{CD}) \textbf{A}_R(\mathrm{x}),\\
         \textbf{A}_B^*(\mathrm{x}) &= \omega_{CD} \textbf{A}_B(\mathrm{x}) + (1-\omega_{CD}) \textbf{A}_R(\mathrm{x}).\\
    \end{aligned}
\end{equation}
where $\omega_{CD}$ is a tunable parameter which controls the dependence among channels. It is reasonable to perform CDMU among different channels, since in natural images the RGB channels are usually related to each other. Such update avoids aggressive modifications on an individual channel which may result in unbalanced tone performance. Note that any a single channel can be used as the reference channel, \ie,~$\textbf{A}_R$. The ablation study in Section 5.2 reveals that a totally independent update leads to tone failure and unstable training.

\subsection{Enlightening-guided Recursive Refinement}
%, even though doing so may further improve the visual pleasure
For low-light images, the degeneration model can be hybrid in practice. For instance, the image noise in shadows may become more pronounced after operations of brightness bossting. However, rare existing methods consider explicit denoising during enlightening process. To this end, this work introduces an optional denoising block to perform enlightening-guided recursive refinement (RF). In general, many existing denoisers can be good candidates for the denoising block. In light of the competitive performance of pretrained FFDNet~\cite{zhang2018ffdnet}, we adopt FFDNet as the denoiser block, as well as an additional noise level map as a guidance to handle the spatially variant noise. Here the noise level map refers to the ratio that each pixel enlightened, inspired by the empirical evidence the noise level map can indicate the degree of involved noise~\cite{zhu2020zero}.

%since no specific denoiser is used to train the agent
We note that the denoising blocks are \emph{totally optional} in our framework, as they are not involved in the training process. Our agent learns the policy in a ``denoising-free'' setting, and users can use FFDNet to denoise the enhanced images optionally at each step of the testing phase. Such regime not only makes training more stable, but also allows a larger flexibility of using other denoisers in testing phase. Moreover, compared with the supervised one-one image translation methods, our method allows to address various types of degeneration rather than the noise by simply employing the restoration methods accordingly.

\begin{figure*}[!t]
	\begin{center}
		\begin{tabular}{c@{ }c@{ }c@{ }c@{ }c@{ }c@{ }c@{ }c}
			\includegraphics[width=.23\linewidth, height=.15\linewidth]{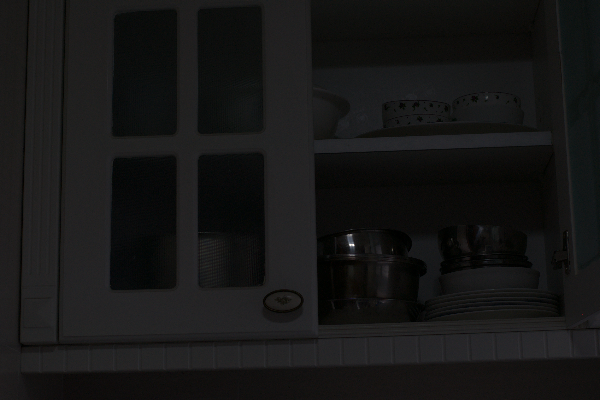}~&
			\includegraphics[width=.23\linewidth, height=.15\linewidth]{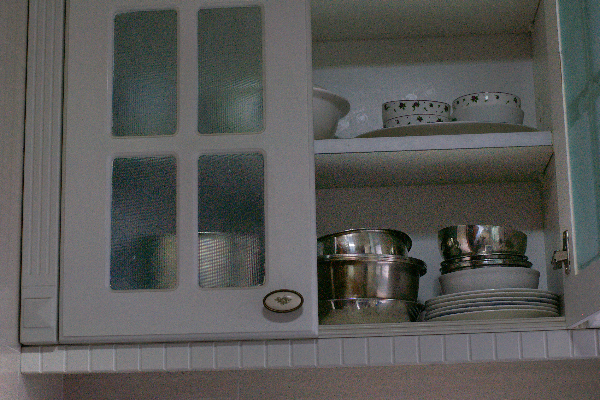}~&
			\includegraphics[width=.23\linewidth, height=.15\linewidth]{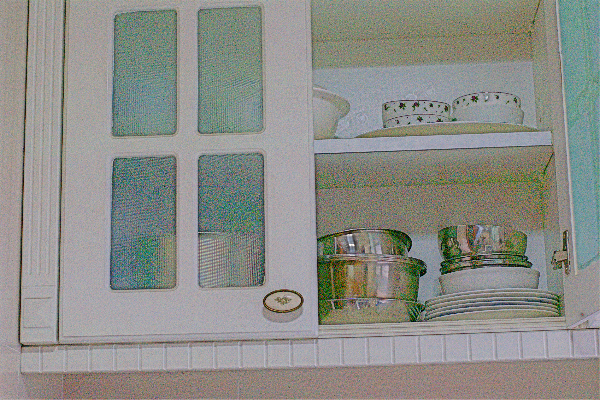}~&
			\includegraphics[width=.23\linewidth, height=.15\linewidth]{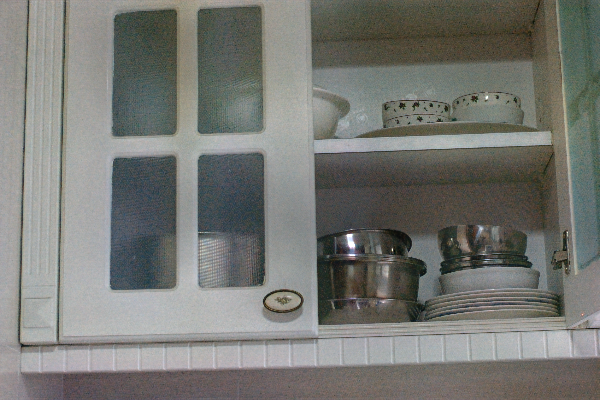}\\
% 			(a) Input~& (b) LIME~& (c) RetinexNet~& (d) EnlightenGAN\\
				\includegraphics[width=.23\linewidth, height=.15\linewidth]{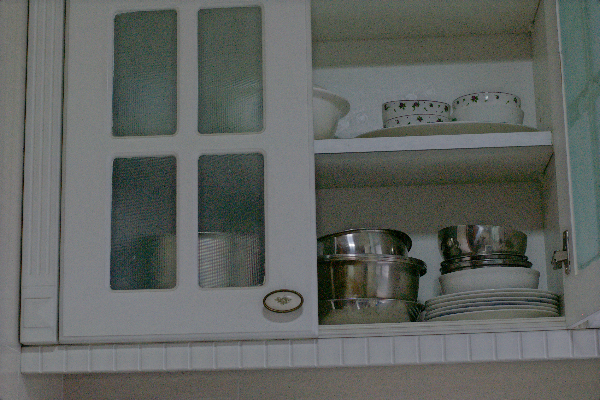}~&
			\includegraphics[width=.23\linewidth, height=.15\linewidth]{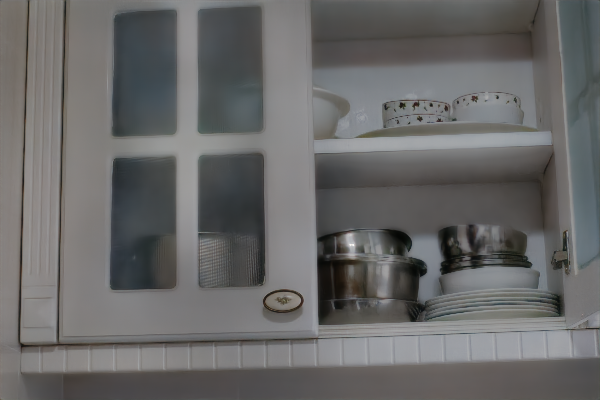}~&
			\includegraphics[width=.23\linewidth, height=.15\linewidth]{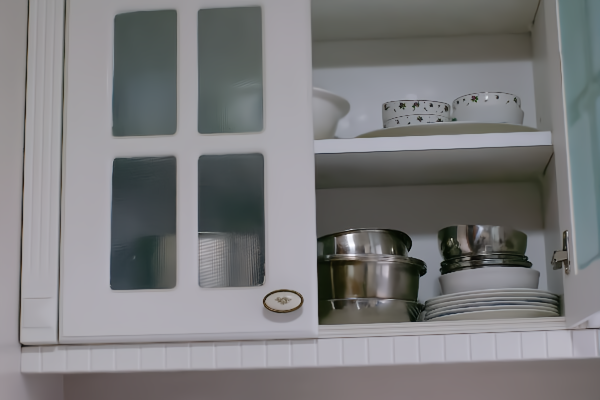}~&
			\includegraphics[width=.23\linewidth, height=.15\linewidth]{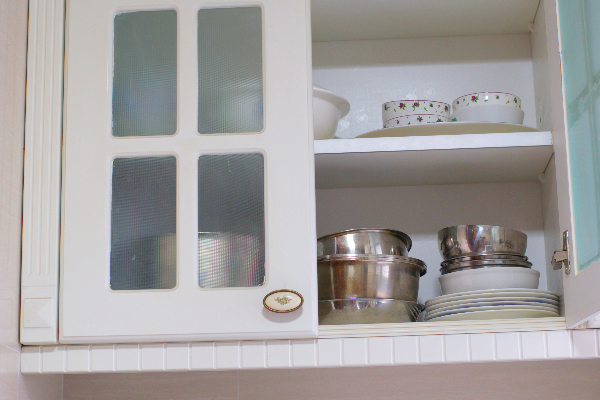}\\
% 			(e) Zero-DCE~& (f) KinD~& (g) Ours~& (h) GT\\
				\includegraphics[width=.23\linewidth, height=.15\linewidth]{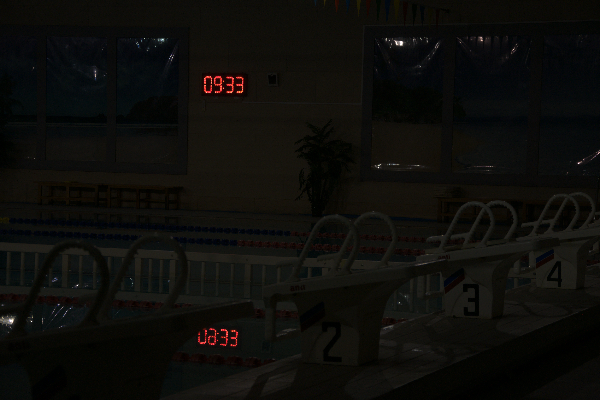}~&
			\includegraphics[width=.23\linewidth, height=.15\linewidth]{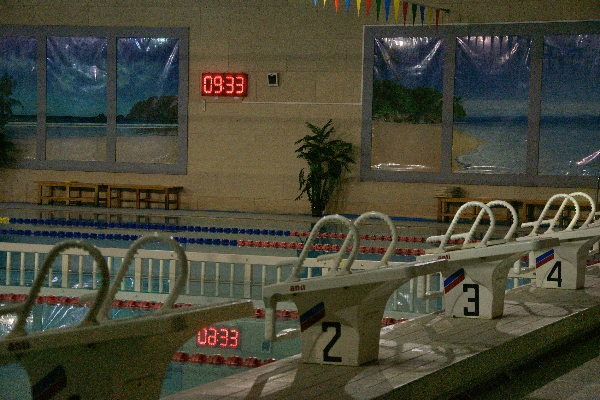}~&
			\includegraphics[width=.23\linewidth, height=.15\linewidth]{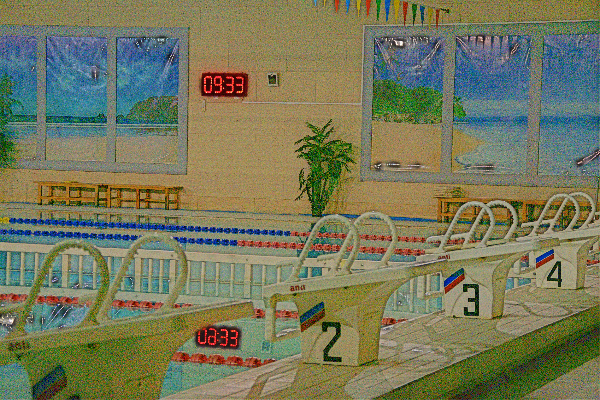}~&
			\includegraphics[width=.23\linewidth, height=.15\linewidth]{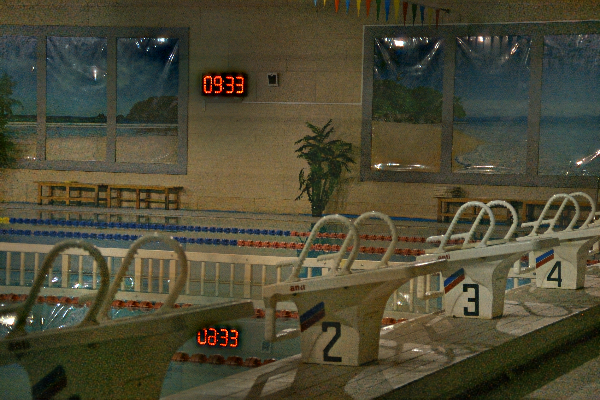}\\
% 			(a) Input~& (b) LIME~& (c) RetinexNet~& (d) EnlightenGAN\\
				\includegraphics[width=.23\linewidth, height=.15\linewidth]{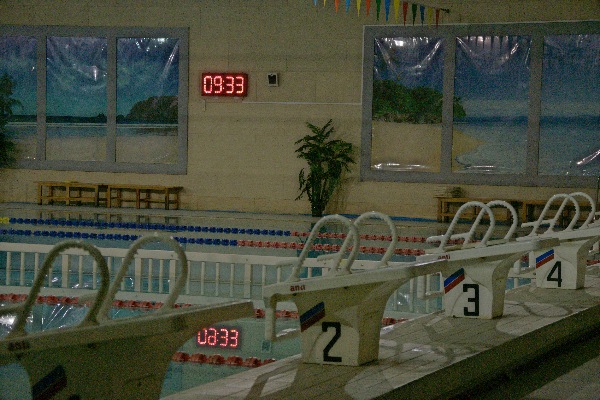}~&
			\includegraphics[width=.23\linewidth, height=.15\linewidth]{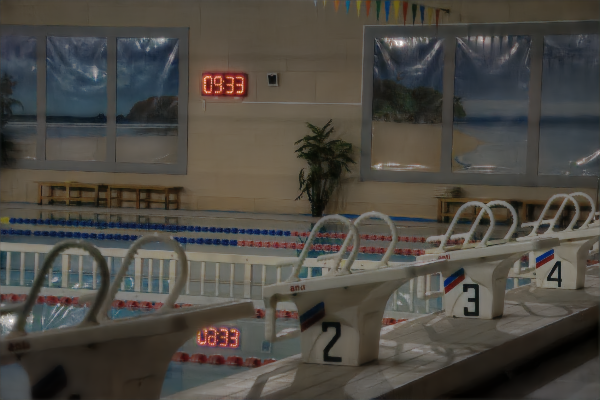}~&
			\includegraphics[width=.23\linewidth, height=.15\linewidth]{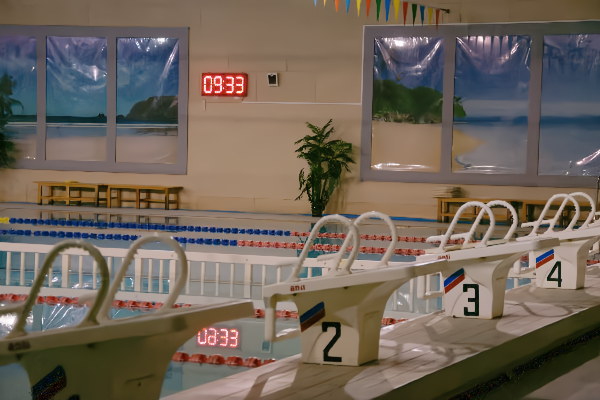}~&
			\includegraphics[width=.23\linewidth, height=.15\linewidth]{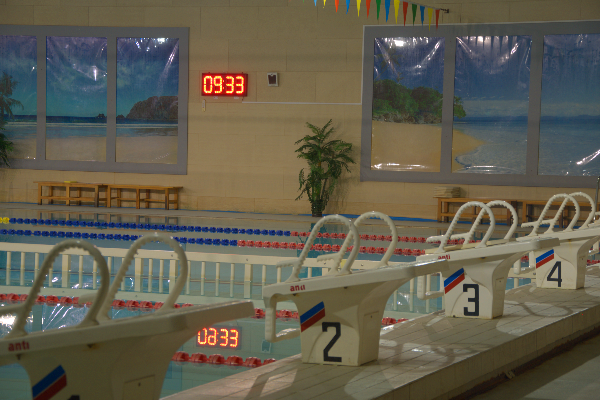}\\
% 			(e) Zero-DCE~& (f) KinD~& (g) Ours~& (h) GT\\
			\includegraphics[width=.23\linewidth, height=.15\linewidth]{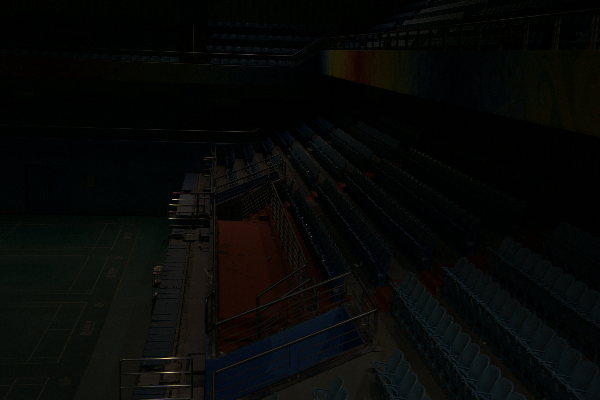}~&
			\includegraphics[width=.23\linewidth, height=.15\linewidth]{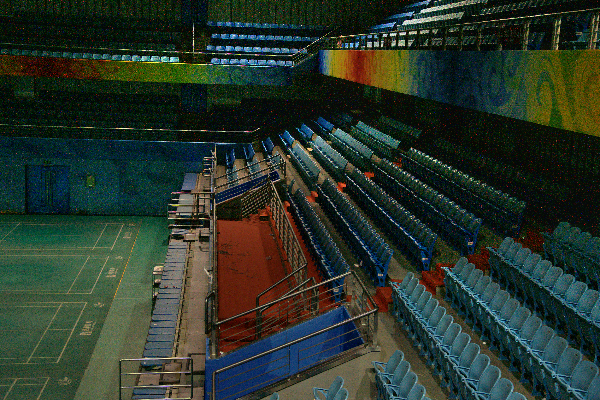}~&
			\includegraphics[width=.23\linewidth, height=.15\linewidth]{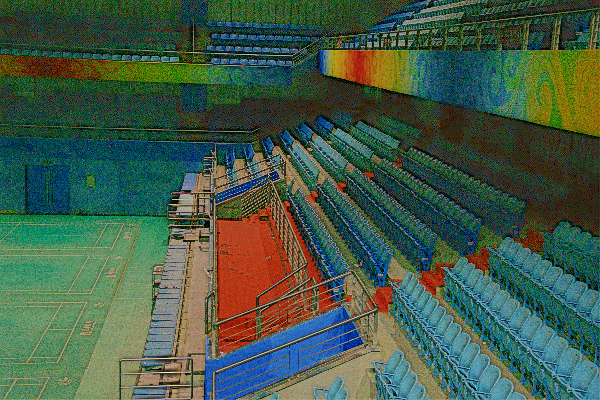}~&
			\includegraphics[width=.23\linewidth, height=.15\linewidth]{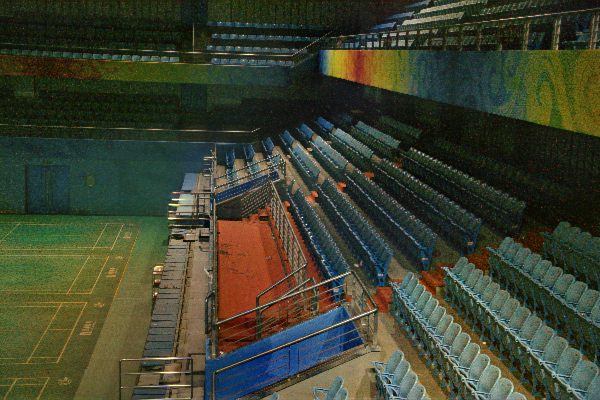}\\
				% 		(a) Input~& (b) LIME~& (c) RetinexNet~& (d) EnlightenGAN\\
				\includegraphics[width=.23\linewidth, height=.15\linewidth]{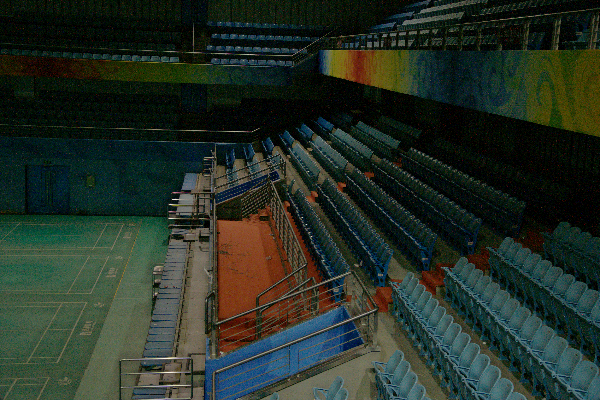}~&
			\includegraphics[width=.23\linewidth, height=.15\linewidth]{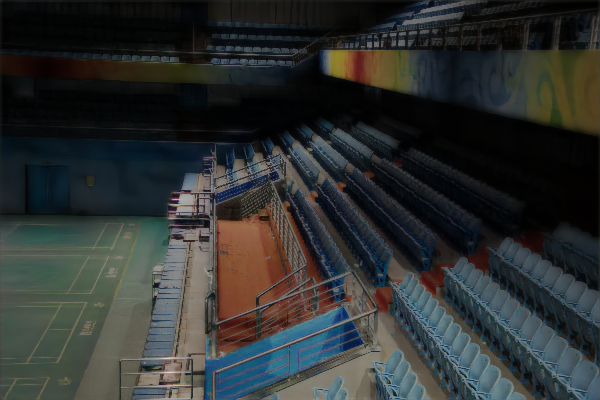}~&
			\includegraphics[width=.23\linewidth, height=.15\linewidth]{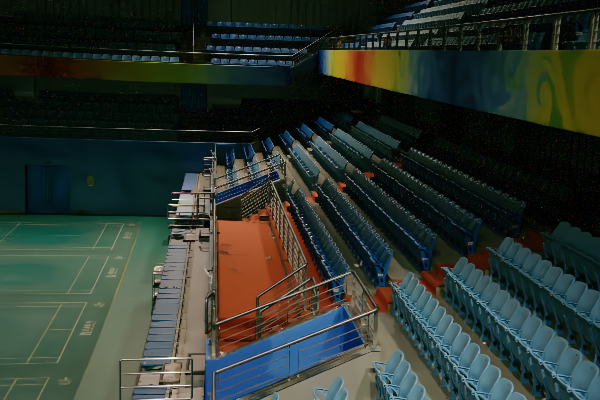}~&
			\includegraphics[width=.23\linewidth, height=.15\linewidth]{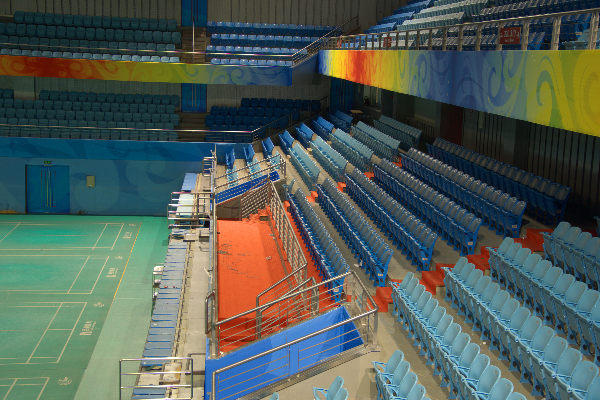}\\
% 			(e) Zero-DCE~& (f) KinD~& (g) Ours~& (h) GT\\
		\end{tabular}
	\end{center}
	\vspace{-1em}
	\caption{Examples of enhancement results on LOL evaluation dataset. For each two rows: Input image, LIME~\protect\cite{guo2016lime}, RetinexNet~\protect\cite{wei2018deep}, EnlightenGAN~\protect\cite{jiang2019enlightengan}, Zero-DCE~\protect\cite{guo2020zero}, 
	KinD~\protect\cite{zhang2019kindling},
	ReLLIE (ours),
	ground truth. 
	Zoom in to better see the details.
% 	Examples of enhancement results on LOL evaluation dataset. We show the estimated results of (b) LIME~\protect\cite{guo2016lime}, (c) RetinexNet~\protect\cite{wei2018deep}, (d) EnlightenGAN~\protect\cite{jiang2019enlightengan}, (e) Zero-DCE~\protect\cite{guo2020zero}, (f) KinD~\protect\cite{zhang2019kindling}, and (g) Our ReLLIE. Zoom in to see the details.
	}
	\label{fig:lol}
\end{figure*}

\begin{figure*}[!t]
	\begin{center}
		\begin{tabular}{c@{ }c@{ }c@{ }c@{ }c@{ }c}
			\includegraphics[width=.18\textwidth]{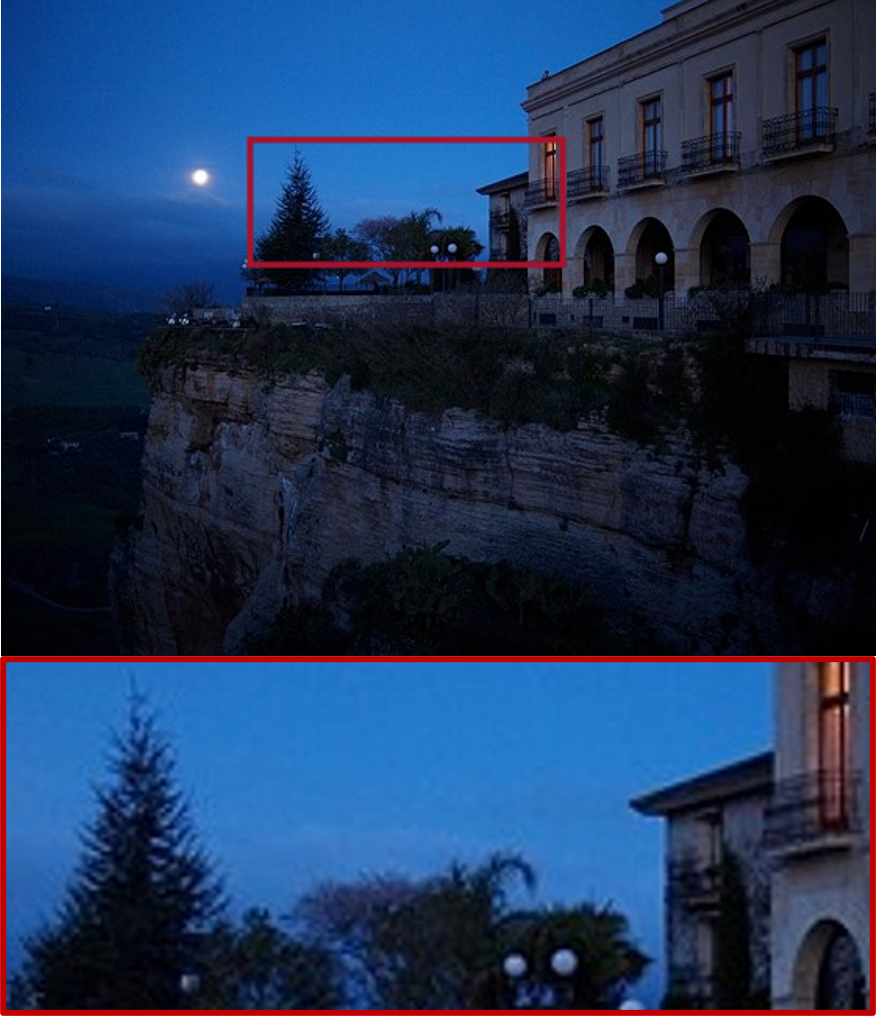}~&
			\includegraphics[width=.18\textwidth]{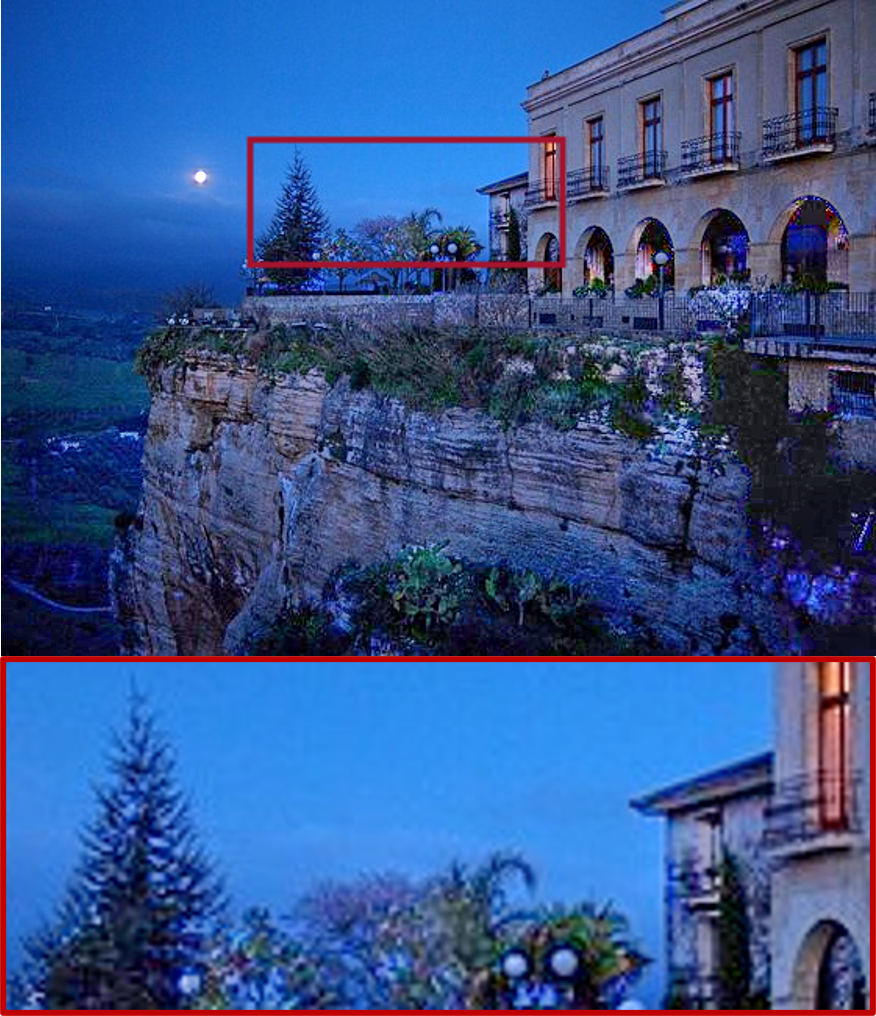}~&
			\includegraphics[width=.18\textwidth]{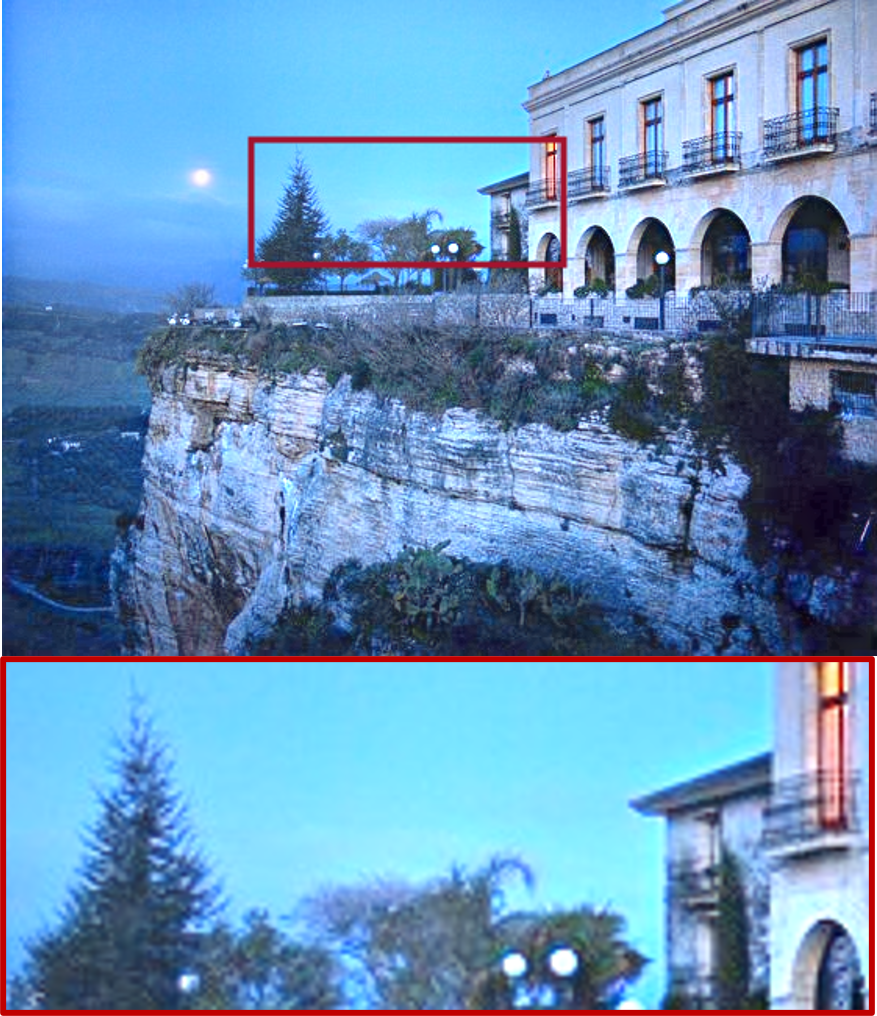}~&
			\includegraphics[width=.18\textwidth]{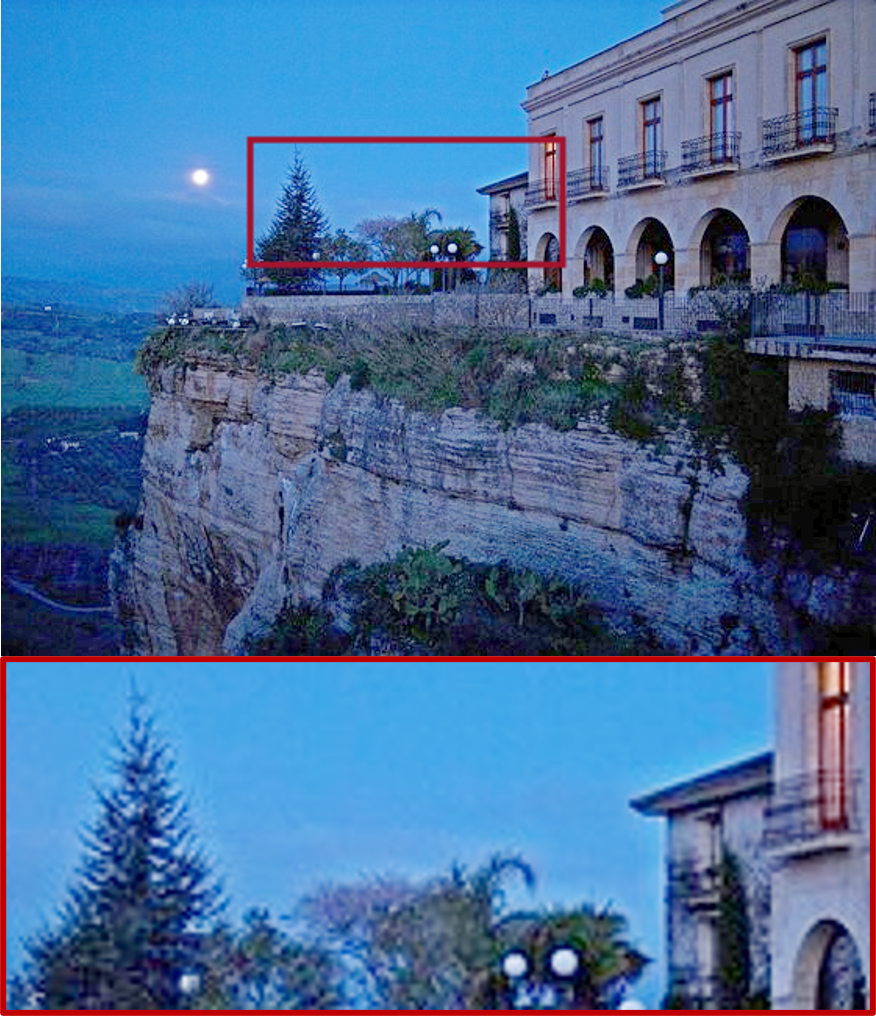}~&
			\includegraphics[width=.18\textwidth]{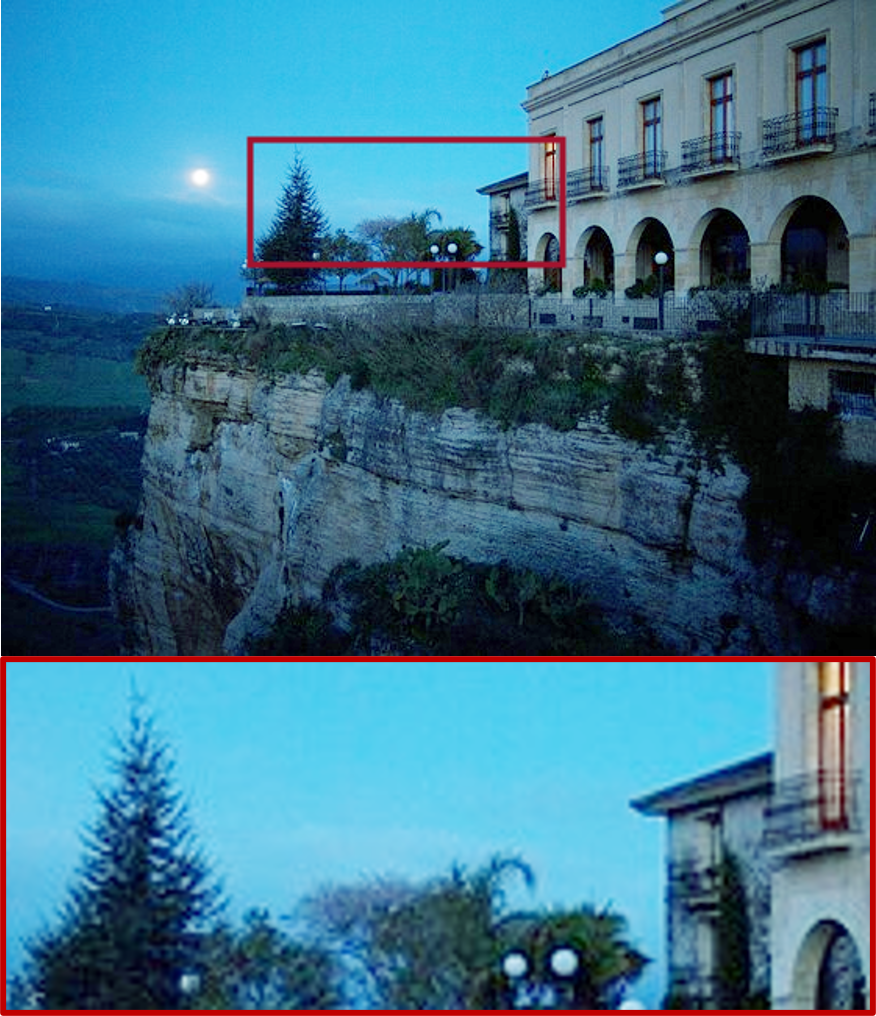}\\
						\includegraphics[width=.18\textwidth]{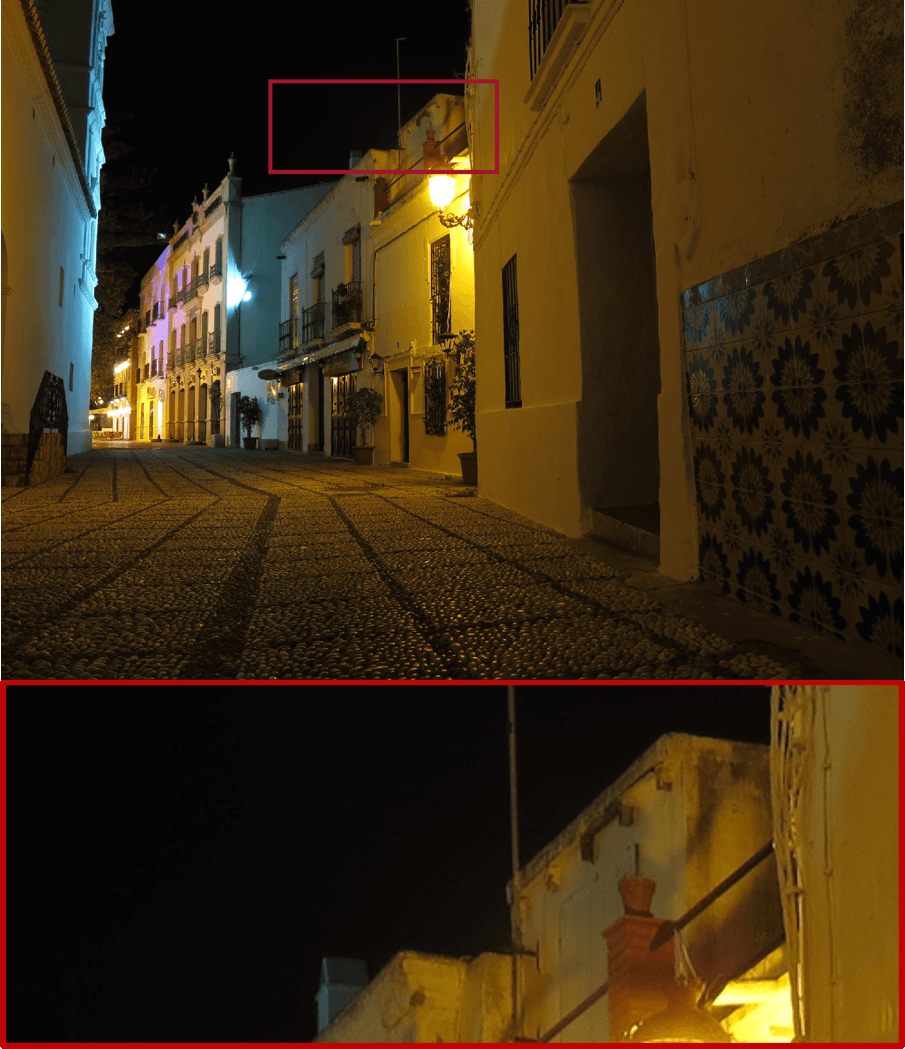}~&
			\includegraphics[width=.18\textwidth]{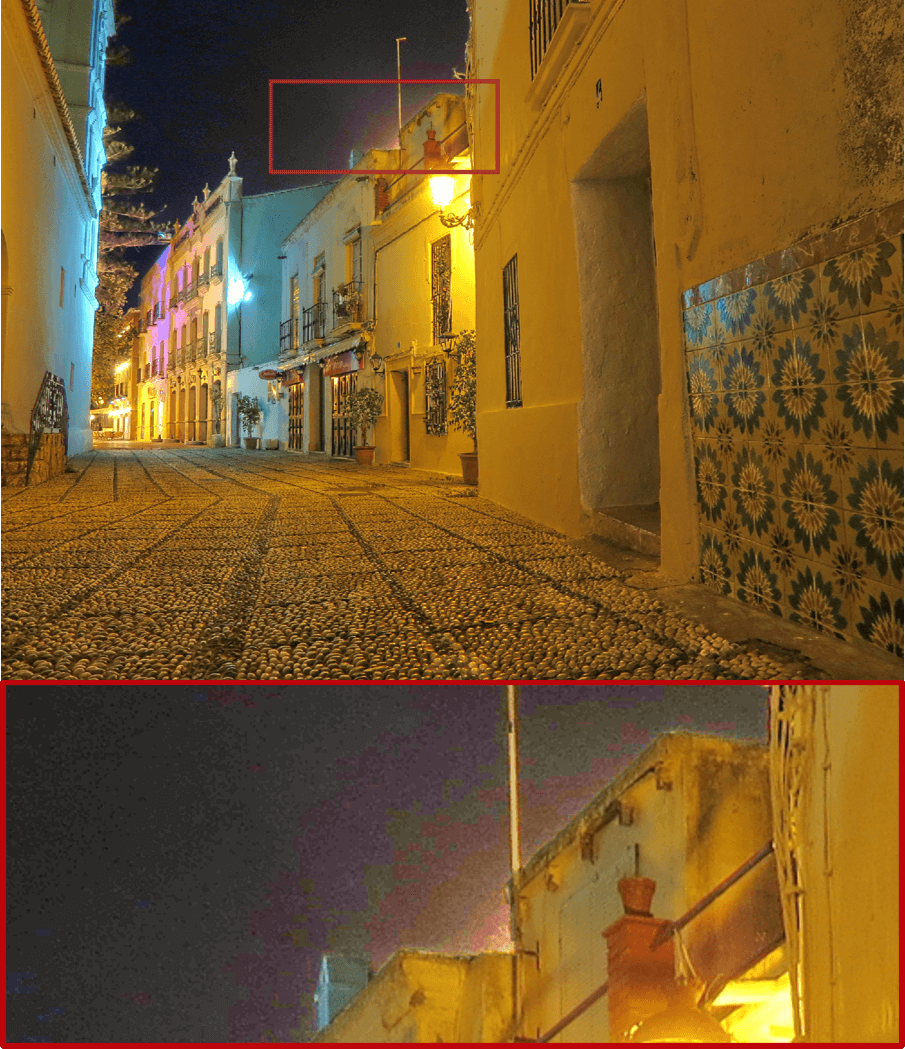}~&
			\includegraphics[width=.18\textwidth]{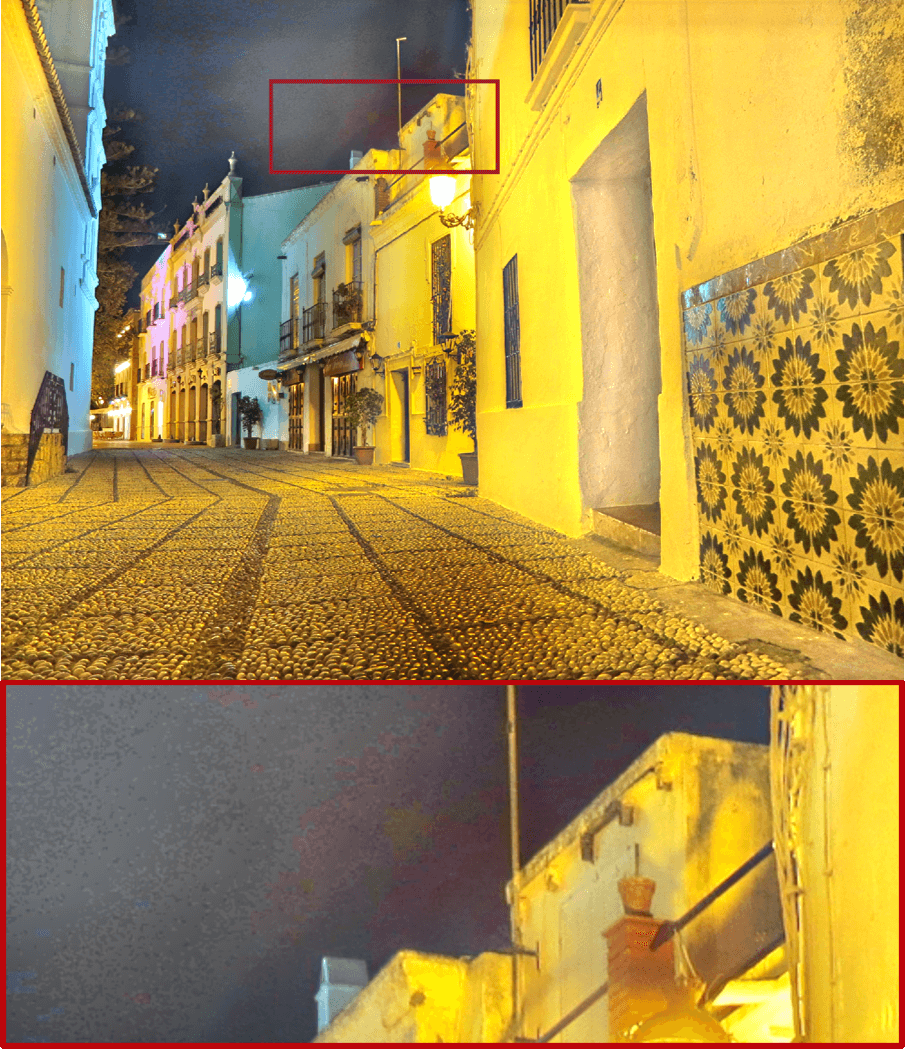}~&
			\includegraphics[width=.18\textwidth]{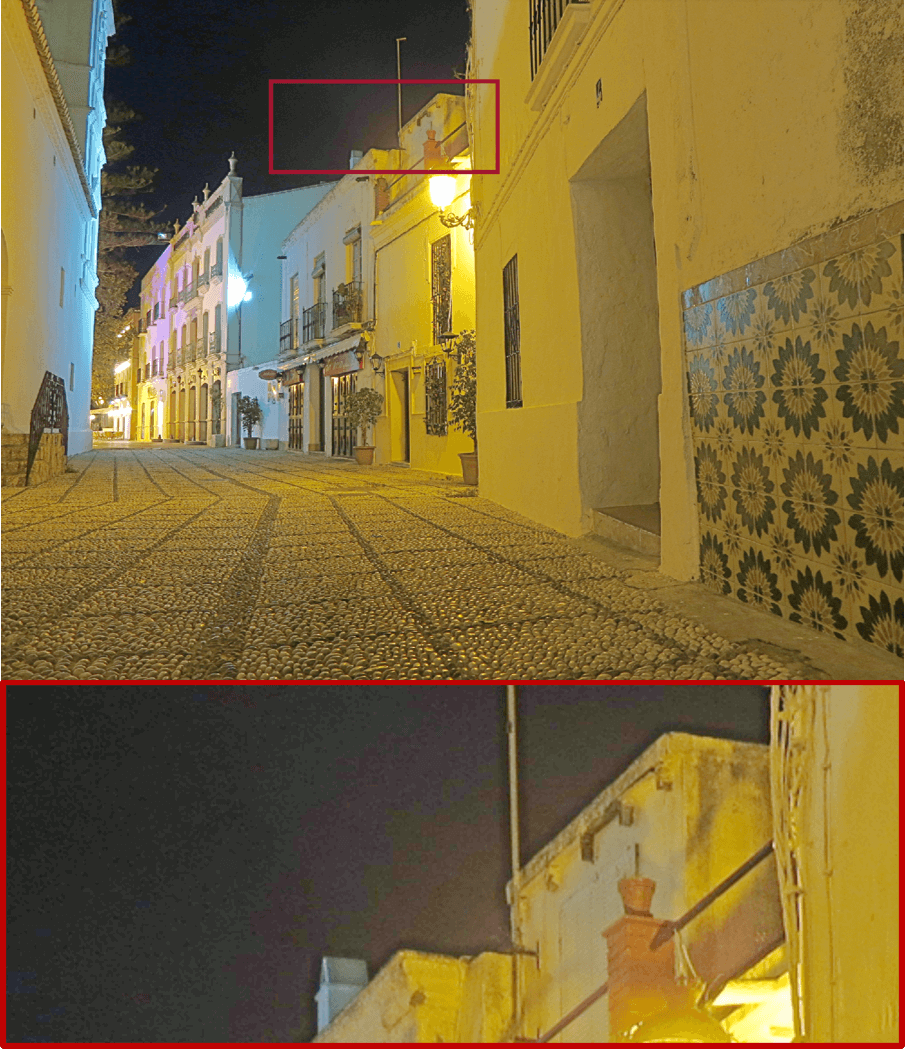}~&
			\includegraphics[width=.18\textwidth]{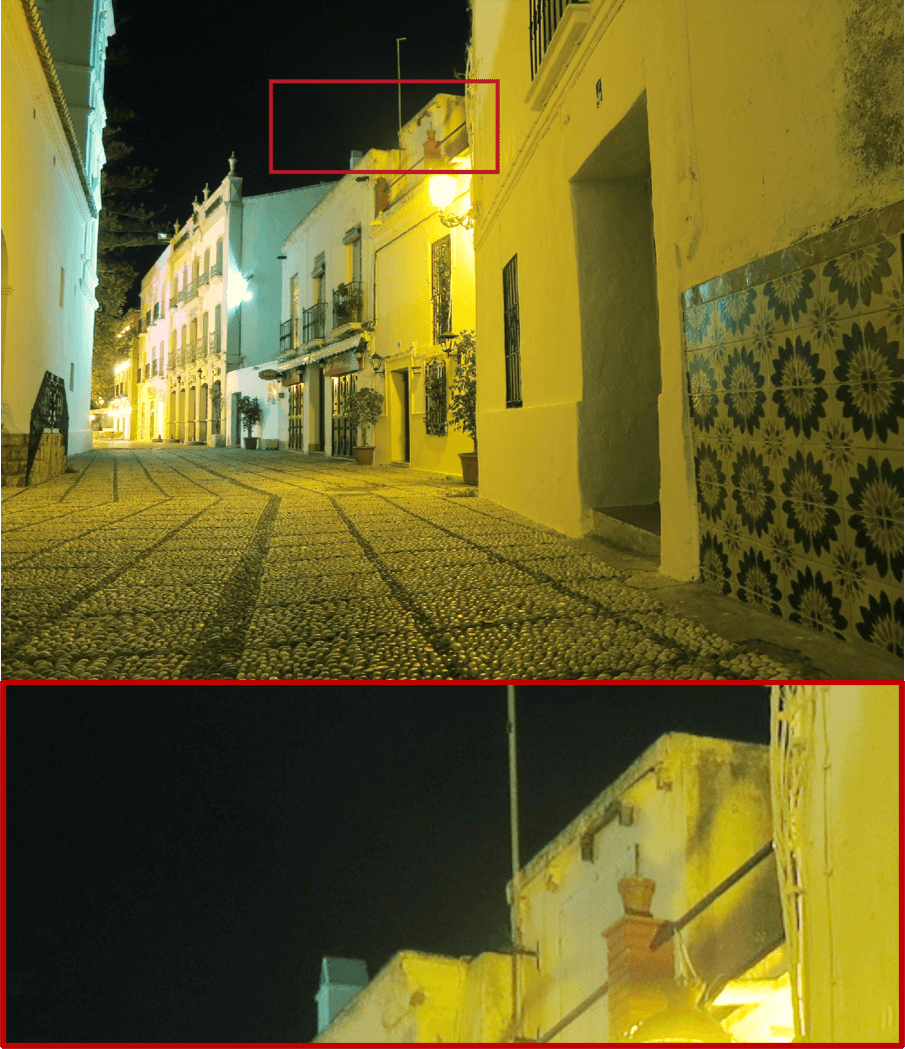}\\
			(a) Input~& (b) LIME~& (c) EnlightenGAN~& (d) Zero-DCE~& (e) Ours\\
		\end{tabular}
	\end{center}
    \vspace{-1em}
	\caption{Examples of enhancement results on LIME evaluation dataset. We show the estimated results of (b) LIME~\protect\cite{guo2016lime}, (c) EnlightenGAN~\protect\cite{jiang2019enlightengan}, (d) Zero-DCE~\protect\cite{guo2020zero}, and (e) Our ReLLIE. 
	%Zoom in to see the details.
	}
	\label{fig:lime}
\end{figure*}

\section{Experiments and Results}

In this section, we show how the proposed method ReLLIE achieves a more customized LLIE in real-world scenarios and thereby boosts the LLIE performance. 

\subsection{Experiments Setting}
%but also conduct qualitative and quantitative comparisons.\par

\noindent
\textbf{Datasets and baselines.} We conduct experiments on two types of LLIE datasets, the standard dataset with paired data (LOL dataset~\cite{wei2018deep}), and the datasets without ground truth images (LIME~ \cite{guo2016lime}, NPE~\cite{wang2013naturalness}, and DICM~\cite{lee2012contrast}). We compare our methods against several state-of-the-art LLIE baselines. The baselines can be classified into three categories, the supervised methods (Retinex-Net~\cite{wei2018deep} and KinD~\cite{zhang2019kindling}), the unsupervised methods (EnlightenGAN~\cite{jiang2019enlightengan} and Zero-DCE~\cite{guo2020zero}), and the zero-shot methods (LIME~\cite{guo2016lime} and Kar \etal \cite{Kar_2021_CVPR}). All the baselines are implemented using the publicly available codes as well as recommended parameters.

We note that the definition of ``zero-shot'' in this paper is different from the conventional ``zero-shot learning'' which often refers to using the learned models to handle images of unseen categories. In this paper, the ``zero-shot'' setting indicates the model can only observe \emph{a single image} in training process. This setting is very challenging, since most of the learning-based models have much more parameters which require sufficient data in training. 
%All the comparison in terms of subjective visual quality and quantitative measurements are reported in the following subsections.

\noindent\textbf{Implementation details.}
We implement the proposed method using PyTorch framework~\cite{paszke2017automatic}. We implement two versions of ReLLIE for both unsupervised and zero-shot settings. For unsupervised learning with sufficient training data, we adopt a seven-layer neural network as the policy agent. For zero-shot learning, we adopt a four-layer neural network as the policy agent. Except the number of layers, all the hyperparameters are identical for both of them. The coefficients in the loss is set as $W_{spa} = 1$, $W_{exp} = 100$, $W_{crl} = 20$, and $W_{tvA} = 200$. In CDMU, $\omega_{CD} = 0.2$ is set as the default. For agent learning, the discount factor $\gamma$ is 0.95, the learning rate is 0.001, and the number of training iterations is 20,000 and 1,000 for unsupervised and zero-shot setting, respectively. All the experiments are conducted on a GTX 1080Ti GPU. 

\begin{table}[t]
\caption{Quantitative results on LOL dataset~\cite{wei2018deep}. +FFDNet denotes employing an external FFDNet~\protect\cite{zhang2018ffdnet} denoiser for post-processing the enhanced results.
% The best result of each category is in \textbf{bold}.
}
\resizebox{1\hsize}{!}{
\begin{tabular}{@{}c|c|ccc@{}}
\toprule
\multicolumn{2}{c|}{\textbf{Methods}}                      & \multicolumn{3}{c}{\textbf{Metrics}}                       \\
\multicolumn{2}{c|}{}                             & LPIPS $\downarrow$           & SSIM $\uparrow$            & PSNR $\uparrow$           \\ \midrule 
\multirow{2}{*}{Supervised} & Retinex-Net~\cite{wei2018deep}         & 0.4739          & 0.5336          & 16.77          \\
                            & KinD~\cite{zhang2019kindling}                & \textbf{0.1593} & \textbf{0.8784} & \textbf{20.38} \\ \midrule \midrule
                            & EnlightenGAN~\cite{guo2020zero}        & 0.3661          & 0.6601          & 17.02          \\
                            & EnlightenGAN+FFDNet & 0.2219          & 0.8130          & 17.63          \\
Unsupervised                & Zero-DCE~\cite{guo2020zero}            & 0.3352          & 0.6632          & 14.86          \\
                            & Zero-DCE+FFDNet     & 0.2179          & 0.7674          & 15.03          \\
                            & ReLLIE+FFDNet       & \textbf{0.1974} & \textbf{0.8268} & \textbf{19.52} \\ \midrule
                            & LIME~\cite{guo2016lime}                & 0.3724          & 0.6216          & 14.02          \\
                            & LIME+FFDNet                & 0.2819          & 0.7419          & 14.20          \\
Zero-shot                   & {Kar~\etal  \cite{Kar_2021_CVPR} \protect\footnotemark}               & -               & 0.6950          & 17.50          \\
                            & ReLLIE               & 0.3976               & 0.6413          & 18.37          \\
                            & ReLLIE+FFDNet (ZS)  & \textbf{0.2618} & \textbf{0.7733} & \textbf{18.99} \\ \bottomrule
\end{tabular}
}
\label{tab:efficiency}
\end{table}
\footnotetext{Since the authors \cite{Kar_2021_CVPR} have not released the code, we only report the SSIM and PSNR referring to their paper.}
\subsection{Quantitative Comparison}
For quantitative comparison with existing methods, we employ three metrics including Peak Signal-to-Noise Ratio (PSNR, dB), Structural Similarity (SSIM), and Learned Perceptual Image Patch Similarity (LPIPS) \cite{zhang2018perceptual}. Table~\ref{tab:efficiency} summarizes the performances of ReLLIE and baselines on  the test images of LOL dataset. Guided by the paired data (\ie, supervised learning), KinD~\cite{zhang2019kindling} achieves the best performance. Except KinD~\cite{zhang2019kindling}, our ReLLIE outperforms all the other baselines under both unsupervised and zero-shot settings. It demonstrates the efficacy of DRL for LLIE tasks.
%It should be noticed that for zero-shot learning, the PSNR is slightly better than which of the supervised version of our proposed RELE. This is achieved by slightly tuning different $N$ for different images during the testing, and the $N$ is selected differently for different images by the user.

In Fig.~\ref{fig:lime}, we show the results of zero-shot LLIE. The upper row shows that ReLLIE preserves more contextual information with a better contrast. The lower row shows that ReLLIE avoids artifacts which exist in all the other baselines. More details are zoomed in with red boxes for further comparison. The results of NPE and DICM can be found in supplementary materials.

%, we present the performance of all the baselines and our proposed RELE
\subsection{Visual Quality Comparison}
Figs.~\ref{fig:lol} and \ref{fig:lime} compare subjective visual quality on low-light images. Fig.~\ref{fig:lol} shows the unsupervised LLIE setting that the ground truth is available. The enhanced images provided by our ReLLIE is more visually pleasing without obvious noise and color casts. Moreover, the results of ReLLIE are more sharp with more details remained and therefore preserving more visual information. It should be noticed that we adopt $N=6$ for all the images, even though $N$ can be changed according to users' preference for better performance (see Fig. \ref{fig:ablation}). Results on the third sample reveal that for some very dark low-light images, $N=6$ may result in under-enhancement. However, ReLLIE can still enhance the image with a relatively good contrast and yields visually pleasure results.

\subsection{Visualization of Customized LLIE}
\begin{figure}[t]
\centering 
\includegraphics[width=1.\linewidth]{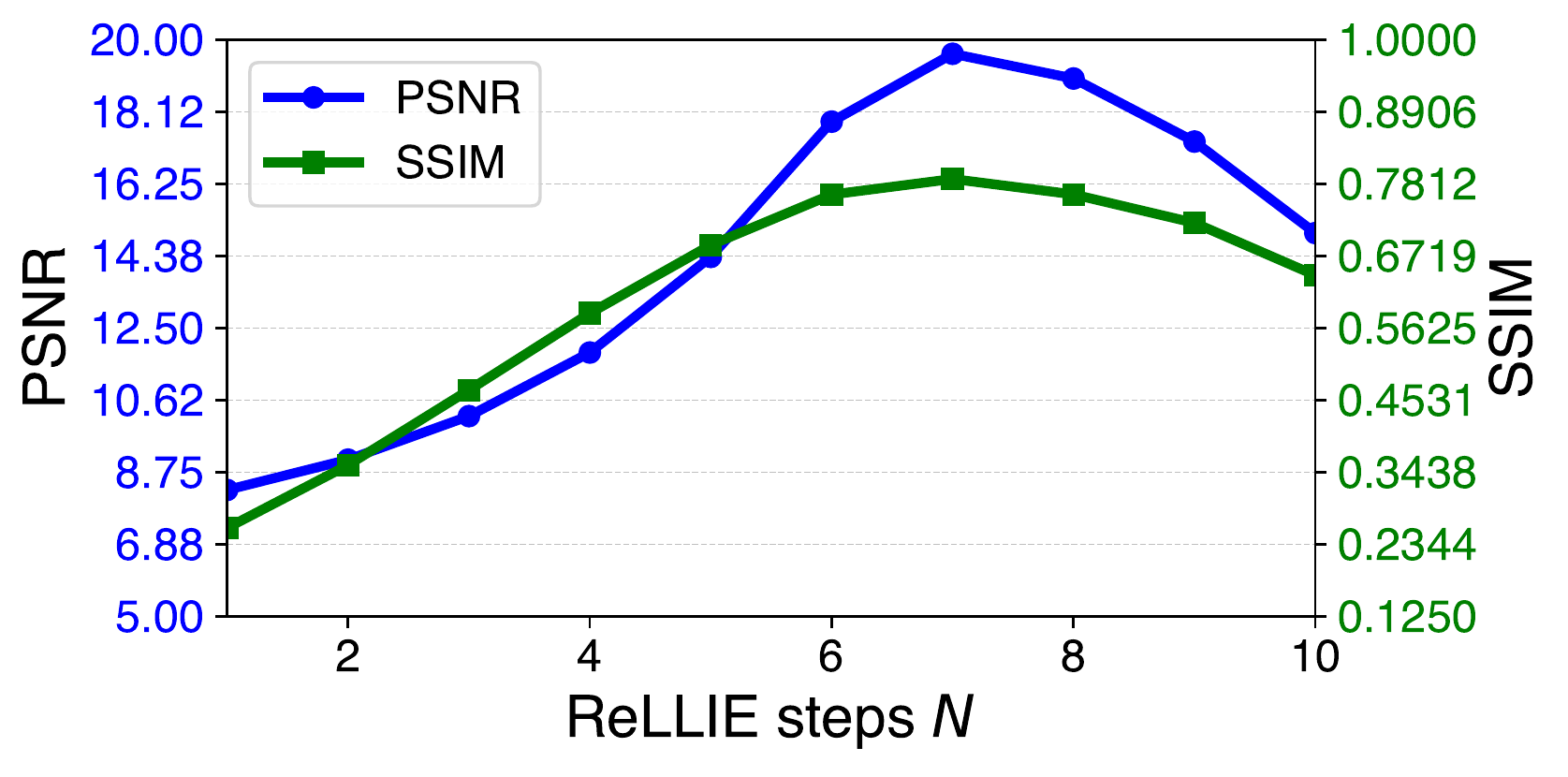}
\caption{Quantitative performance of ReLLIE on LOL evaluation set with different number of enhancement steps $N$.}
%changed for pixels from low-light images to the enhanced one }
\label{fig:p}
\end{figure}

\begin{figure}[t]
\centering 
\begin{minipage}[b]{1.0\linewidth}
\centering 
\includegraphics[width=1\linewidth]{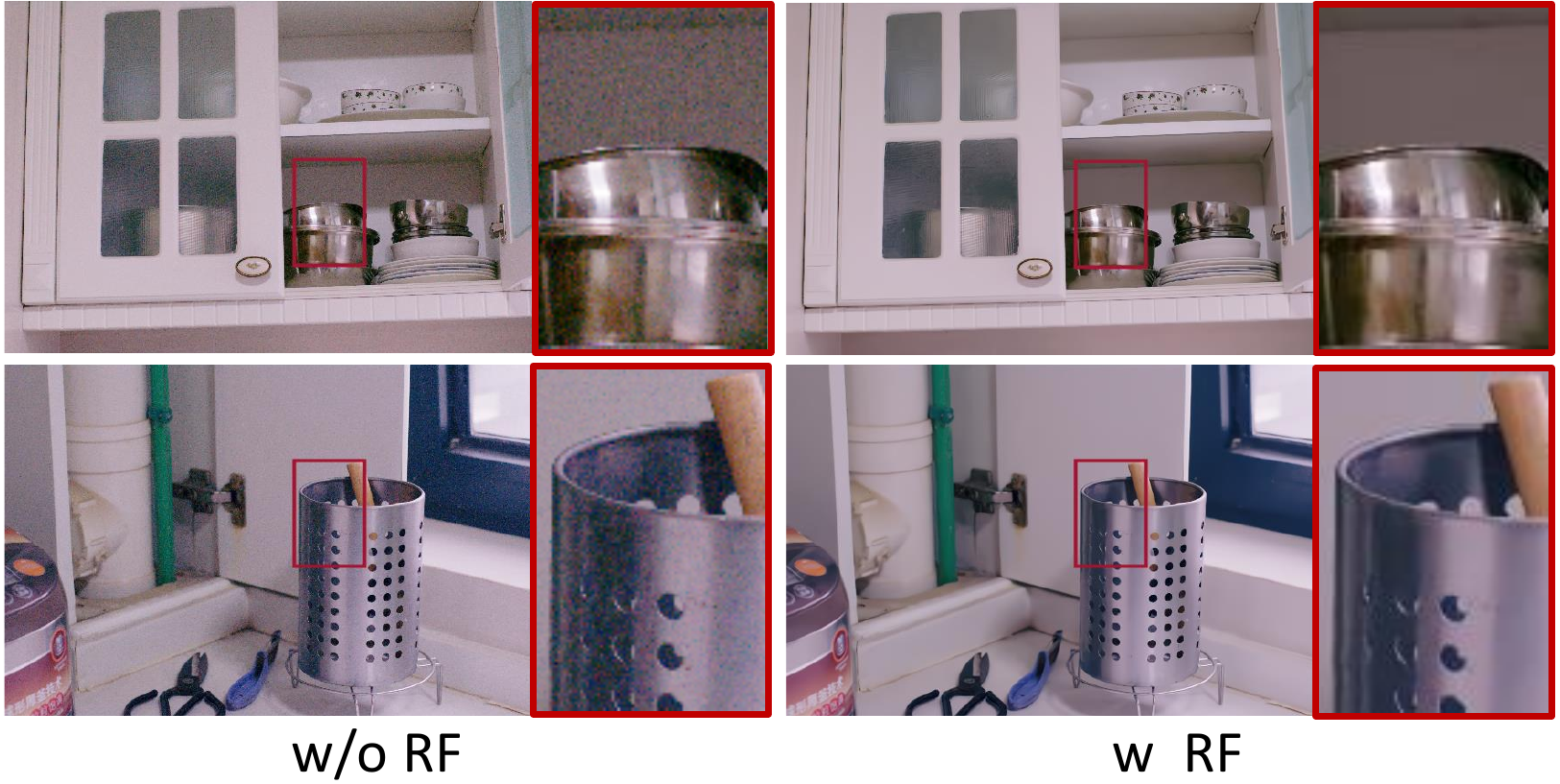}
\caption{Enhancement examples of ReLLIE (zero-shot) without and with RF, respectively.}
\label{fig:ablation2}
\end{minipage}
\end{figure}

\begin{figure*}[!ht]
\centering 
\begin{minipage}[b]{1.0\textwidth}
\centering 
\includegraphics[width=1\textwidth,height=0.25\textwidth]{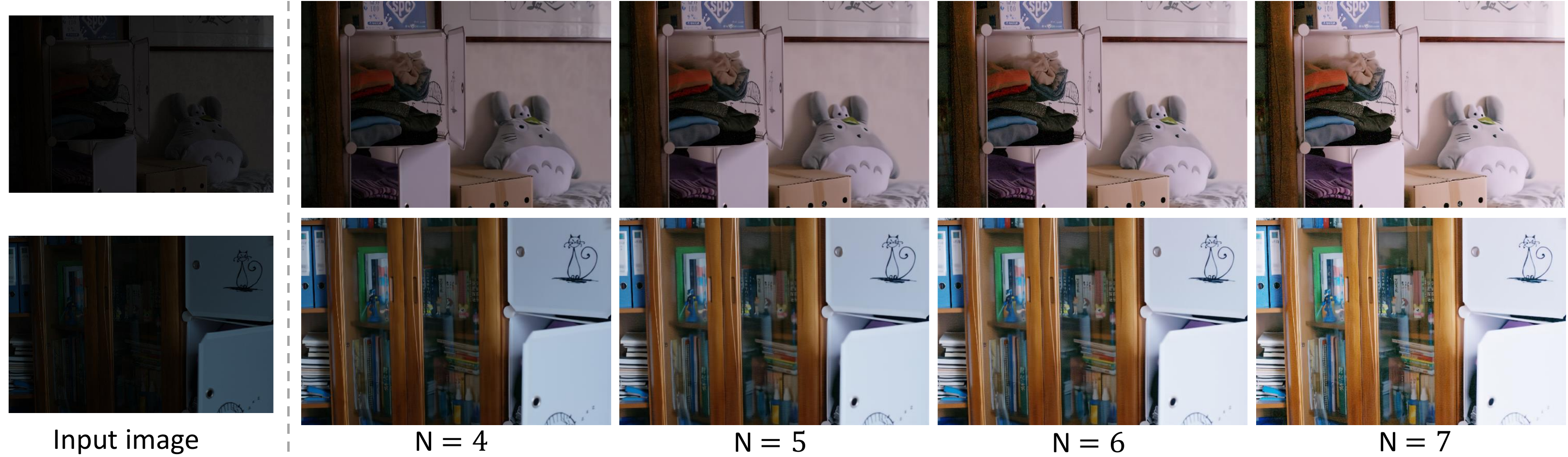}
\caption{Examples of customized LLIE with different enhancement steps. 
%All these images are candidate outputs given by our approach which can be selected by customers according to their preference.
}
\label{fig:ablation}
\end{minipage}
\end{figure*}

The favorite illumination strengths of different persons may be pretty diverse. Therefore, a practical approach needs to consider the user-orientated goals by providing various enhancement options. Fig. \ref{fig:ablation} shows the customized enhanced images provided by the proposed ReLLIE in zero-shot scenarios and Fig. \ref{fig:p} demonstrates the different SSIM and PNSR achieved with different $N$. Given a single low-light image, we train a randomly initialized agent with a fixed amount of steps, \ie,~$N=8$, for 1,000 iterations until it converges. ReLLIE is customized, because: 1) Even though the policy network is trained with $N=8$, in testing phase the images can be enhanced for arbitrary steps; 2) Although no refinement module is involved in training, it can be employed at arbitrary steps to improve the testing performance, as shown in Fig. \ref{fig:ablation2}. In one word, our ReLLIE provides more candidate enhanced images to users. Therefore, it is more customized and suitable for real-world applications.

\begin{figure}[t]
\centering 
\includegraphics[scale=0.18]{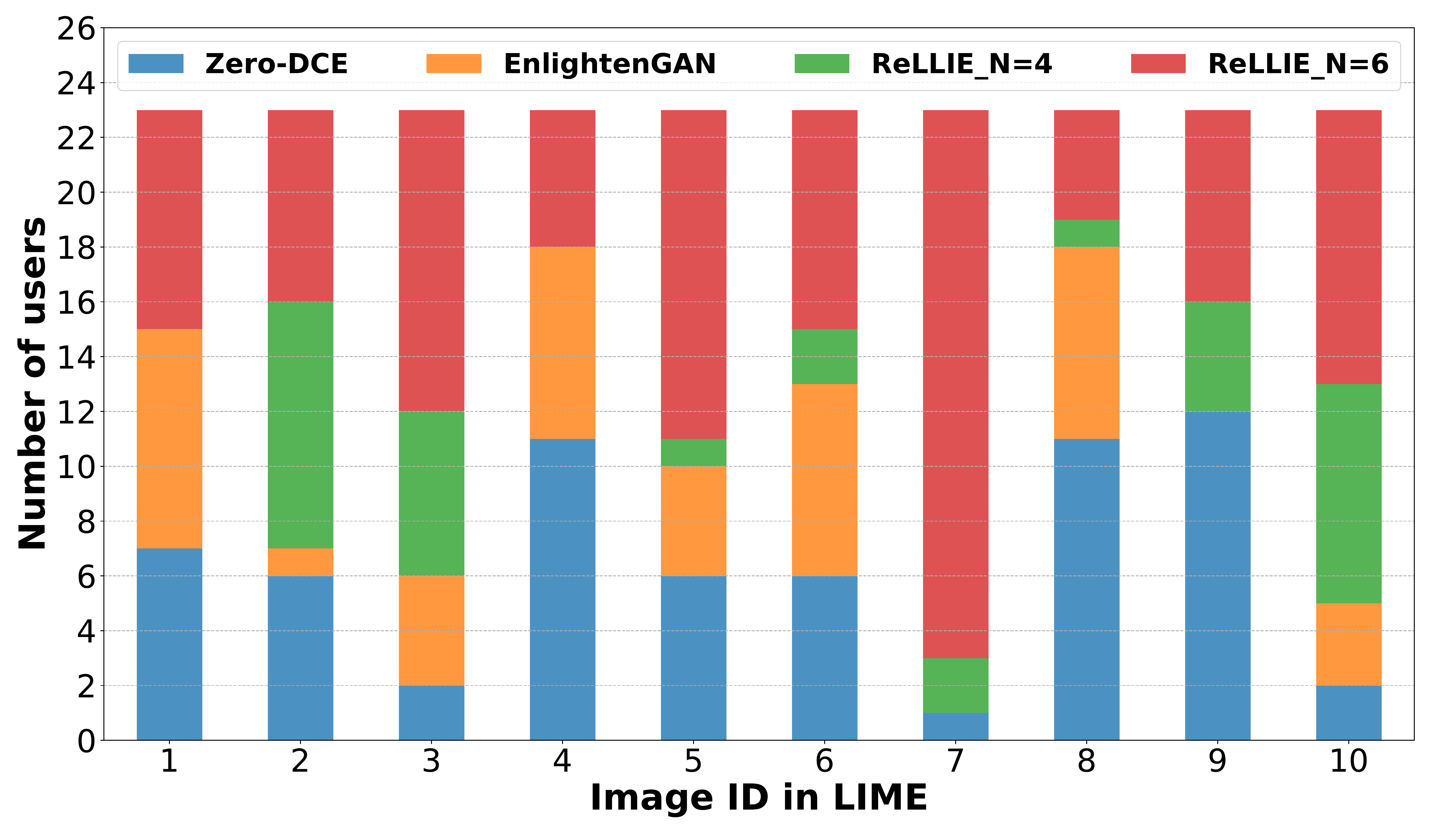}
\caption{User study results on LIME \cite{guo2016lime} evaluation set. }
%changed for pixels from low-light images to the enhanced one }
\label{fig:user}
\end{figure}
%\subsection{Comparison with State-of-the-art}
%We evaluate our RELE model on widely-adopted datasets, including LOL~\cite{wei2018deep}, LIME~\cite{guo2016lime}, NPE~\cite{}, and MEF~\cite{}. 
%We compare our RELE with several state-of-the-art methods: one conventional method (LIME~\cite{guo2016lime}), three reference required methods (Retinex-Net~\cite{wei2018deep}, EnlightenGAN~\cite{jiang2019enlightengan}, KinD~\cite{zhang2019kindling}), and one non-reference method (Zero-DCE~\cite{guo2020zero}). The results are reproduced by using publicly released source codes with recommended parameters.\par

\subsection{User Study}
To investigate the subjective assessment of LLIE approaches, we further conduct a user study on the LIME testing dataset with 10 images. A total of 23 users, who cover various ages, genders, and occupations, are invited to select their favorite images from the enhancement results provided by Zero-DCE \cite{guo2020zero}, EnglightenGAN \cite{jiang2019enlightengan}, and our ReLLIE (with enhancement steps $N=4$ and $N=6$) on their own devices. Fig. \ref{fig:user} summarizes the user study results, and 
it demonstrates that ReLLIE can better meet users' preference for most of the images.

\subsection{Ablation Study}
\begin{figure}[!t]
	\begin{center}
		\begin{tabular}{c@{ }c@{ }c}
			\includegraphics[width=.3\linewidth,height=.3\linewidth]{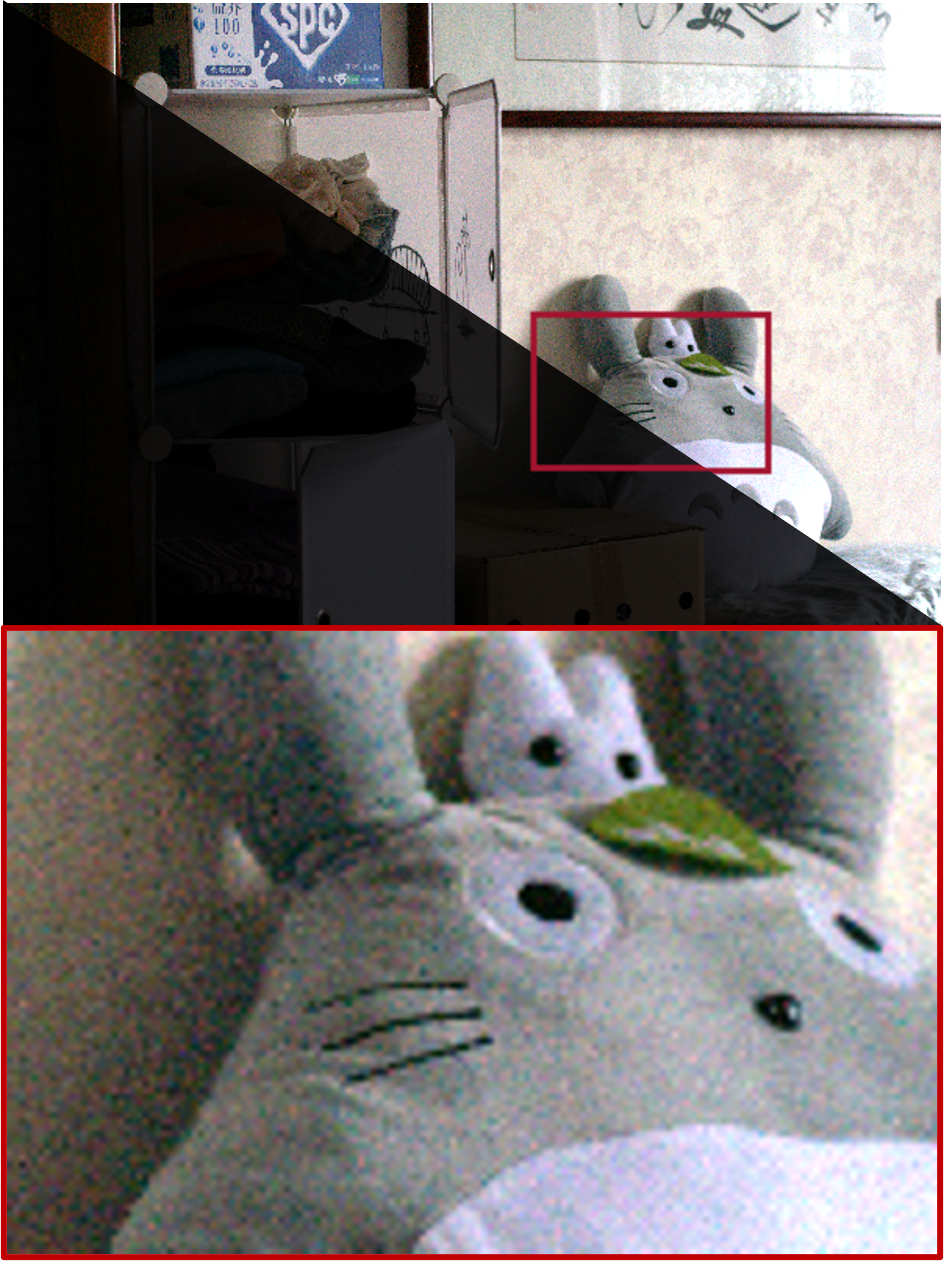}~&
				\includegraphics[width=.3\linewidth,height=.3\linewidth]{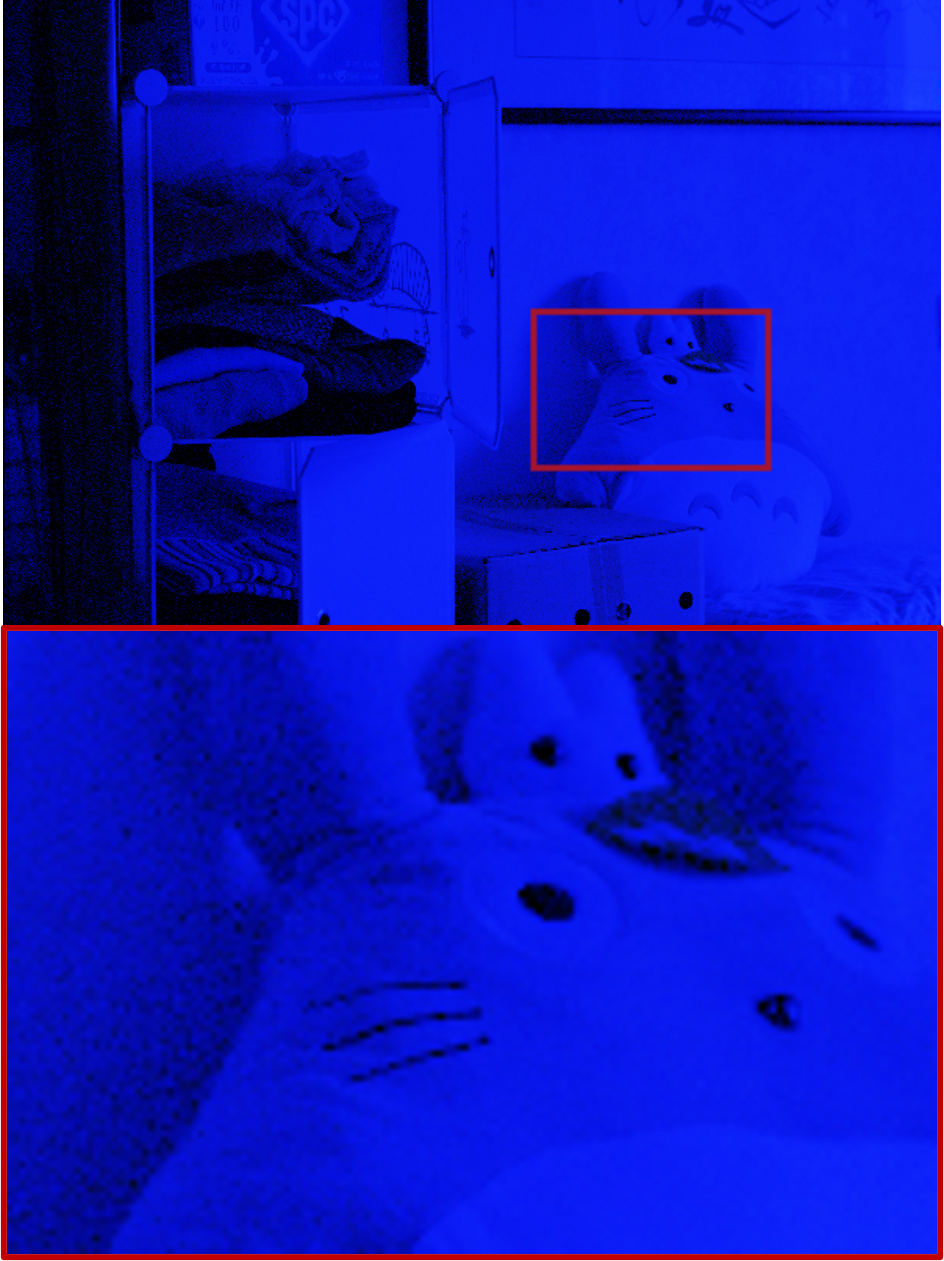}~&

			\includegraphics[width=.3\linewidth,height=.3\linewidth]{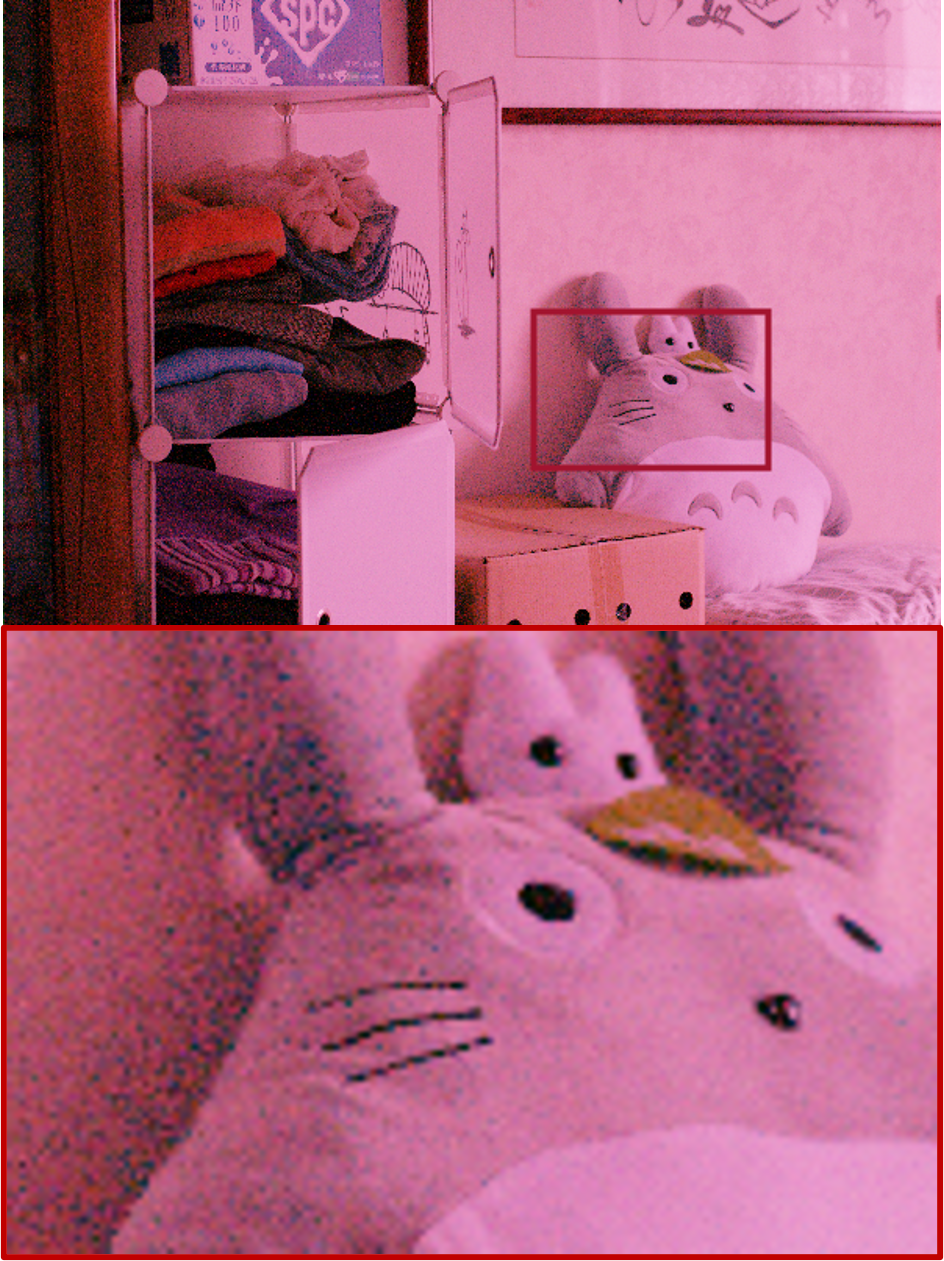}\\	
				(a) Input~& (b) Baseline ~&(c)  +CRL \\\includegraphics[width=.3\linewidth,height=.3\linewidth]{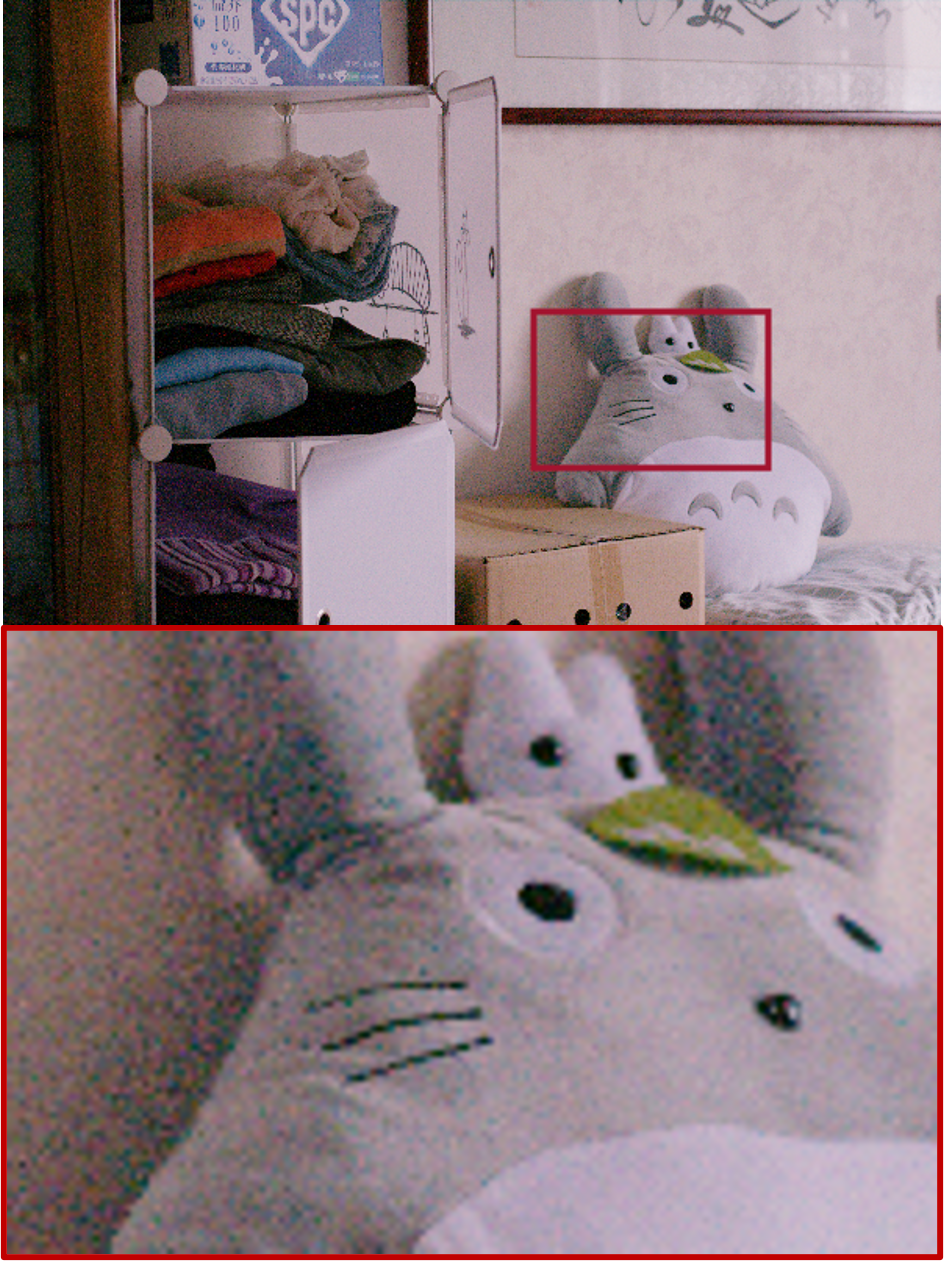}~&
				\includegraphics[width=.3\linewidth,height=.3\linewidth]{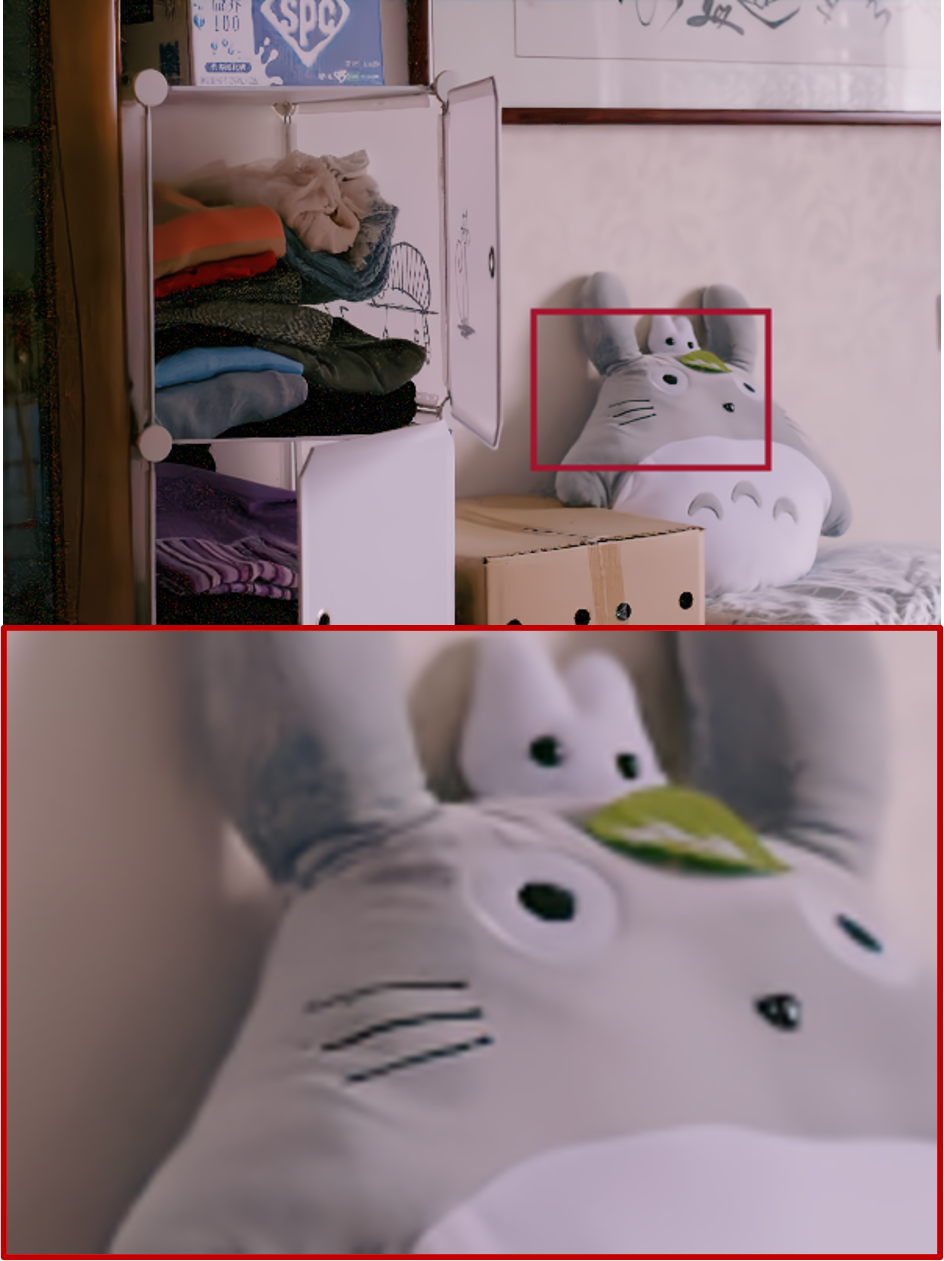}~&

			\includegraphics[width=.3\linewidth,height=.3\linewidth]{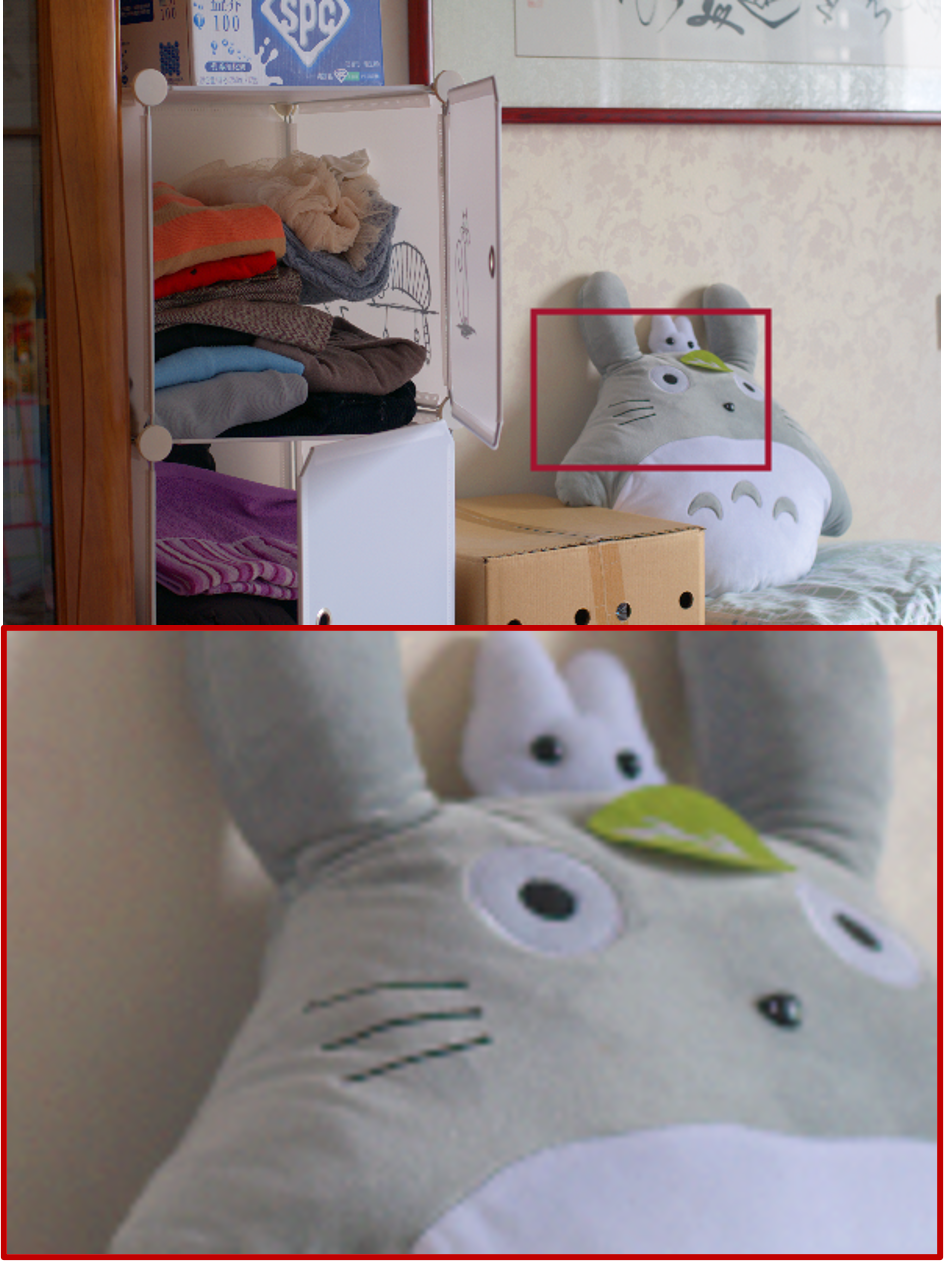}\\	
			(d)+CRL, CDMU~& (e) +CRL, CDMU, RF~& (f) Ground truth\\
		\end{tabular}
	\end{center}
	\caption{The effect of components in ReLLIE. (b) is the baseline without using all the components.}
	\label{fig:compare}
\end{figure}

To study the effectiveness of proposed components in ReLLIE, including CRL, CDMU, and RF, we further perform ablation studies on unsupervised LLIE and summarize the results in Table~\ref{tab:adaptation}. We observe that by adding the components progressively, the model performance is significantly improved from 7.76 dB to 19.52 dB in PSNR. More specifically, compared with the baseline without all the components, CRL can alleviate the color casts and accompanied with CDMU this issue can be handled well. It can also be observed that RF can boost the visual quality by removing the noise.  Fig. \ref{fig:compare} shows a qualitative example to reveal how each component influences the outputs. 

\begin{table}[t]
    \centering
    % \footnotesize
    % \setlength{\tabcolsep}{0.4em}
    % \renewcommand\tabcolsep{2pt}
       \caption{Ablation study on the components of ReLLIE. 
       %CRL, CDMU and RF stand for channel-ratio constancy loss, channel dependent momentum update, and refinement respectively. \checkmark means trained with this components. 
       }
       \vspace{-1em}
    % \adjustbox{width=\linewidth}{
     \resizebox{0.8\hsize}{!}{
\begin{tabular}{@{}ccc|ccc@{}}
\toprule
\multirow{2}{*}{CRL} & \multirow{2}{*}{CDMU} & \multirow{2}{*}{RF} & \multicolumn{3}{c}{ReLLIE}                         \\
                     &                       &                     &  LPIPS $\downarrow$           & SSIM $\uparrow$            & PSNR $\uparrow$           \\ \midrule
                     &                       &                     & 0.6746          & 0.3798          & 7.76           \\
     \checkmark      &                &                     & 0.4824          & 0.6653          & 15.40          \\
     \checkmark      & \checkmark            &                     & 0.3450          & 0.6730          & 18.74          \\
     \checkmark      & \checkmark            & \checkmark             & \textbf{0.1974} & \textbf{0.8268} & \textbf{19.52} \\ \bottomrule
\end{tabular}
    }
    % }

    \label{tab:adaptation}
    % \vspace{-0.4cm}
\end{table}

\section{Conclusion}
In this paper, we have proposed a non-reference DRL based framework, ReLLIE, for efficient, robust, and customized low-light image enhancement. By learning a stochastic image translation policy instead of a one-one translation model, ReLLIE provides diverse image enhancement candidates to meet different individuals' preference. In addition, we have proposed a series of learning modules including CRL, CDMU and RF to enhance the robustness of LLIE methods. Extensive qualitative and quantitative experiments and user study have validated the superiority of ReLLIE against existing methods on unsupervised/zero-shot LLIE scenarios. 
%In future work, we will try to consider overexposure scenarios and  other real-life degeneration.



\newpage

\bibliographystyle{ACM-Reference-Format}
\bibliography{bibfile}


\begin{thebibliography}{30}

%%% the Web address.

\ifx \showCODEN    \undefined \def \showCODEN     #1{\unskip}     \fi
\ifx \showDOI      \undefined \def \showDOI       #1{#1}\fi
\ifx \showISBNx    \undefined \def \showISBNx     #1{\unskip}     \fi
\ifx \showISBNxiii \undefined \def \showISBNxiii  #1{\unskip}     \fi
\ifx \showISSN     \undefined \def \showISSN      #1{\unskip}     \fi
\ifx \showLCCN     \undefined \def \showLCCN      #1{\unskip}     \fi
\ifx \shownote     \undefined \def \shownote      #1{#1}          \fi
\ifx \showarticletitle \undefined \def \showarticletitle #1{#1}   \fi
\ifx \showURL      \undefined \def \showURL       {\relax}        \fi
% invisible to TeX
\providecommand\bibfield[2]{#2}
\providecommand\bibinfo[2]{#2}
\providecommand\natexlab[1]{#1}
\providecommand\showeprint[2][]{arXiv:#2}

\bibitem[\protect\citeauthoryear{Abdullah-Al-Wadud, Kabir, Dewan, and
  Chae}{Abdullah-Al-Wadud et~al\mbox{.}}{2007}]%
        {abdullah2007dynamic}
\bibfield{author}{\bibinfo{person}{Mohammad Abdullah-Al-Wadud},
  \bibinfo{person}{Md~Hasanul Kabir}, \bibinfo{person}{M~Ali~Akber Dewan},
  {and} \bibinfo{person}{Oksam Chae}.} \bibinfo{year}{2007}\natexlab{}.
\newblock \showarticletitle{A dynamic histogram equalization for image contrast
  enhancement}.
\newblock \bibinfo{journal}{\emph{IEEE Transactions on Consumer Electronics}}
  \bibinfo{volume}{53}, \bibinfo{number}{2} (\bibinfo{year}{2007}),
  \bibinfo{pages}{593--600}.
\newblock


\bibitem[\protect\citeauthoryear{Fu, Zeng, Huang, Liao, Ding, and Paisley}{Fu
  et~al\mbox{.}}{2016}]%
        {fu2016fusion}
\bibfield{author}{\bibinfo{person}{Xueyang Fu}, \bibinfo{person}{Delu Zeng},
  \bibinfo{person}{Yue Huang}, \bibinfo{person}{Yinghao Liao},
  \bibinfo{person}{Xinghao Ding}, {and} \bibinfo{person}{John Paisley}.}
  \bibinfo{year}{2016}\natexlab{}.
\newblock \showarticletitle{A fusion-based enhancing method for weakly
  illuminated images}.
\newblock \bibinfo{journal}{\emph{Signal Processing}}  \bibinfo{volume}{129}
  (\bibinfo{year}{2016}), \bibinfo{pages}{82--96}.
\newblock


\bibitem[\protect\citeauthoryear{Furuta, Inoue, and Yamasaki}{Furuta
  et~al\mbox{.}}{2019}]%
        {furuta2019fully}
\bibfield{author}{\bibinfo{person}{Ryosuke Furuta}, \bibinfo{person}{Naoto
  Inoue}, {and} \bibinfo{person}{Toshihiko Yamasaki}.}
  \bibinfo{year}{2019}\natexlab{}.
\newblock \showarticletitle{Fully convolutional network with multi-step
  reinforcement learning for image processing}. In
  \bibinfo{booktitle}{\emph{Proceedings of the AAAI Conference on Artificial
  Intelligence}}. \bibinfo{pages}{3598--3605}.
\newblock


\bibitem[\protect\citeauthoryear{Gharbi, Chen, Barron, Hasinoff, and
  Durand}{Gharbi et~al\mbox{.}}{2017}]%
        {gharbi2017deep}
\bibfield{author}{\bibinfo{person}{Micha{\"e}l Gharbi}, \bibinfo{person}{Jiawen
  Chen}, \bibinfo{person}{Jonathan~T Barron}, \bibinfo{person}{Samuel~W
  Hasinoff}, {and} \bibinfo{person}{Fr{\'e}do Durand}.}
  \bibinfo{year}{2017}\natexlab{}.
\newblock \showarticletitle{Deep bilateral learning for real-time image
  enhancement}.
\newblock \bibinfo{journal}{\emph{ACM Transactions on Graphics (TOG)}}
  \bibinfo{volume}{36}, \bibinfo{number}{4} (\bibinfo{year}{2017}),
  \bibinfo{pages}{118}.
\newblock


\bibitem[\protect\citeauthoryear{Guo, Li, Guo, Loy, Hou, Kwong, and Cong}{Guo
  et~al\mbox{.}}{2020}]%
        {guo2020zero}
\bibfield{author}{\bibinfo{person}{Chunle Guo}, \bibinfo{person}{Chongyi Li},
  \bibinfo{person}{Jichang Guo}, \bibinfo{person}{Chen~Change Loy},
  \bibinfo{person}{Junhui Hou}, \bibinfo{person}{Sam Kwong}, {and}
  \bibinfo{person}{Runmin Cong}.} \bibinfo{year}{2020}\natexlab{}.
\newblock \showarticletitle{Zero-Reference Deep Curve Estimation for Low-Light
  Image Enhancement}. In \bibinfo{booktitle}{\emph{Proceedings of the IEEE/CVF
  Conference on Computer Vision and Pattern Recognition}}.
  \bibinfo{pages}{1780--1789}.
\newblock


\bibitem[\protect\citeauthoryear{Guo, Li, and Ling}{Guo et~al\mbox{.}}{2016}]%
        {guo2016lime}
\bibfield{author}{\bibinfo{person}{Xiaojie Guo}, \bibinfo{person}{Yu Li}, {and}
  \bibinfo{person}{Haibin Ling}.} \bibinfo{year}{2016}\natexlab{}.
\newblock \showarticletitle{LIME: Low-light image enhancement via illumination
  map estimation}.
\newblock \bibinfo{journal}{\emph{IEEE Transactions on image processing}}
  \bibinfo{volume}{26}, \bibinfo{number}{2} (\bibinfo{year}{2016}),
  \bibinfo{pages}{982--993}.
\newblock


\bibitem[\protect\citeauthoryear{Hu, He, Xu, Wang, and Lin}{Hu
  et~al\mbox{.}}{2018}]%
        {hu2018exposure}
\bibfield{author}{\bibinfo{person}{Yuanming Hu}, \bibinfo{person}{Hao He},
  \bibinfo{person}{Chenxi Xu}, \bibinfo{person}{Baoyuan Wang}, {and}
  \bibinfo{person}{Stephen Lin}.} \bibinfo{year}{2018}\natexlab{}.
\newblock \showarticletitle{Exposure: A white-box photo post-processing
  framework}.
\newblock \bibinfo{journal}{\emph{ACM Transactions on Graphics (TOG)}}
  \bibinfo{volume}{37}, \bibinfo{number}{2} (\bibinfo{year}{2018}),
  \bibinfo{pages}{1--17}.
\newblock


\bibitem[\protect\citeauthoryear{Jiang, Gong, Liu, Cheng, Fang, Shen, Yang,
  Zhou, and Wang}{Jiang et~al\mbox{.}}{2019}]%
        {jiang2019enlightengan}
\bibfield{author}{\bibinfo{person}{Yifan Jiang}, \bibinfo{person}{Xinyu Gong},
  \bibinfo{person}{Ding Liu}, \bibinfo{person}{Yu Cheng}, \bibinfo{person}{Chen
  Fang}, \bibinfo{person}{Xiaohui Shen}, \bibinfo{person}{Jianchao Yang},
  \bibinfo{person}{Pan Zhou}, {and} \bibinfo{person}{Zhangyang Wang}.}
  \bibinfo{year}{2019}\natexlab{}.
\newblock \showarticletitle{Enlightengan: Deep light enhancement without paired
  supervision}.
\newblock \bibinfo{journal}{\emph{arXiv preprint arXiv:1906.06972}}
  (\bibinfo{year}{2019}).
\newblock


\bibitem[\protect\citeauthoryear{Kar, Kanti~Dhara, Sen, and Kumar~Biswas}{Kar
  et~al\mbox{.}}{2021}]%
        {Kar_2021_CVPR}
\bibfield{author}{\bibinfo{person}{Aupendu Kar}, \bibinfo{person}{Sobhan
  Kanti~Dhara}, \bibinfo{person}{Debashis Sen}, {and} \bibinfo{person}{Prabir
  Kumar~Biswas}.} \bibinfo{year}{2021}\natexlab{}.
\newblock \showarticletitle{Zero-shot Single Image Restoration through
  Controlled Perturbation of Koschmieder’s Model}. In
  \bibinfo{booktitle}{\emph{IEEE/CVF Conference on Computer Vision and Pattern
  Recognition (CVPR)}}.
\newblock


\bibitem[\protect\citeauthoryear{Land}{Land}{1977}]%
        {land1977retinex}
\bibfield{author}{\bibinfo{person}{Edwin~H Land}.}
  \bibinfo{year}{1977}\natexlab{}.
\newblock \showarticletitle{The retinex theory of color vision}.
\newblock \bibinfo{journal}{\emph{Scientific american}} \bibinfo{volume}{237},
  \bibinfo{number}{6} (\bibinfo{year}{1977}), \bibinfo{pages}{108--129}.
\newblock


\bibitem[\protect\citeauthoryear{Lee, Lee, and Kim}{Lee et~al\mbox{.}}{2012}]%
        {lee2012contrast}
\bibfield{author}{\bibinfo{person}{Chulwoo Lee}, \bibinfo{person}{Chul Lee},
  {and} \bibinfo{person}{Chang-Su Kim}.} \bibinfo{year}{2012}\natexlab{}.
\newblock \showarticletitle{Contrast enhancement based on layered difference
  representation}. In \bibinfo{booktitle}{\emph{2012 19th IEEE International
  Conference on Image Processing}}. IEEE, \bibinfo{pages}{965--968}.
\newblock


\bibitem[\protect\citeauthoryear{Lee, Lee, and Kim}{Lee et~al\mbox{.}}{2013}]%
        {lee2013contrast}
\bibfield{author}{\bibinfo{person}{Chulwoo Lee}, \bibinfo{person}{Chul Lee},
  {and} \bibinfo{person}{Chang-Su Kim}.} \bibinfo{year}{2013}\natexlab{}.
\newblock \showarticletitle{Contrast enhancement based on layered difference
  representation of 2D histograms}.
\newblock \bibinfo{journal}{\emph{IEEE transactions on image processing}}
  \bibinfo{volume}{22}, \bibinfo{number}{12} (\bibinfo{year}{2013}),
  \bibinfo{pages}{5372--5384}.
\newblock


\bibitem[\protect\citeauthoryear{Lore, Akintayo, and Sarkar}{Lore
  et~al\mbox{.}}{2017}]%
        {lore2017llnet}
\bibfield{author}{\bibinfo{person}{Kin~Gwn Lore}, \bibinfo{person}{Adedotun
  Akintayo}, {and} \bibinfo{person}{Soumik Sarkar}.}
  \bibinfo{year}{2017}\natexlab{}.
\newblock \showarticletitle{LLNet: A deep autoencoder approach to natural
  low-light image enhancement}.
\newblock \bibinfo{journal}{\emph{Pattern Recognition}}  \bibinfo{volume}{61}
  (\bibinfo{year}{2017}), \bibinfo{pages}{650--662}.
\newblock


\bibitem[\protect\citeauthoryear{Mertens, Kautz, and Van~Reeth}{Mertens
  et~al\mbox{.}}{2009}]%
        {mertens2009exposure}
\bibfield{author}{\bibinfo{person}{Tom Mertens}, \bibinfo{person}{Jan Kautz},
  {and} \bibinfo{person}{Frank Van~Reeth}.} \bibinfo{year}{2009}\natexlab{}.
\newblock \showarticletitle{Exposure fusion: A simple and practical alternative
  to high dynamic range photography}. In \bibinfo{booktitle}{\emph{Computer
  graphics forum}}, Vol.~\bibinfo{volume}{28}. Wiley Online Library,
  \bibinfo{pages}{161--171}.
\newblock


\bibitem[\protect\citeauthoryear{Mnih, Badia, Mirza, Graves, Lillicrap, Harley,
  Silver, and Kavukcuoglu}{Mnih et~al\mbox{.}}{2016}]%
        {mnih2016asynchronous}
\bibfield{author}{\bibinfo{person}{Volodymyr Mnih},
  \bibinfo{person}{Adria~Puigdomenech Badia}, \bibinfo{person}{Mehdi Mirza},
  \bibinfo{person}{Alex Graves}, \bibinfo{person}{Timothy Lillicrap},
  \bibinfo{person}{Tim Harley}, \bibinfo{person}{David Silver}, {and}
  \bibinfo{person}{Koray Kavukcuoglu}.} \bibinfo{year}{2016}\natexlab{}.
\newblock \showarticletitle{Asynchronous methods for deep reinforcement
  learning}. In \bibinfo{booktitle}{\emph{International conference on machine
  learning}}. \bibinfo{pages}{1928--1937}.
\newblock


\bibitem[\protect\citeauthoryear{Park, Lee, Yoo, and Kweon}{Park
  et~al\mbox{.}}{2018}]%
        {park2018distort}
\bibfield{author}{\bibinfo{person}{Jongchan Park}, \bibinfo{person}{Joon-Young
  Lee}, \bibinfo{person}{Donggeun Yoo}, {and} \bibinfo{person}{In~So Kweon}.}
  \bibinfo{year}{2018}\natexlab{}.
\newblock \showarticletitle{Distort-and-recover: Color enhancement using deep
  reinforcement learning}. In \bibinfo{booktitle}{\emph{Proceedings of the IEEE
  conference on computer vision and pattern recognition}}.
  \bibinfo{pages}{5928--5936}.
\newblock


\bibitem[\protect\citeauthoryear{Paszke, Gross, Chintala, Chanan, Yang, DeVito,
  Lin, Desmaison, Antiga, and Lerer}{Paszke et~al\mbox{.}}{2017}]%
        {paszke2017automatic}
\bibfield{author}{\bibinfo{person}{Adam Paszke}, \bibinfo{person}{Sam Gross},
  \bibinfo{person}{Soumith Chintala}, \bibinfo{person}{Gregory Chanan},
  \bibinfo{person}{Edward Yang}, \bibinfo{person}{Zachary DeVito},
  \bibinfo{person}{Zeming Lin}, \bibinfo{person}{Alban Desmaison},
  \bibinfo{person}{Luca Antiga}, {and} \bibinfo{person}{Adam Lerer}.}
  \bibinfo{year}{2017}\natexlab{}.
\newblock \showarticletitle{Automatic differentiation in pytorch}.
\newblock  (\bibinfo{year}{2017}).
\newblock


\bibitem[\protect\citeauthoryear{Sutton and Barto}{Sutton and Barto}{2018}]%
        {sutton2018reinforcement}
\bibfield{author}{\bibinfo{person}{Richard~S Sutton} {and}
  \bibinfo{person}{Andrew~G Barto}.} \bibinfo{year}{2018}\natexlab{}.
\newblock \bibinfo{booktitle}{\emph{Reinforcement learning: An introduction}}.
\newblock \bibinfo{publisher}{MIT press}.
\newblock


\bibitem[\protect\citeauthoryear{Sutton, McAllester, Singh, and Mansour}{Sutton
  et~al\mbox{.}}{[n.d.]}]%
        {NIPS1999_464d828b}
\bibfield{author}{\bibinfo{person}{Richard~S Sutton}, \bibinfo{person}{David
  McAllester}, \bibinfo{person}{Satinder Singh}, {and} \bibinfo{person}{Yishay
  Mansour}.} \bibinfo{year}{[n.d.]}\natexlab{}.
\newblock \showarticletitle{Policy Gradient Methods for Reinforcement Learning
  with Function Approximation}. In \bibinfo{booktitle}{\emph{Advances in Neural
  Information Processing Systems}}. \bibinfo{pages}{1057--1063}.
\newblock


\bibitem[\protect\citeauthoryear{Wang, Zheng, Hu, and Li}{Wang
  et~al\mbox{.}}{2013}]%
        {wang2013naturalness}
\bibfield{author}{\bibinfo{person}{Shuhang Wang}, \bibinfo{person}{Jin Zheng},
  \bibinfo{person}{Hai-Miao Hu}, {and} \bibinfo{person}{Bo Li}.}
  \bibinfo{year}{2013}\natexlab{}.
\newblock \showarticletitle{Naturalness preserved enhancement algorithm for
  non-uniform illumination images}.
\newblock \bibinfo{journal}{\emph{IEEE Transactions on Image Processing}}
  \bibinfo{volume}{22}, \bibinfo{number}{9} (\bibinfo{year}{2013}),
  \bibinfo{pages}{3538--3548}.
\newblock


\bibitem[\protect\citeauthoryear{Wei, Wang, Yang, and Liu}{Wei
  et~al\mbox{.}}{2018}]%
        {wei2018deep}
\bibfield{author}{\bibinfo{person}{Chen Wei}, \bibinfo{person}{Wenjing Wang},
  \bibinfo{person}{Wenhan Yang}, {and} \bibinfo{person}{Jiaying Liu}.}
  \bibinfo{year}{2018}\natexlab{}.
\newblock \showarticletitle{Deep retinex decomposition for low-light
  enhancement}.
\newblock \bibinfo{journal}{\emph{arXiv preprint arXiv:1808.04560}}
  (\bibinfo{year}{2018}).
\newblock


\bibitem[\protect\citeauthoryear{Yang, Wang, Fang, Wang, and Liu}{Yang
  et~al\mbox{.}}{2020}]%
        {Yang_2020_CVPR}
\bibfield{author}{\bibinfo{person}{Wenhan Yang}, \bibinfo{person}{Shiqi Wang},
  \bibinfo{person}{Yuming Fang}, \bibinfo{person}{Yue Wang}, {and}
  \bibinfo{person}{Jiaying Liu}.} \bibinfo{year}{2020}\natexlab{}.
\newblock \showarticletitle{From Fidelity to Perceptual Quality: A
  Semi-Supervised Approach for Low-Light Image Enhancement}. In
  \bibinfo{booktitle}{\emph{IEEE/CVF Conference on Computer Vision and Pattern
  Recognition (CVPR)}}.
\newblock


\bibitem[\protect\citeauthoryear{Yu, Dong, Lin, and Loy}{Yu
  et~al\mbox{.}}{2018}]%
        {yu2018crafting}
\bibfield{author}{\bibinfo{person}{Ke Yu}, \bibinfo{person}{Chao Dong},
  \bibinfo{person}{Liang Lin}, {and} \bibinfo{person}{Chen~Change Loy}.}
  \bibinfo{year}{2018}\natexlab{}.
\newblock \showarticletitle{Crafting a toolchain for image restoration by deep
  reinforcement learning}. In \bibinfo{booktitle}{\emph{Proceedings of the IEEE
  conference on computer vision and pattern recognition}}.
  \bibinfo{pages}{2443--2452}.
\newblock


\bibitem[\protect\citeauthoryear{Zhang, Zuo, and Zhang}{Zhang
  et~al\mbox{.}}{2018b}]%
        {zhang2018ffdnet}
\bibfield{author}{\bibinfo{person}{Kai Zhang}, \bibinfo{person}{Wangmeng Zuo},
  {and} \bibinfo{person}{Lei Zhang}.} \bibinfo{year}{2018}\natexlab{b}.
\newblock \showarticletitle{FFDNet: Toward a fast and flexible solution for
  CNN-based image denoising}.
\newblock \bibinfo{journal}{\emph{IEEE Transactions on Image Processing}}
  \bibinfo{volume}{27}, \bibinfo{number}{9} (\bibinfo{year}{2018}),
  \bibinfo{pages}{4608--4622}.
\newblock


\bibitem[\protect\citeauthoryear{Zhang, Isola, Efros, Shechtman, and
  Wang}{Zhang et~al\mbox{.}}{2018a}]%
        {zhang2018perceptual}
\bibfield{author}{\bibinfo{person}{Richard Zhang}, \bibinfo{person}{Phillip
  Isola}, \bibinfo{person}{Alexei~A Efros}, \bibinfo{person}{Eli Shechtman},
  {and} \bibinfo{person}{Oliver Wang}.} \bibinfo{year}{2018}\natexlab{a}.
\newblock \showarticletitle{The Unreasonable Effectiveness of Deep Features as
  a Perceptual Metric}. In \bibinfo{booktitle}{\emph{CVPR}}.
\newblock


\bibitem[\protect\citeauthoryear{Zhang, Zhu, Zha, Dauwels, and Wen}{Zhang
  et~al\mbox{.}}{2021}]%
        {zhang2021r3l}
\bibfield{author}{\bibinfo{person}{Rongkai Zhang}, \bibinfo{person}{Jiang Zhu},
  \bibinfo{person}{Zhiyuan Zha}, \bibinfo{person}{Justin Dauwels}, {and}
  \bibinfo{person}{Bihan Wen}.} \bibinfo{year}{2021}\natexlab{}.
\newblock \bibinfo{title}{R3L: Connecting Deep Reinforcement Learning to
  Recurrent Neural Networks for Image Denoising via Residual Recovery}.
\newblock
\newblock
\showeprint[arxiv]{2107.05318}


\bibitem[\protect\citeauthoryear{Zhang, Zhang, and Guo}{Zhang
  et~al\mbox{.}}{2019}]%
        {zhang2019kindling}
\bibfield{author}{\bibinfo{person}{Yonghua Zhang}, \bibinfo{person}{Jiawan
  Zhang}, {and} \bibinfo{person}{Xiaojie Guo}.}
  \bibinfo{year}{2019}\natexlab{}.
\newblock \showarticletitle{Kindling the darkness: A practical low-light image
  enhancer}. In \bibinfo{booktitle}{\emph{Proceedings of the 27th ACM
  International Conference on Multimedia}}. \bibinfo{pages}{1632--1640}.
\newblock


\bibitem[\protect\citeauthoryear{Zheng, Wu, Han, and Shi}{Zheng
  et~al\mbox{.}}{2020}]%
        {zheng_2020_ECCV}
\bibfield{author}{\bibinfo{person}{Ziqiang Zheng}, \bibinfo{person}{Yang Wu},
  \bibinfo{person}{Xinran Han}, {and} \bibinfo{person}{Jianbo Shi}.}
  \bibinfo{year}{2020}\natexlab{}.
\newblock \showarticletitle{ForkGAN: Seeing into the Rainy Night}. In
  \bibinfo{booktitle}{\emph{The IEEE European Conference on Computer Vision
  (ECCV)}}.
\newblock


\bibitem[\protect\citeauthoryear{Zhu, Zhang, Shen, Ma, Zhao, and Zhou}{Zhu
  et~al\mbox{.}}{2020}]%
        {zhu2020zero}
\bibfield{author}{\bibinfo{person}{Anqi Zhu}, \bibinfo{person}{Lin Zhang},
  \bibinfo{person}{Ying Shen}, \bibinfo{person}{Yong Ma},
  \bibinfo{person}{Shengjie Zhao}, {and} \bibinfo{person}{Yicong Zhou}.}
  \bibinfo{year}{2020}\natexlab{}.
\newblock \showarticletitle{Zero-shot restoration of underexposed images via
  robust retinex decomposition}. In \bibinfo{booktitle}{\emph{2020 IEEE
  International Conference on Multimedia and Expo (ICME)}}. IEEE,
  \bibinfo{pages}{1--6}.
\newblock


\bibitem[\protect\citeauthoryear{Zhu, Park, Isola, and Efros}{Zhu
  et~al\mbox{.}}{2017}]%
        {zhu2017unpaired}
\bibfield{author}{\bibinfo{person}{Jun-Yan Zhu}, \bibinfo{person}{Taesung
  Park}, \bibinfo{person}{Phillip Isola}, {and} \bibinfo{person}{Alexei~A
  Efros}.} \bibinfo{year}{2017}\natexlab{}.
\newblock \showarticletitle{Unpaired image-to-image translation using
  cycle-consistent adversarial networks}. In
  \bibinfo{booktitle}{\emph{Proceedings of the IEEE international conference on
  computer vision}}. \bibinfo{pages}{2223--2232}.
\newblock


\end{thebibliography}
\section*{Appendix:}

\begin{figure*}[h]
	\begin{center}
		%\vspace{-3mm}
		\begin{tabular}{c@{ }c@{ }c@{ }c}
							\includegraphics[width=.2\textwidth]{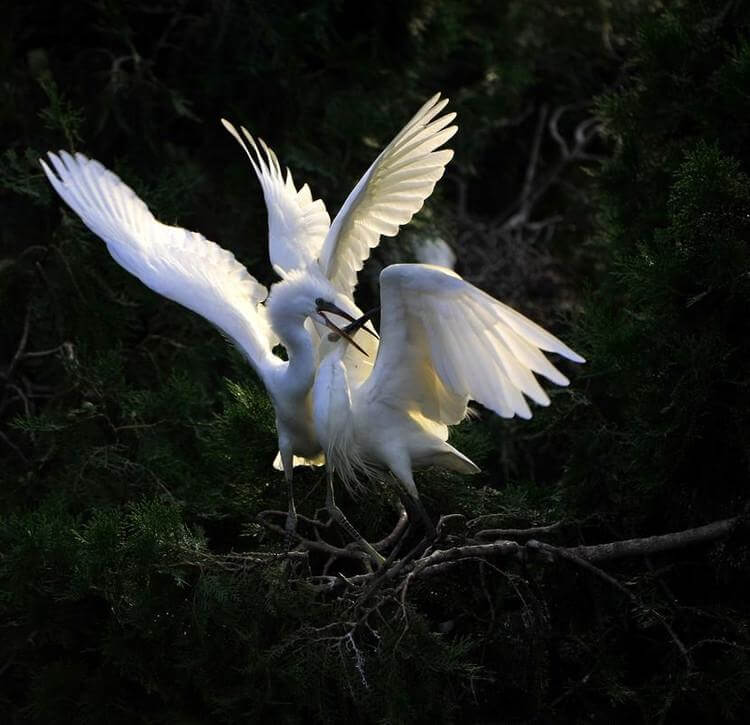}~&
			\includegraphics[width=.2\textwidth]{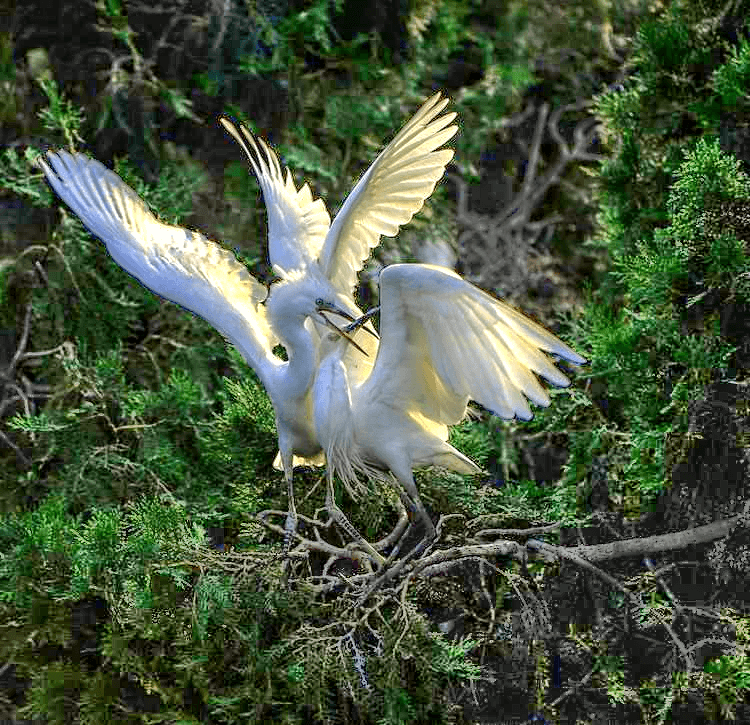}~&
			\includegraphics[width=.2\textwidth]{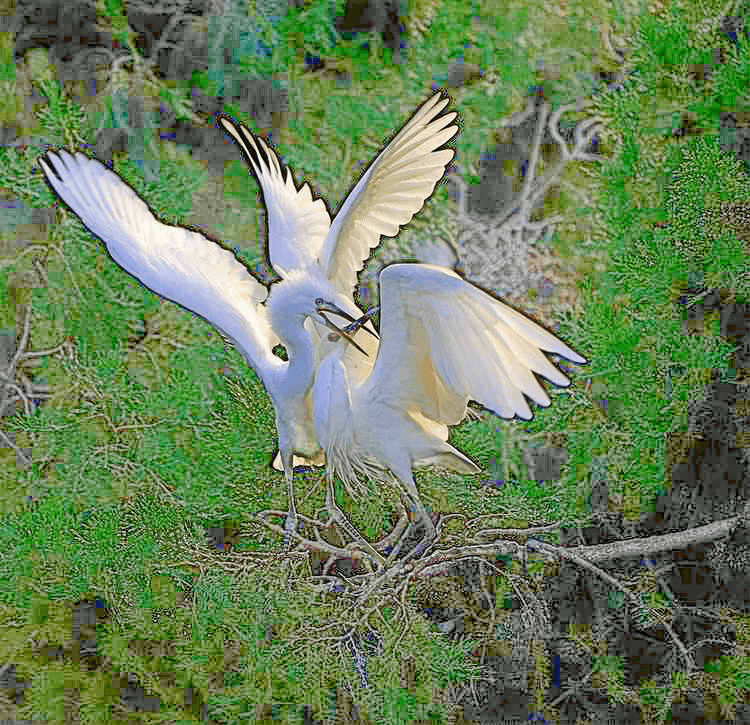}~&
			\includegraphics[width=.2\textwidth]{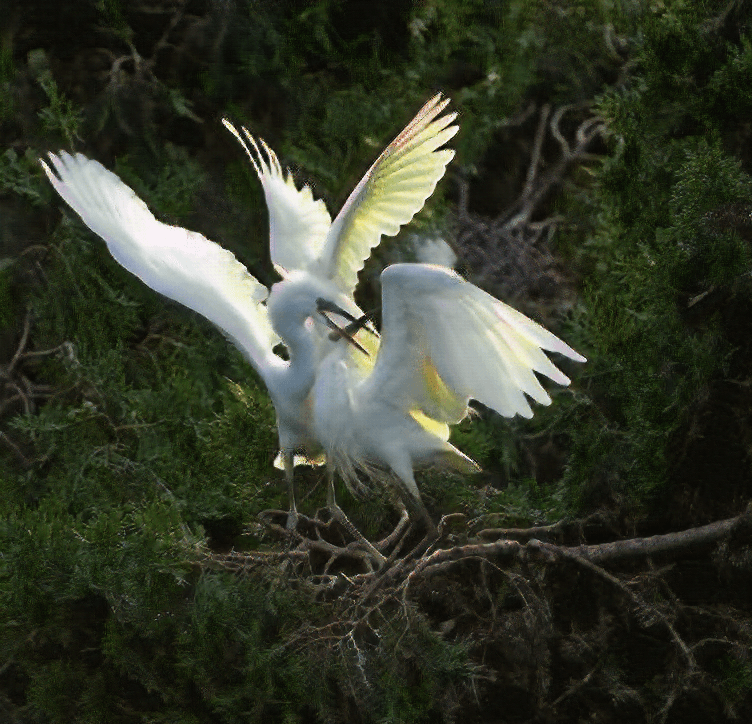}\\
			(a) Input~& (b) LIME~\cite{guo2016lime}~& (c) RetinexNet~\cite{wei2018deep}~& (d) CycleGAN~\cite{zhu2017unpaired}\\
			\includegraphics[width=.2\textwidth]{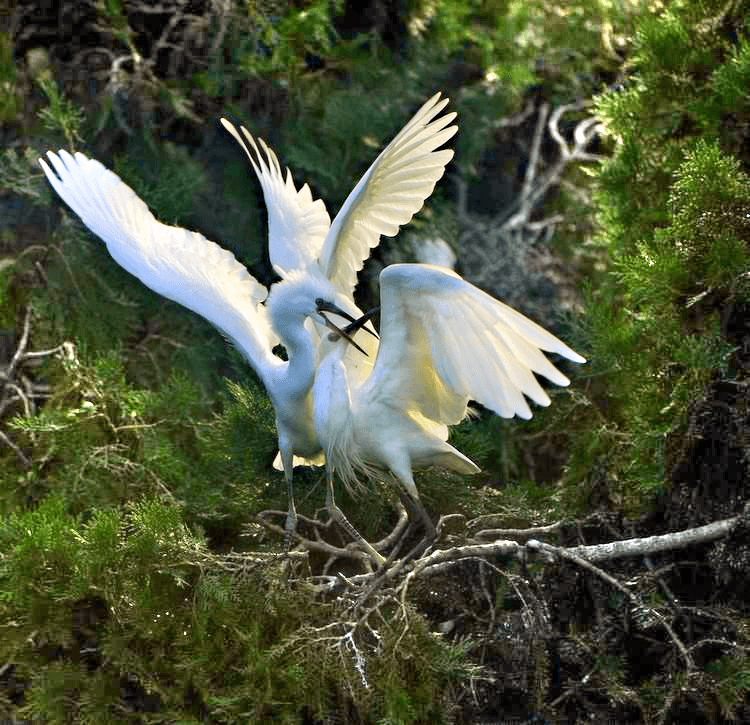}~&
			\includegraphics[width=.2\textwidth]{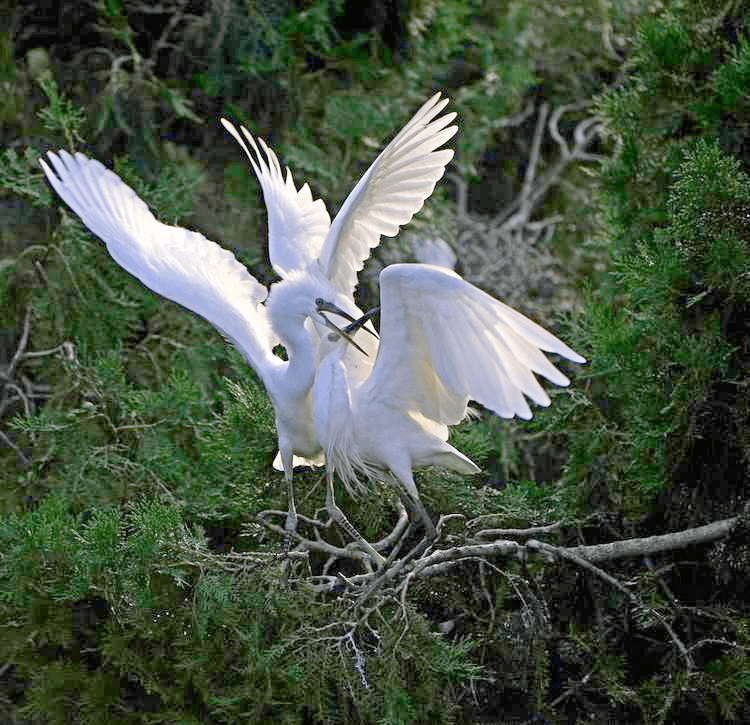}~&
			\includegraphics[width=.2\textwidth]{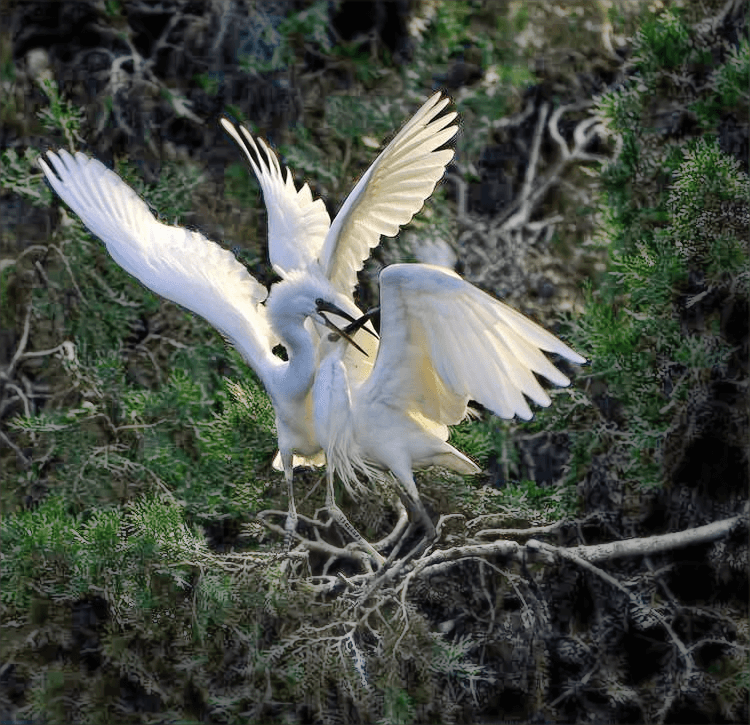}~&
			\includegraphics[width=.2\textwidth]{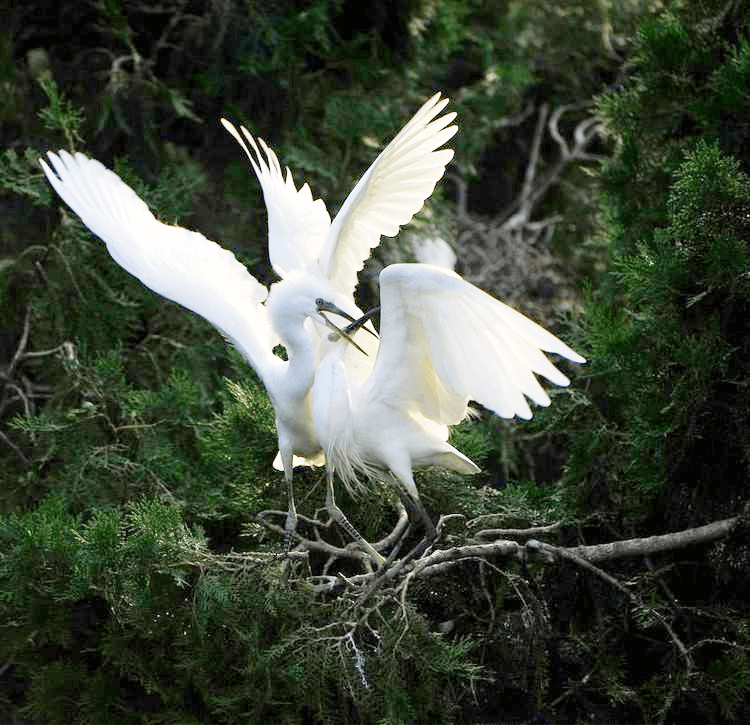}\\
						(e) EnlightenGAN~\cite{jiang2019enlightengan}-M~& (f) Zero-DCE~\cite{guo2020zero}~& (g) KinD~\cite{zhang2019kindling}~& (h) Ours\\
			\includegraphics[width=.2\textwidth]{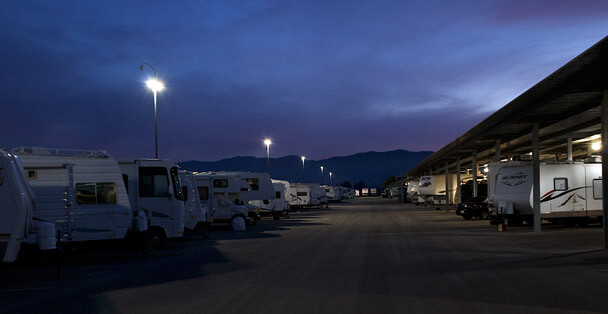}~&
			\includegraphics[width=.2\textwidth]{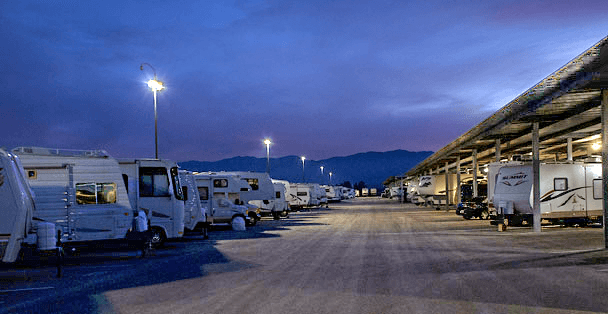}~&
			\includegraphics[width=.2\textwidth]{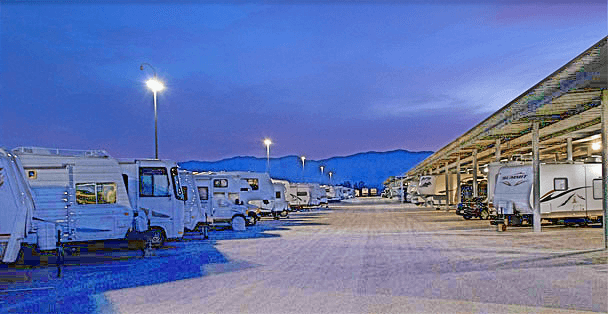}~&
			\includegraphics[width=.2\textwidth]{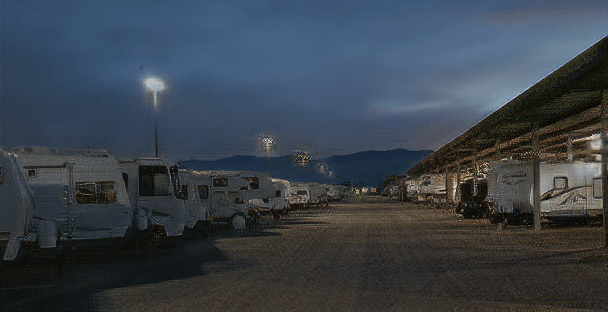}\\
			(a) Input~& (b) LIME~\cite{guo2016lime}~& (c) RetinexNet~\cite{wei2018deep}~& (d) CycleGAN~\cite{zhu2017unpaired}\\
			\includegraphics[width=.2\textwidth]{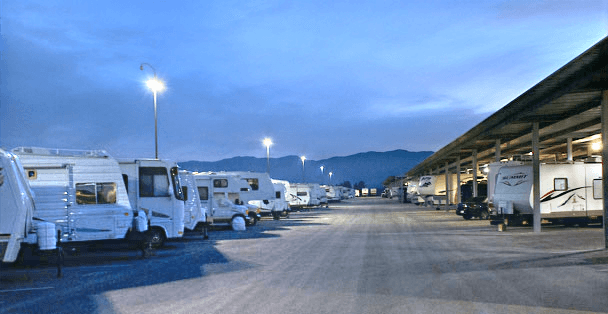}~&
			\includegraphics[width=.2\textwidth]{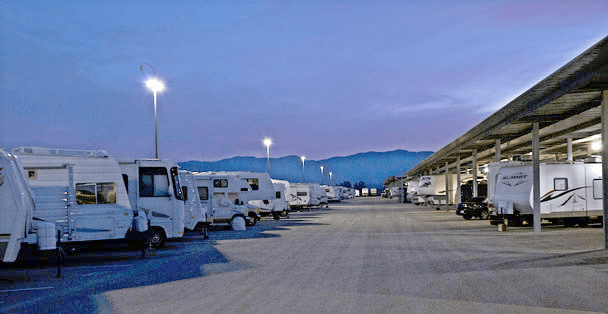}~&
			\includegraphics[width=.2\textwidth]{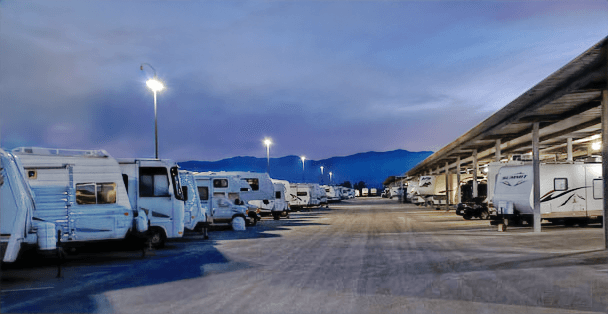}~&
			\includegraphics[width=.2\textwidth]{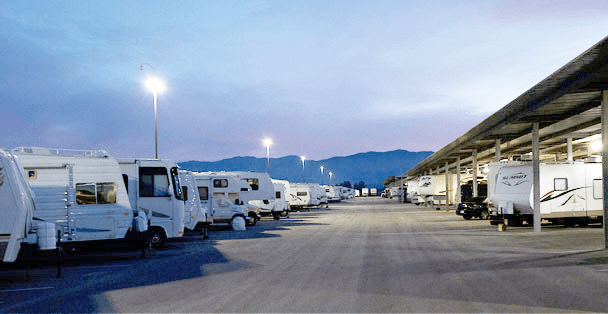}\\
						(e) EnlightenGAN~\cite{jiang2019enlightengan}-M~& (f) Zero-DCE~\cite{guo2020zero}~& (g) KinD~\cite{zhang2019kindling}~& (h) Ours\\	\includegraphics[width=.2\textwidth]{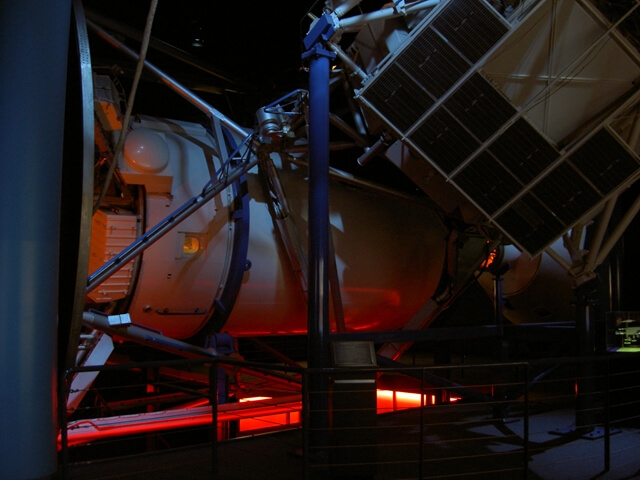}~&
			\includegraphics[width=.2\textwidth]{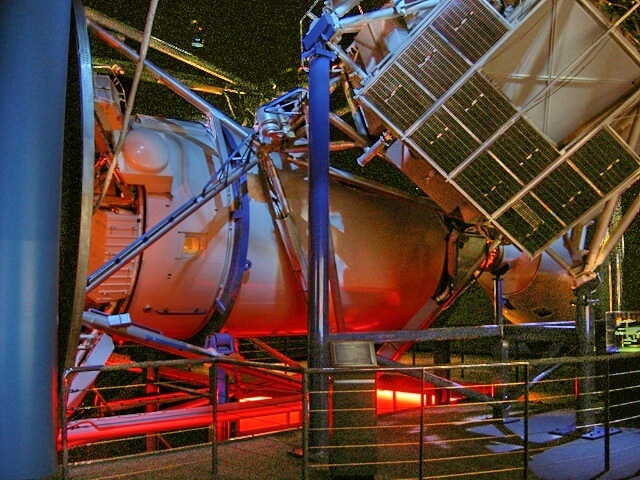}~&
			\includegraphics[width=.2\textwidth]{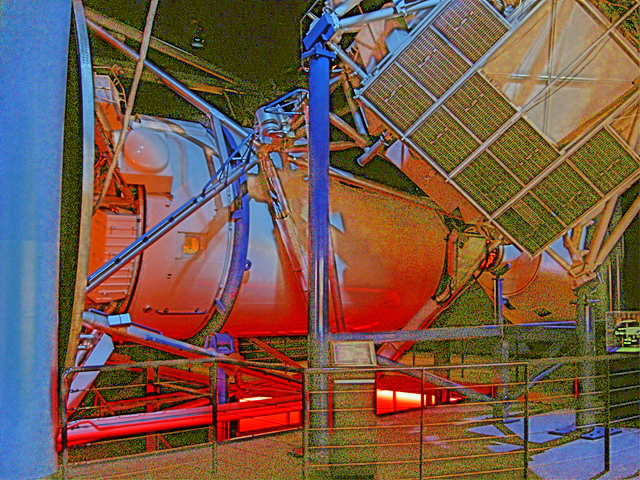}~&
			\includegraphics[width=.2\textwidth]{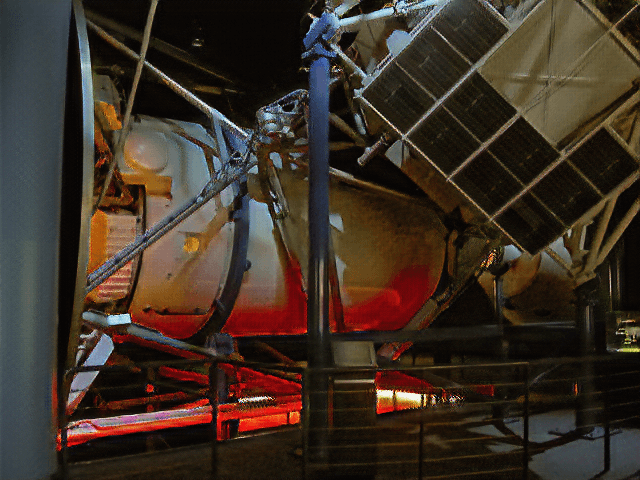}\\
			(a) Input~& (b) LIME~\cite{guo2016lime}~& (c) RetinexNet~\cite{wei2018deep}~& (d) CycleGAN~\cite{zhu2017unpaired}\\
			\includegraphics[width=.2\textwidth]{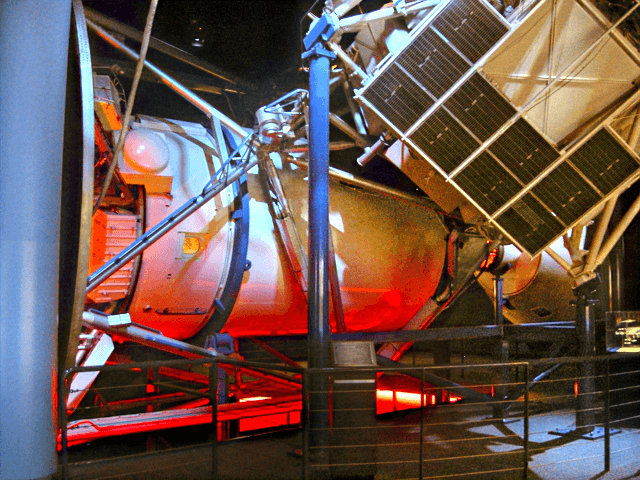}~&
			\includegraphics[width=.2\textwidth]{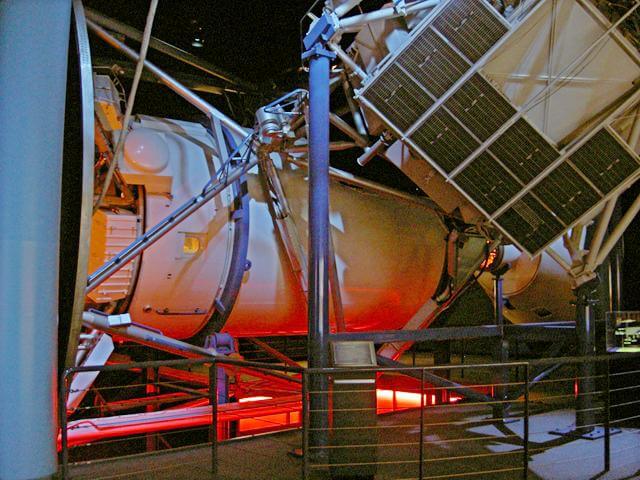}~&
			\includegraphics[width=.2\textwidth]{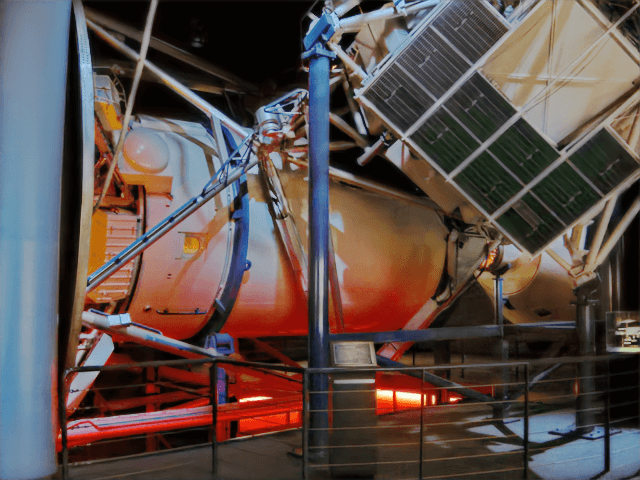}~&
			\includegraphics[width=.2\textwidth]{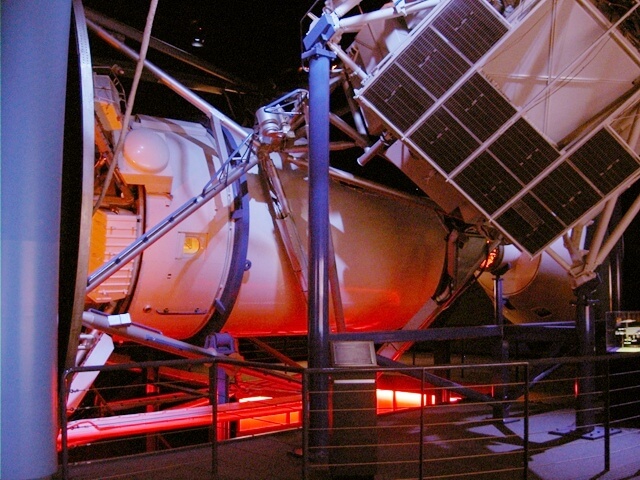}\\
						(e) EnlightenGAN~\cite{jiang2019enlightengan}-M~& (f) Zero-DCE~\cite{guo2020zero}~& (g) KinD~\cite{zhang2019kindling}~& (h) Ours\\

% 									\includegraphics[width=.23\textwidth]{img/NONREF_3/input.png}~&
% 			\includegraphics[width=.23\textwidth]{img/NONREF_3/input.png}~&
% 			\includegraphics[width=.23\textwidth]{img/NONREF_3/retinexnet.png}~&
% 			\includegraphics[width=.23\textwidth]{img/NONREF_3/cyclegan.png}\\
% 			(a) Input~& (b) LIME~\cite{guo2016lime}~& (c) RetinexNet~\cite{wei2018deep}~& (d) CycleGAN~\cite{zhu2017unpaired}\\
% 			\includegraphics[width=.23\textwidth]{img/NONREF_3/eg.png}~&
% 			\includegraphics[width=.23\textwidth]{img/NONREF_3/input.png}~&
% 			\includegraphics[width=.23\textwidth]{img/NONREF_3/kind.png}~&
% 			\includegraphics[width=.23\textwidth]{img/NONREF_3/ours.png}\\
% 						(e) EnlightenGAN~\cite{jiang2021enlightengan}-M~& (f) Zero-DCE~\cite{guo2020zero}~& (g) KinD~\cite{zhang2019kindling}~& (h) Ours\\
														
		\end{tabular}
	\end{center}
	\vspace{-3mm}
	\caption{Examples of enhancement results on NPE~\cite{wang2013naturalness} evaluation dataset. We show the estimated results of (b) LIME~\cite{guo2016lime}, (c) RetinexNet~\cite{wei2018deep}, (d) CycleGAN~\protect\cite{zhu2017unpaired}, (e) EnlightenGAN~\cite{jiang2019enlightengan}, (f) Zero-DCE~\cite{guo2020zero}, (g) KinD~\cite{zhang2019kindling} and (h) Ours. Zoom in to better see the details.}
	\label{fig:nonref1}
\end{figure*}

\begin{figure*}[!t]
	\begin{center}
		\begin{tabular}{c@{ }c@{ }c@{ }c}
			\includegraphics[width=.2\textwidth]{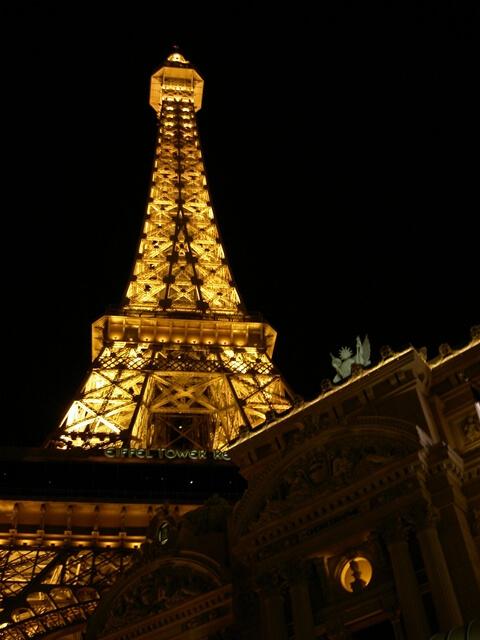}~&
			\includegraphics[width=.2\textwidth]{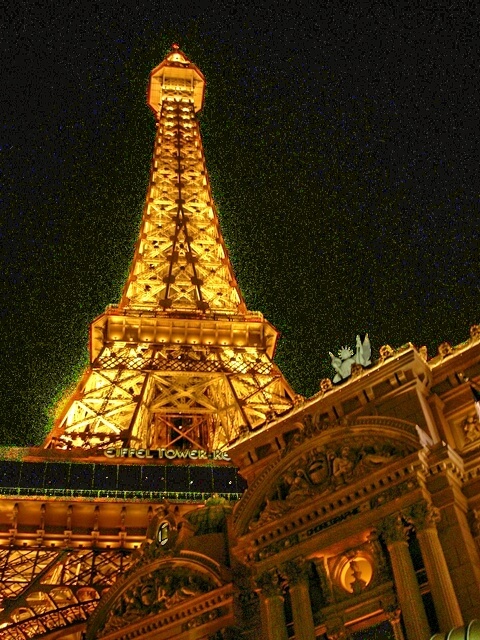}~&
			\includegraphics[width=.2\textwidth]{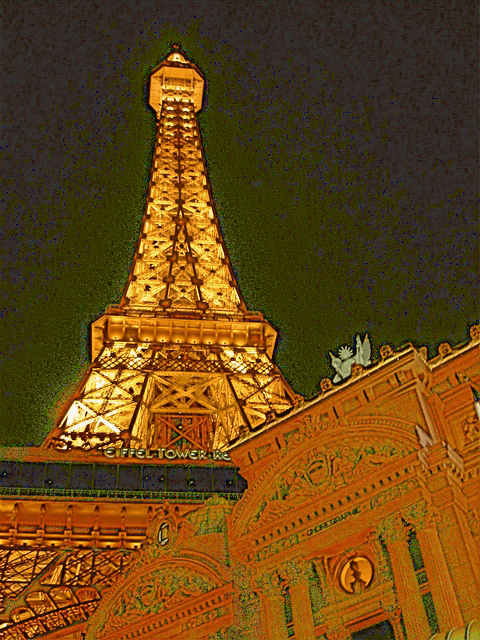}~&
			\includegraphics[width=.2\textwidth]{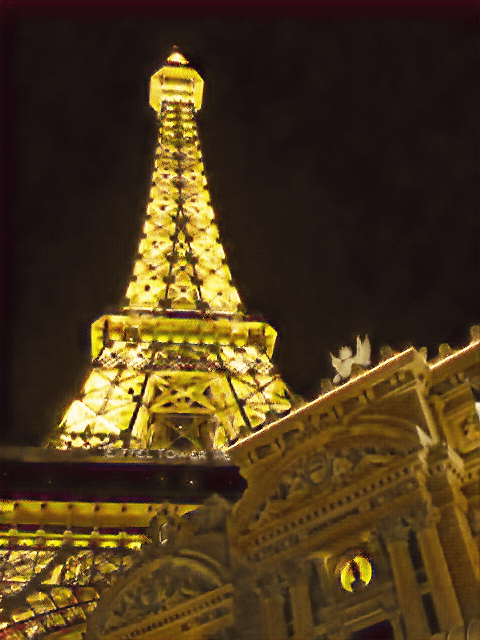}\\
			(a) Input~& (b) LIME~\cite{guo2016lime}~& (c) RetinexNet~\cite{wei2018deep}~& (d) CycleGAN~\cite{zhu2017unpaired}\\
			\includegraphics[width=.2\textwidth]{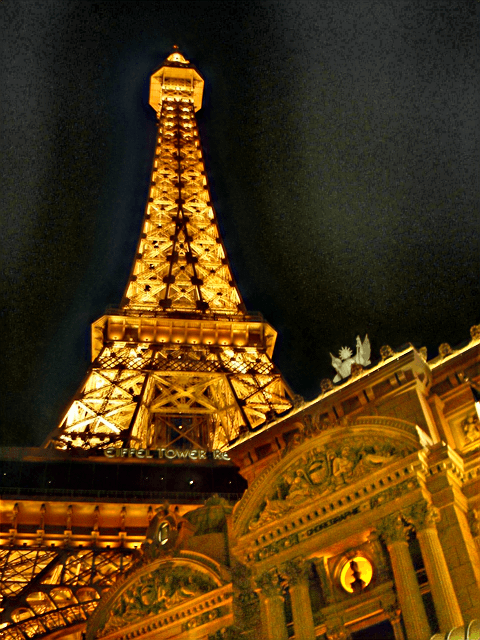}~&
			\includegraphics[width=.2\textwidth]{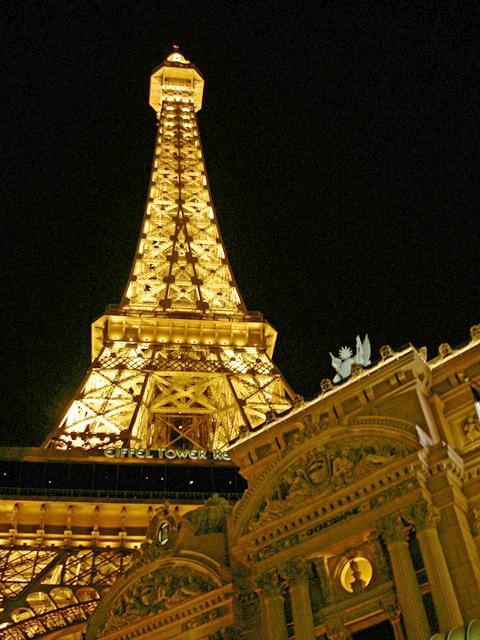}~&
			\includegraphics[width=.2\textwidth]{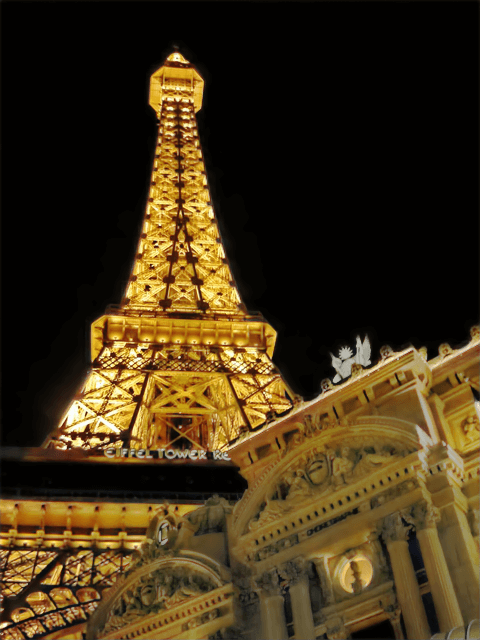}~&
			\includegraphics[width=.2\textwidth]{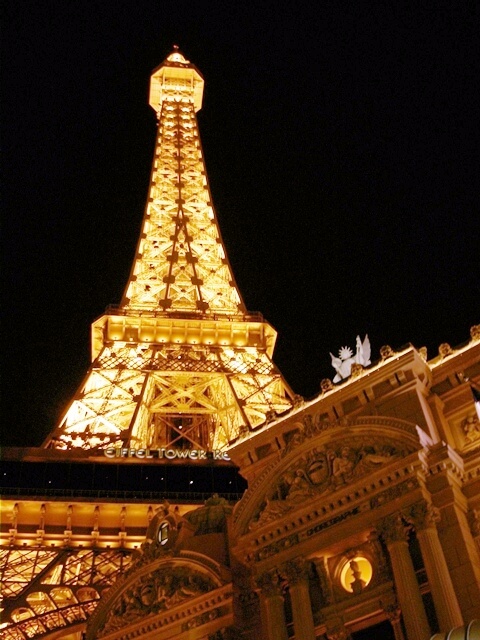}\\
						(e) EnlightenGAN~\cite{jiang2021enlightengan}-M~& (f) Zero-DCE~\cite{guo2020zero}~& (g) KinD~\cite{zhang2019kindling}~& (h) Ours\\
			\includegraphics[width=.2\textwidth]{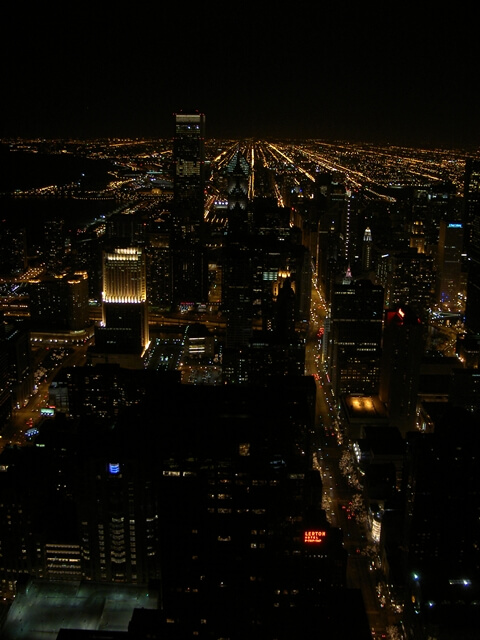}~&
			\includegraphics[width=.2\textwidth]{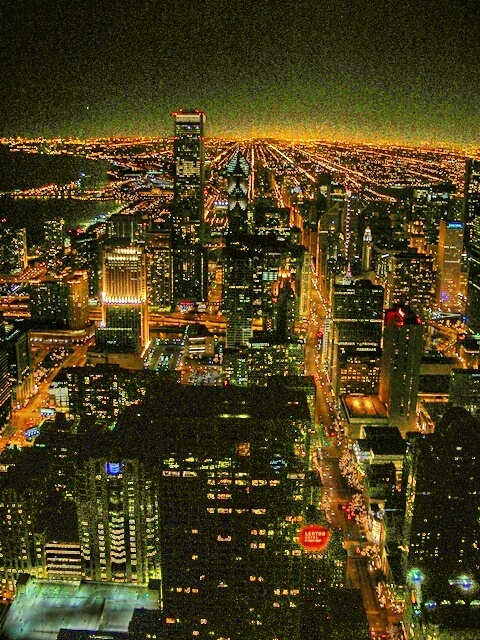}~&
			\includegraphics[width=.2\textwidth]{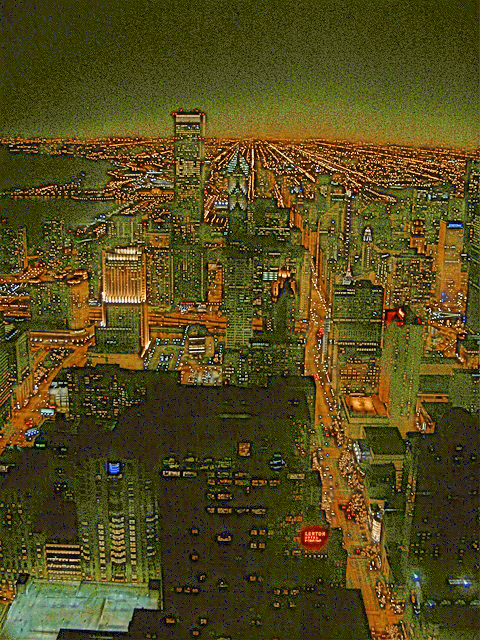}~&
			\includegraphics[width=.2\textwidth]{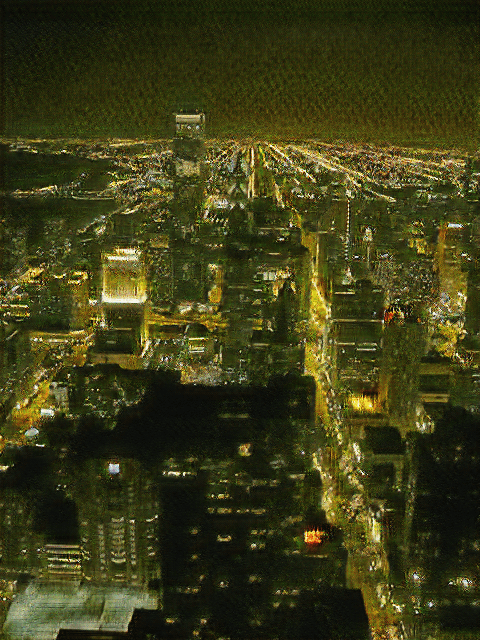}\\
			(a) Input~& (b) LIME~\cite{guo2016lime}~& (c) RetinexNet~\cite{wei2018deep}~& (d) CycleGAN~\cite{zhu2017unpaired}\\
			\includegraphics[width=.2\textwidth]{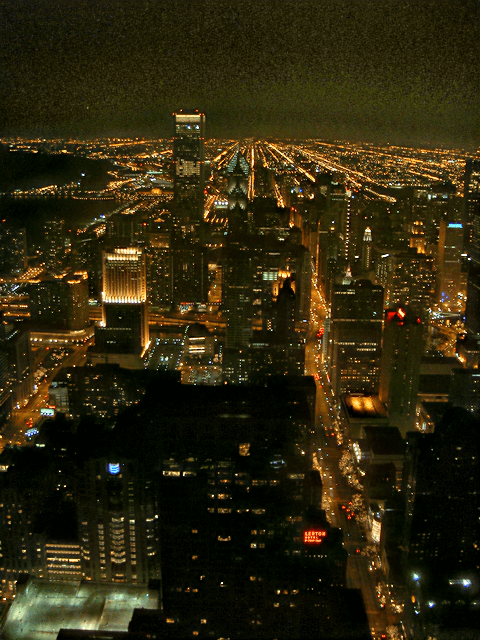}~&
			\includegraphics[width=.2\textwidth]{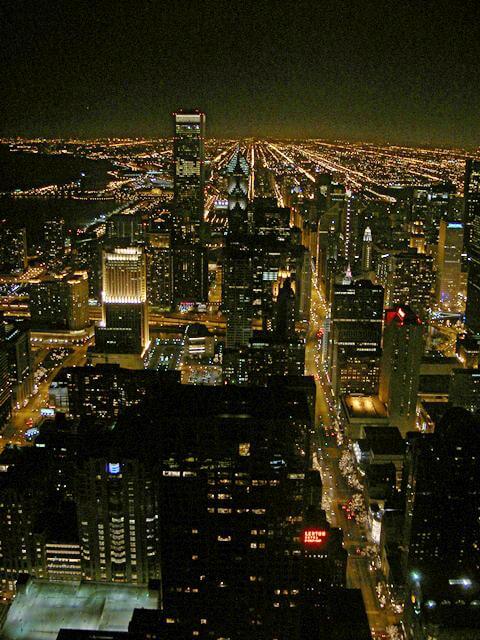}~&
			\includegraphics[width=.2\textwidth]{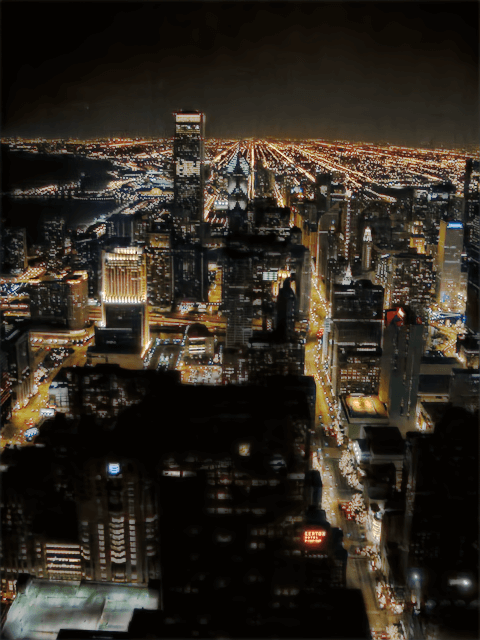}~&
			\includegraphics[width=.2\textwidth]{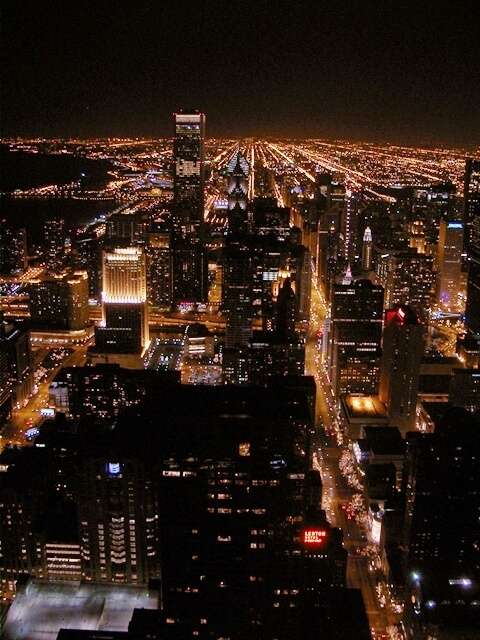}\\
						(e) EnlightenGAN~\cite{jiang2021enlightengan}-M~& (f) Zero-DCE~\cite{guo2020zero}~& (g) KinD~\cite{zhang2019kindling}~& (h) Ours\\
% 									\includegraphics[width=.23\textwidth]{img/NONREF_3/input.png}~&
% 			\includegraphics[width=.23\textwidth]{img/NONREF_3/input.png}~&
% 			\includegraphics[width=.23\textwidth]{img/NONREF_3/retinexnet.png}~&
% 			\includegraphics[width=.23\textwidth]{img/NONREF_3/cyclegan.png}\\
% 			(a) Input~& (b) LIME~\cite{guo2016lime}~& (c) RetinexNet~\cite{wei2018deep}~& (d) CycleGAN~\cite{zhu2017unpaired}\\
% 			\includegraphics[width=.23\textwidth]{img/NONREF_3/eg.png}~&
% 			\includegraphics[width=.23\textwidth]{img/NONREF_3/input.png}~&
% 			\includegraphics[width=.23\textwidth]{img/NONREF_3/kind.png}~&
% 			\includegraphics[width=.23\textwidth]{img/NONREF_3/ours.png}\\
% 						(e) EnlightenGAN~\cite{jiang2021enlightengan}-M~& (f) Zero-DCE~\cite{guo2020zero}~& (g) KinD~\cite{zhang2019kindling}~& (h) Ours\\
														
		\end{tabular}
	\end{center}
	\vspace{-3mm}
	\caption{Examples of enhancement results on DICM~\cite{lee2012contrast} evaluation dataset. We show the estimated results of (b) LIME~\cite{guo2016lime}, (c) RetinexNet~\cite{wei2018deep}, (d) CycleGAN~\protect\cite{zhu2017unpaired}, (e) EnlightenGAN~\cite{jiang2019enlightengan}, (f) Zero-DCE~\cite{guo2020zero}, (g) KinD~\cite{zhang2019kindling} and (h) Ours. Zoom in to better see the details.}
	\label{fig:nonref1}
\end{figure*}

\end{document}